# Fastener-centric acquisition and AND/OR graph representation of state-dependent disassembly knowledge from CAD-derived virtual assemblies

Shiki Iwase[a],*

[a]Toyota Central R&D Labs., Inc., 41-1 Yokomichi, Nagakute, Aichi 480-1192, Japan
*Corresponding author: shiki.iwase.me@mosk.tytlabs.co.jp

## Abstract

Design-stage disassemblability assessment requires knowledge of both product structure and the operations available in each assembly state. CAD assemblies provide geometric and structural information but do not directly specify state-dependent disassembly operations. This paper proposes a fastener-centric virtual-disassembly framework that acquires operation-state knowledge from CAD-derived virtual assemblies. Fasteners jointly releasing the same constraint are grouped into operation-level sets, avoiding exploration of arbitrary individual-fastener orders. Their removability is evaluated through state-dependent visibility, tool-access, and torque-driven virtual-removal trials. Following fastener removal, graph-based structural reasoning identifies candidate moving units, whose separation is evaluated through initial-motion screening, path generation, and physics-based validation. The acquired operations, states, dependencies, validated paths, and simulation-derived attributes are recursively organized into a Transformed AND/OR Graph for subsequent planning and design evaluation. Exhaustive verification using a controlled toy assembly found no branching violations for deliberately introduced tool-accessibility and precedence conditions. Evaluation using a 54-component OpenManipulator Chain gripper demonstrated fastener-removal and part- or subassembly-separation operations across successive assembly states. In this product-derived assembly, visibility and tool-accessibility evaluations pruned 51.4% of fastener-set candidate evaluations. Tool accessibility induced state-dependent precedence by enabling previously inaccessible fastener operations after obstructing components had been removed. Together, operation-level fastener grouping and state-dependent screening limited detailed physical validation to mechanically plausible candidates. The results demonstrate the technical feasibility of acquiring product-specific, state-dependent disassembly knowledge from CAD-derived virtual assemblies before manufacture, providing an upstream basis for subsequent disassembly reasoning and design-for-disassembly evaluation during product design.

## 1. Introduction

The increasing volume and complexity of end-of-life electromechanical products have intensified the need for efficient disassembly, reuse, remanufacturing, and recycling. Regulatory developments are also placing greater emphasis on circular product design and end-of-life recovery. In the European Union, Regulation (EU) 2026/1738 introduces requirements for vehicle design, dismantling, reuse, recycling, and recycled content, while the Ecodesign for Sustainable Products Regulation establishes a broader framework addressing product durability, reparability, reusability, and recyclability [1,2]. Because up to 80% of a product's environmental impacts are determined during the design phase [3], product designers are increasingly expected to consider end-of-life recovery, maintainability, and disassemblability alongside manufacturing and performance requirements. Design for Disassembly (DfD) supports this life-cycle-oriented approach by incorporating disassembly and recovery considerations into product design [4,5].

Improving disassembly involves two complementary approaches. One addresses products already in use or entering the end-of-life stage by developing processes and robotic systems capable of disassembling them efficiently [6]. Such applications must cope with product-condition uncertainty, including wear, fracture, deformation, and corrosion [7]. The other seeks to improve future products by incorporating disassemblability into product design before manufacturing. Both approaches are important; this study focuses on reducing future end-of-life burdens by acquiring the operation-level knowledge required to evaluate and support DfD during the design stage.

Assembly and disassembly planning has traditionally relied on liaison graphs, precedence constraints, interference matrices, Petri nets, and AND/OR-graph representations [8–13]. More recently, a hypergraph-based formulation derived precedence relations from a single cut-set matrix, eliminating the need to prepare separate precedence matrices for individual disassembly states [14]. In parallel with these developments in planning representations, disassembly-related entities and relations have been extracted from unstructured technical documents [15], while reusable assembly-hierarchy and shape-similarity knowledge has been organized from large CAD repositories [16]. These approaches provide useful representations for reasoning about product structure, subassembly hierarchies, shape similarity, and design reuse. However, neither approach directly addresses the state-dependent acquisition of constraint-release and physical-separation operations from a target CAD-derived assembly.

Beyond the acquisition of product-specific information, another challenge concerns the representation of fasteners. Many conventional formulations primarily describe disassembly as a part-ordering problem: fasteners may be omitted, incorporated implicitly into liaison constraints, represented through manually defined precedence relations, or treated as ordinary parts within the same combinatorial space as functional components. Such abstractions overlook both the effort and the structural role associated with fastener handling. Santochi et al. reported that connector separation and tooling accounted for 32% and 10%, respectively, of total disassembly time for electrical and electronic equipment [17]. Ghazilla et al. similarly found in their disassembly experiments that fastener separation accounted for approximately 30% to 40% of total disassembly time [18]. The practical importance of this operation has also motivated research on robotic screw removal from end-of-life products, including learning-based positioning of a robotic screwdriver relative to the target screw [19]. Moreover, fastener release determines when adjacent parts or subassemblies can be separated and thereby establishes operation precedence. Fasteners that jointly release the same constraint should therefore be represented as an operation-level set rather than as independent elements in the same part-removal permutation.

Reflecting the practical and structural importance of operation-level disassembly knowledge, product-specific DfD methods use such information for quantitative design evaluation. LeanDfD calculates target-component disassembly sequences using component liaisons, precedence relations, standard operation times, and corrective factors to assess disassemblability and recyclability [20]. The Disassembly Map represents alternative routes to target components and evaluates individual steps in terms of sequence depth, required tools, fastener characteristics, and disassembly-time estimates based on the Ease of Disassembly Metric [21,22]. Recent design-for-robotic-disassembly work has further adapted disassembly mapping to robot capabilities, tools, and robot action planning [23]. These methods provide detailed design feedback, but they assume that the relevant product structure, disassembly operations, and precedence relations have already been identified through product-specific analysis or specified in advance.

CAD models are typically available before physical prototypes are manufactured and therefore provide a promising information source for design-stage disassemblability analysis. CAD data describe the nominal product geometry and component structure, and some CAD-based methods derive sequence-relevant liaison and accessibility information directly from this geometry [24]. Raw CAD geometry does not itself encode which fastening operations are executable in a particular assembly state, whether the required tools can access the fasteners, or whether the resulting parts or subassemblies can be physically separated. A gap therefore remains between the product information encoded in a componentized CAD assembly and the integrated knowledge of fastener-mediated constraint release and subsequent physical separation required for disassembly planning and DfD evaluation. To address this gap, this paper proposes an integrated framework that combines CAD-based recognition, geometric reasoning, and physics-based validation to acquire operation-state knowledge for virtual disassembly.

The main technical contributions are as follows.

First, a fastener-centric method is proposed for acquiring state-dependent constraint-release and physical-separation knowledge from CAD-derived virtual assemblies. The method infers operation-level fastener sets, evaluates their state-dependent removability, distinguishes constraint release from subsequent separation, and preserves the released intermediate states required for further operation acquisition.

Second, the framework automatically acquires operation-related attributes for subsequent sequence evaluation and DfD analysis, including visibility, tool accessibility, removal direction, workpiece poses required for tool access, validated separation paths, moving-unit mass, and simulation-derived operation duration. This reduces the need to obtain such product-specific information through prior physical-disassembly observation or manual trajectory specification.

The acquired knowledge is organized into a Transformed AND/OR Graph (T-AOG) [25] and can be used to support subsequent selective and complete disassembly planning.

The remainder of this paper is organized as follows. Section 2 reviews CAD-derived relation acquisition, geometry-based disassembly planning, and physics-based virtual assembly and disassembly planning. Section 3 describes the proposed framework. Section 4 presents the experimental setup and results. Section 5 discusses the implications and limitations of the acquired knowledge. Section 6 concludes the paper and outlines its future use and extension.

## 2. Related Work

CAD-based assembly and disassembly research can be broadly distinguished according to whether motion feasibility is evaluated geometrically or through physics simulation. Geometry-based methods evaluate accessibility and separation using CAD-derived geometry and collision relations without simulating motion governed by equations of motion. Physics-based methods additionally use dynamic simulation and can account for mass, contact forces, friction, gravity, and related physical effects. The following sections examine how these two groups of methods derive product-specific relations, evaluate executable operations, and generate or validate removal motions.

### 2.1 CAD-derived relation acquisition and geometry-based disassembly planning

CAD-based studies have sought to derive disassembly-relevant information directly from product geometry. Learning-based methods have been used to recognize assembly features such as fasteners and to infer semantic relations between components. Adesso et al. proposed a framework that identifies fasteners in industrial 3D CAD assemblies using a convolutional neural network and prioritizes them during disassembly-sequence generation while accounting for initial and unavoidable collisions [26]. JoinABLe predicts joinable relations between parts for the bottom-up assembly of parametric CAD models [27], demonstrating that geometric and parametric information can support relation inference beyond contact detection. These studies demonstrate the feasibility of automatically recognizing assembly features and relations. The remaining issue addressed in this study is how recognized fasteners and component relations can be converted into operation-level constraint releases and distinguished from subsequent part or subassembly separation.

At the part level, earlier CAD-based methods reduced manual specification by deriving contact, interference, accessibility, and precedence information from assembly models. Contact, surface, mating, and interference information extracted from CAD or STEP assemblies has been used to construct precedence matrices and determine feasible part-removal directions [28,29]. Geometric accessibility has also been evaluated from product geometry, fastener information, and available removal space to determine whether individual parts can be disassembled [30]. Collectively, these methods derive planning inputs from product geometry but primarily evaluate part-level accessibility and geometric separation.

Beyond static part-level feasibility, CAD-derived relations have been used to update accessibility across assembly states and construct representations of alternative sequences. Rehal and Sen introduced shell structures derived from tessellated CAD assemblies to generate liaison information and update part accessibility during disassembly [24]. Münker and Schmitt extracted liaison and geometric precedence information from CAD models and applied top-down and bottom-up algorithms to generate AND/OR graphs for complex products [31]. These studies connect CAD-derived structural and precedence information to state-dependent sequence reasoning, although the executable operations represented by those relations remain defined primarily at the component level.

More recently, GLARE more closely integrated relation acquisition with disassembly-sequence generation. Harlacher and Friedrich combined semi-automatic learning and extraction of semantic spatial relations from imperfect CAD data with geometric-symbolic reasoning to generate component-removal actions and disassembly sequences [32]. The framework reconstructs a relational assembly model using congruent, concentric, and screwed relations and sequentially removes individual components while updating the product graph. However, its planning unit remains the sequential removal of individual components; it does not separately represent operation-level fastener-mediated constraint release and the subsequent physical separation of a part or subassembly.

Whereas the preceding studies focus mainly on relation acquisition, accessibility, and sequence-level reasoning, a complementary line of research has addressed removal-path generation and subassembly planning through geometric collision analysis. Hegewald et al. proposed iterative mesh modification planning for generating collision-free disassembly paths for complex industrial components [33]. Ebinger et al. developed a general framework supporting exhaustive and preemptive search, multiple part-separation techniques, and hierarchical planning through collision-based subassembly identification [34]. Together, these studies demonstrate how geometric feasibility can guide removal-path generation, sequence search, and subassembly formation.

At the integrated planning level, Kiyokawa et al. presented a CAD-based disassembly-planning framework that integrates sequence, task, motion, and scheduling decisions for a reconfigurable multi-robot system equipped with multiple tools and a rotary stage [35]. Their framework prioritizes fastener removal before the removal of connected non-fastener parts during sequence generation and explicitly accounts for tool changes and product reorientation. However, the framework plans component-level disassembly using contact, connection, and constraint relations, rather than separately acquiring operation-level fastener-mediated constraint releases and subsequent part or subassembly separations.

Taken together, these studies establish important foundations for fastener recognition, semantic relation inference, state-dependent accessibility analysis, CAD-derived planning representations, and geometry-based path and sequence planning. However, the reviewed methods do not jointly group fasteners according to the constraints they release, distinguish their removal from subsequent part or subassembly separation, and acquire both as state-dependent executable operations.

## 2.2 Physics-based virtual assembly and disassembly planning

Physics-based studies have incorporated dynamic simulation directly into assembly and disassembly planning, ranging from contact-rich operation simulation to sequence and motion generation. *Factory* demonstrated efficient contact-rich simulation of threaded nut-and-bolt interactions and provided corresponding robotic assembly environments and controllers [36]. *Assemble Them All* uses assembly-by-disassembly reasoning and physics-based path planning to generate physically plausible assembly sequences and motions [37]. Candidate parts are evaluated for removability from the current assembly state, and successful removals are accumulated into a disassembly sequence that is subsequently reversed to obtain the assembly sequence.

*ASAP* extends this approach through a disassembly-tree formulation in which nodes represent partial assemblies and transitions represent candidate part-removal operations [38]. Geometric heuristics or a graph neural network prioritize candidate parts, while path existence, subassembly stability, and robotic executability determine whether a transition is feasible.

Zhu et al. integrated learned part-sequence inference, motion planning, and robot-contact optimization in a multi-level robotic assembly framework [39]. Their method extends part-level assembly planning from sequence inference toward motion and contact selection, but retains part insertion as the principal operation unit rather than explicitly representing preceding constraint-release operations.

Beyond sequence and motion planning, *Fabrica* extends the scope toward robotic execution by integrating precedence, sequence, grasp, fixture, and motion planning with simulation-trained reinforcement learning for contact-rich assembly [40]. RRT-Connect generates transit and transfer motions, while a policy trained in simulation is transferred to the physical system for contact-rich assembly steps.

Collectively, these studies incorporate contact-rich operation simulation and geometric, physical, or robotic feasibility into assembly or disassembly planning. However, the principal sequence-level decision unit generally remains the insertion or removal of a product part, while the preceding fastener-mediated or other constraint-release operations and their state-dependent conditions are not acquired together with the executable separation operation. The proposed framework addresses this gap at the knowledge-acquisition level by recording fastener-mediated constraint-release operations, tool accessibility, product poses required for tool access, and physically validated part- and subassembly-separation motions. It does not optimize a disassembly sequence; instead, the resulting operations, states, attributes, and dependencies are retained as reusable knowledge for subsequent planning and DfD evaluation.

## 3. Fastener-centric virtual disassembly framework

The proposed framework formalizes a common iterative process in practical disassembly: observing the current assembly state, selecting an accessible release operation, and reassessing the resulting state before proceeding. An operator identifies an exposed and tool-accessible fastener, removes it, and then determines whether the constrained part or subassembly can be separated. If separation remains blocked by a non-fastener constraint, an additional release action, such as disengaging an interlock or snap-fit, is considered. The assembly is observed again after each successful operation, and the process is repeated from the resulting state.

This iterative process is implemented using CAD-derived structural information and virtual-disassembly trials. As summarized in Figure 1, Phase 1 constructs an annotated liaison graph representing the product structure, recognized fasteners, and their associated constraints. Phase 2 evaluates executable constraint-release and separation operations in each assembly state. Successful operations and resulting states are recorded as reusable operation-state knowledge in a Transformed AND/OR Graph (T-AOG).

The input is a CAD-derived assembly imported in its nominal configuration into a virtual environment. Each physical component, including each bolt and nut, is represented as an individual CAD part. Meaningful contacts between assembled components are assumed, whereas artificial penetrations producing unintended contacts must be absent. Detailed mating-thread geometry is required only when torque-driven fastener removal is physically simulated; small local penetrations between mating threads are permitted if they do not produce contacts with unrelated components. Product-specific fastener-interface associations, precedence relations, removal trajectories, and operation sequences are not required as inputs.

To represent the stationary support or clamping condition during virtual disassembly, one product component may optionally be designated as fixed. When a fastener-removal or other constraint-release operation causes the remaining assembly to divide into multiple connected units, the unit containing the fixed component is treated as the stay unit, while the other units are evaluated as candidate moving units. If no fixed component is specified, the stay unit is selected automatically during recursive operation acquisition. The stay-unit selection defines the support condition assumed during exploration and therefore affects the separation operations, resulting poses, and branches recorded in the T-AOG.

In Phase 1, inter-component contacts are detected, and operation-level fastener sets and their interface associations are inferred. Fastener-mediated and other active constraints are represented as constraint edges in the annotated liaison graph. Together with the root assembly state, the resulting graph provides the structural basis for Phase 2.

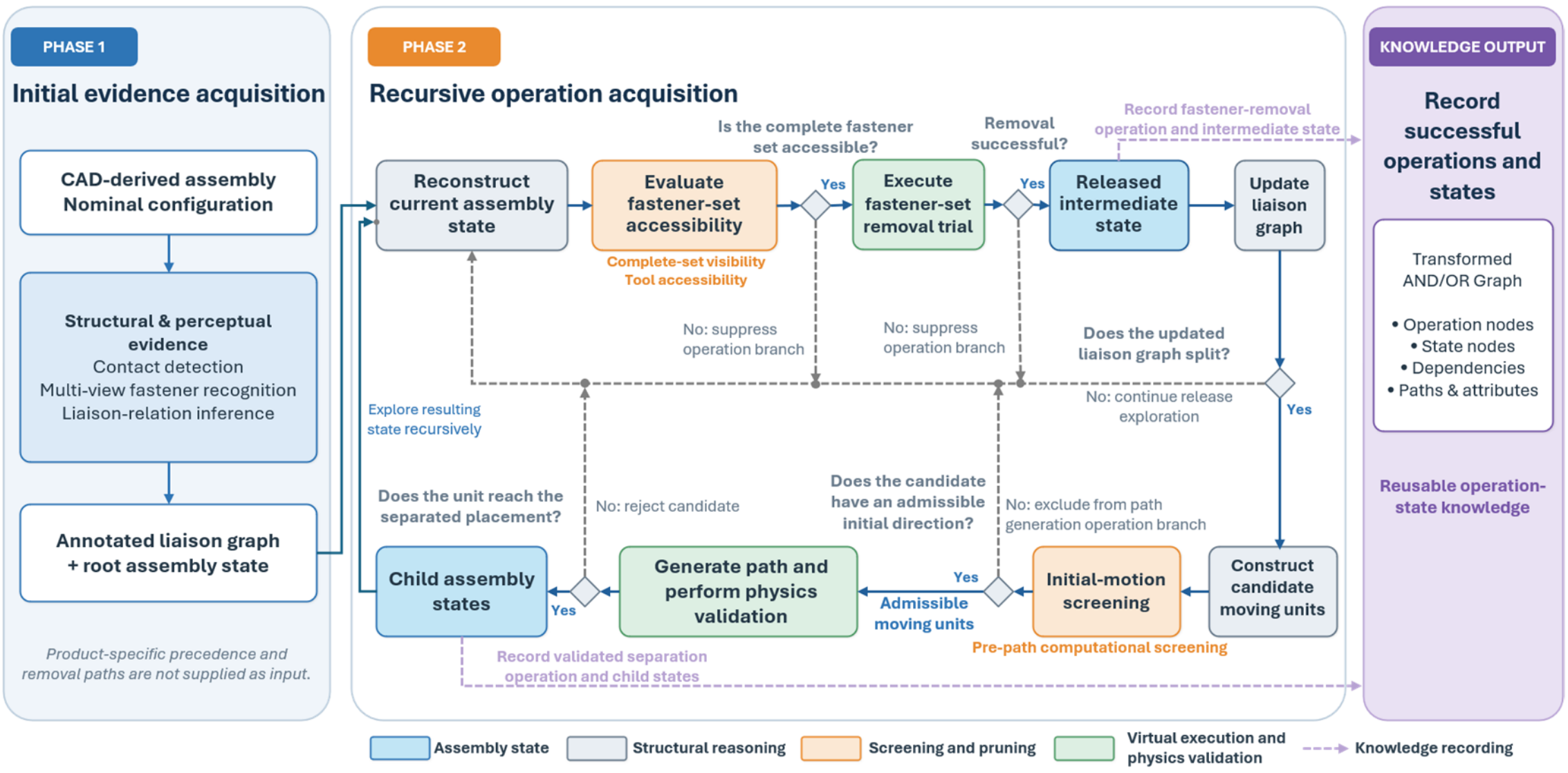

Figure 1. Overall workflow of the proposed fastener-centric virtual-disassembly framework, consisting of initial evidence acquisition, recursive operation acquisition, and T-AOG recording. An initial conceptual layout for this explanatory figure was generated with assistance from Microsoft 365 Copilot using GPT-6.1 Sol, based on the author's code. The author subsequently redrew, restructured, and verified the final figure.

In Phase 2, fastener sets are screened for visibility and tool accessibility in each reconstructed assembly state. Successful virtual removal produces a released intermediate state. The updated assembly structure is then examined for separable moving units. Retained units undergo initial-motion screening, path generation, and physics-based validation. Each validated resulting state is subsequently explored in the same manner.

Algorithm 1 formalizes the queue-based exploration outlined in Figure 1. A reconstructed assembly state may contain multiple fastener sets that must be evaluated as candidate removal operations. Before each candidate fastener-set removal operation is evaluated, the virtual assembly is restored to the same parent state. The candidates are therefore evaluated independently, and their computational evaluation order does not define disassembly precedence.

After a successful fastener-removal or other constraint-release operation, the corresponding constraint edges are removed from the liaison graph, and its connected components are recomputed. The connected component designated as the stay unit is excluded from path validation, while the remaining connected components are evaluated as candidate moving units. Resulting states from successfully validated separations are recorded for subsequent exploration. If the released state does not yield an admissible moving unit for path generation, it is retained for subsequent constraint-release exploration.

Algorithm 1. Top-level workflow for fastener-centric virtual-disassembly knowledge acquisition

```
Input:
    CAD-derived virtual assembly
    Component information
    Perception, accessibility, path-generation, and validation configurations
Output:
    T-AOG-compatible executable disassembly knowledge

records ← empty set
state_queue ← empty queue
graph_0 ← AcquireInitialEvidence()
root_state ← CreateRootState(graph_0)
Push(state_queue, root_state)

while state_queue is not empty do
    state ← Pop(state_queue)
    if state contains a single component then
        continue
    end if

    ReconstructState(state)
    fastener_sets ← GetFastenerSetsPresentInState(state)
    residual_releases ← GetResidualReleasesPresentInState(state)

    for each fastener_set in fastener_sets do
        ReconstructState(state)
        group_orders ← GetToolApproachGroupOrders(state, fastener_set)
        for each group_order in group_orders do
            ReconstructState(state)
            trial ← EvaluateFastenerSetRemoval(state, fastener_set, group_order)
            if trial is successful then
                RecordFastenerRemoval(records, fastener_set, group_order, trial)
                AnalyzeReleasedState(state, trial, records)
            end if
        end for
    end for

    for each residual_release in residual_releases do
        ReconstructState(state)
        if IsResidualReleaseEligible(state, residual_release) then
            result ← ApplyResidualRelease(state, residual_release)
            if result is successful then
                RecordResidualRelease(records, residual_release, result)
            end if
        end if
    end for

    for each child_state in RecordedStates(records) do
        if child_state has not previously been queued then
            Push(state_queue, child_state)
        end if
    end for
end while

return ConvertToTAOG(records)
```

Algorithm 1. Continued

```
Procedure AnalyzeReleasedState(parent_state, trial, records)
constraint_graph ← ActiveConstraintGraph(parent_state, trial)
connected_units ← ConnectedComponents(constraint_graph)

local_queue ← empty queue
Push(local_queue, (trial, connected_units))

while local_queue is not empty do
    released_state, connected_units ← Pop(local_queue)
    ReconstructState(released_state)

    if connected_units contains fewer than two units then
        RetainForFurtherExploration(records, released_state)
        continue
    end if

    moving_units ← SelectMovingUnits(connected_units, released_state)
    admissible_units, directions ←
        ScreenInitialDirections(moving_units, released_state)

    if admissible_units is empty then
        RetainForFurtherExploration(records, released_state )
        continue
    end if

    paths ← GeneratePaths(admissible_units, directions, released_state)
    results ← ValidatePaths(paths, released_state)

    for each successful result in results do
        RecordSeparation(records, result)
        remaining_state, remaining_units ←
            RemainingReleasedState(released_state, connected_units, result )
        if remaining_state exists then
            if remaining_units contains more than one unit then
                Push(local_queue, (remaining_state, remaining_units))
            else
                RetainForFurtherExploration(records, remaining_state)
            end if
        end if
    end for
end while

end procedure
```

RecordSeparation(records, result) records the validated separation operation and all resulting child states, reusing equivalent existing states as described in Section 3.3. After a fastener-mediated constraint release, remaining states containing multiple connected units are processed through a local queue, and their remaining moving units are screened again under the updated obstacle configuration. RecordedStates(records) returns the recorded assembly states retained for subsequent constraint-release exploration, including applicable released intermediate states, while states already expanded through local separation are not expanded again.

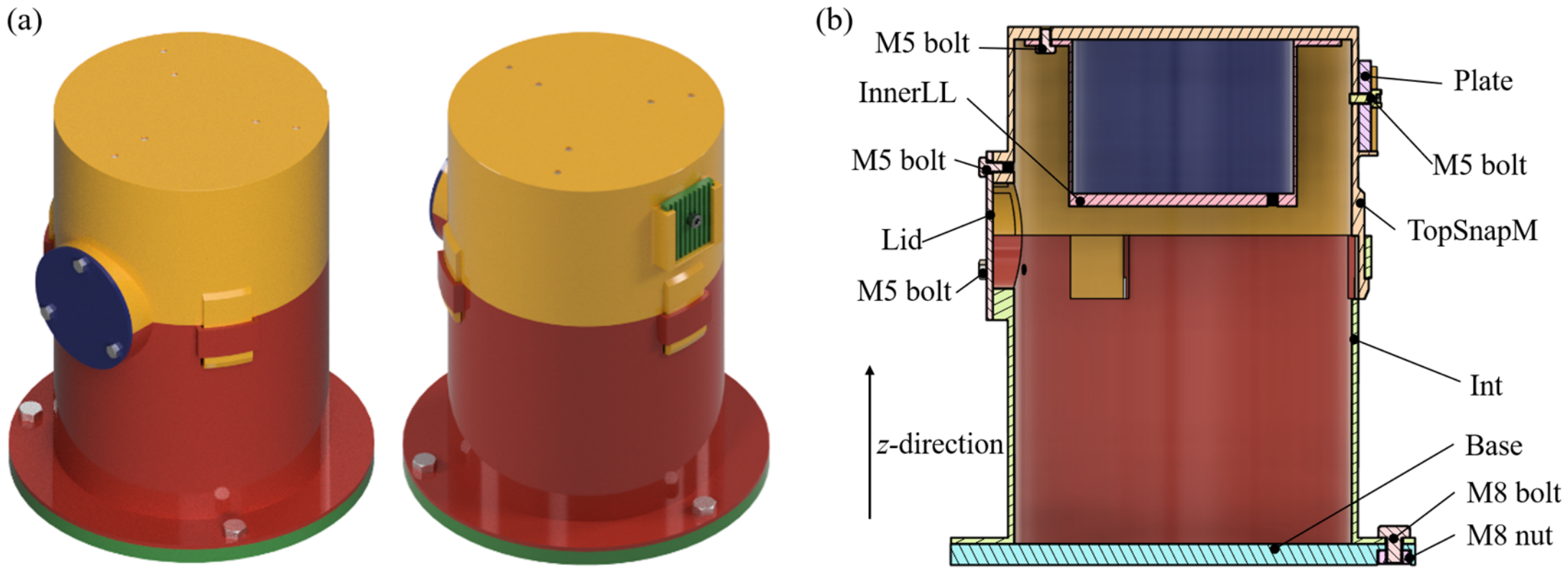

Figure 2. Toy assembly with six non-fastener components used as the running example: (a) nominal assembly shown from two viewpoints and (b) cross-sectional view showing its internal fastened and geometrically constrained interfaces.

A toy assembly comprising six non-fastener components is used as a running example throughout Section 3, as shown in Figure 2. The assembly was adapted from an earlier AND/OR-graph study [14] and modified to include the fastening and residual-interface configurations evaluated in this work. It contains seven M5 bolts, three M8 bolts, and three nuts, as well as a snap-fit-like interface between TopSnapM and Int and a guided sliding component, Plate. These features provide both fastener-mediated and geometry-based constraints.

Throughout this paper, executable denotes satisfaction of the state-dependent evaluation conditions defined within the configured virtual environment; it does not imply executability by a particular robot or under unmodeled physical-product conditions.

The research software used in this study was developed by the author with assistance from OpenAI ChatGPT and Codex for code inspection, implementation, debugging, and the preparation of diagnostic and verification code. All AI-assisted code was reviewed, tested, and verified by the author, who takes full responsibility for the implemented methods and reported results.

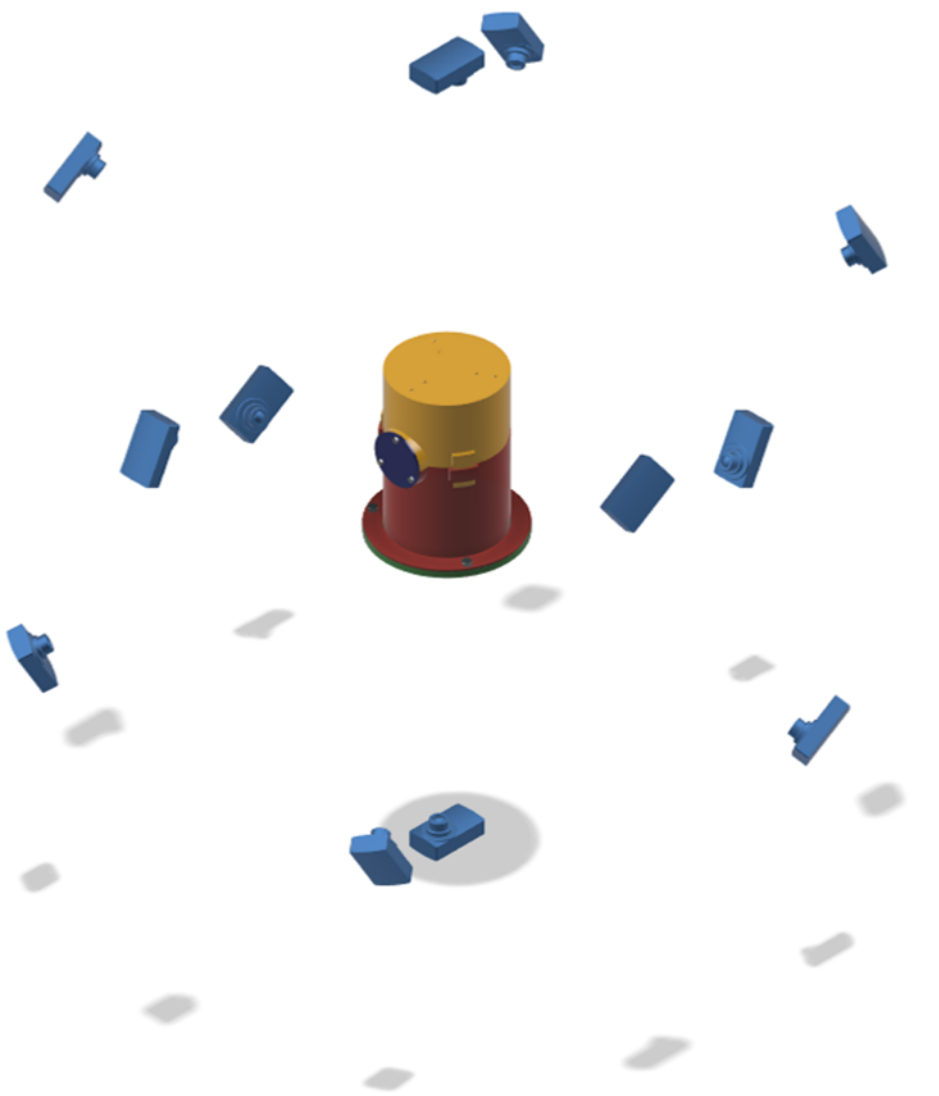

Figure 3. Camera arrangement with 12 icosahedrally distributed viewpoints for isolated component rendering and MVCNN-based fastener recognition.

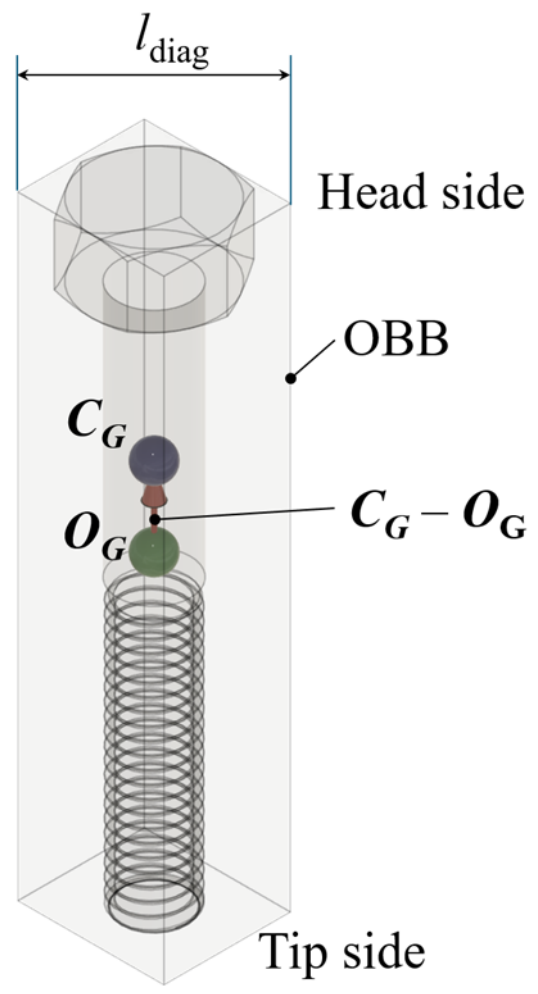

Figure 4. Bolt-removal direction estimated from the offset between the geometry centroid $\boldsymbol{C}_{\boldsymbol{G}}$ and OBB center $\boldsymbol{O}_{\boldsymbol{G}}$.

## 3.1 Phase 1: Initial evidence acquisition

### 3.1.1 Initial liaison graph construction and fastener recognition

Initial liaison-graph construction begins by identifying the component categories to be represented as nodes. Bolts, nuts, and other product components are represented as individual nodes, while inter-component contact evidence defines the graph edges.

When explicit fastener information is unavailable in the CAD model, bolt-like and nut-like candidates are recognized using a multi-view convolutional neural network (MVCNN) [41]. Each target component is isolated by hiding the other components and rendered from 12 approximately uniformly distributed viewpoints using the icosahedral camera arrangement shown in Figure 3. This prevents surrounding geometry from dominating the visual evidence and enables component-wise recognition.

Fastener-category predictions are evaluated together with geometric attributes describing thread, hole, and head features to distinguish bolt-like and nut-like candidates from other components. The detailed acceptance and rejection rules are provided in Appendix A.

For each accepted bolt-like candidate, the longitudinal axis, head side, removal direction, and nominal shank diameter are estimated from its polygonal geometry. As illustrated in Figure 4, an oriented bounding box (OBB) is aligned with the bolt geometry. Let $\boldsymbol{C}_G$ and $\boldsymbol{O}_G$ denote the geometry centroid and OBB center, respectively. Because the enlarged head shifts $\boldsymbol{C}_G$ toward the head relative to $\boldsymbol{O}_G$, the bolt-removal direction is estimated from the threaded end toward the head as

$$\boldsymbol{d}_{\mathbf{bolt}} = \frac{\boldsymbol{C}_G - \boldsymbol{O}_G}{\|\boldsymbol{C}_G - \boldsymbol{O}_G\|}. \tag{1}$$

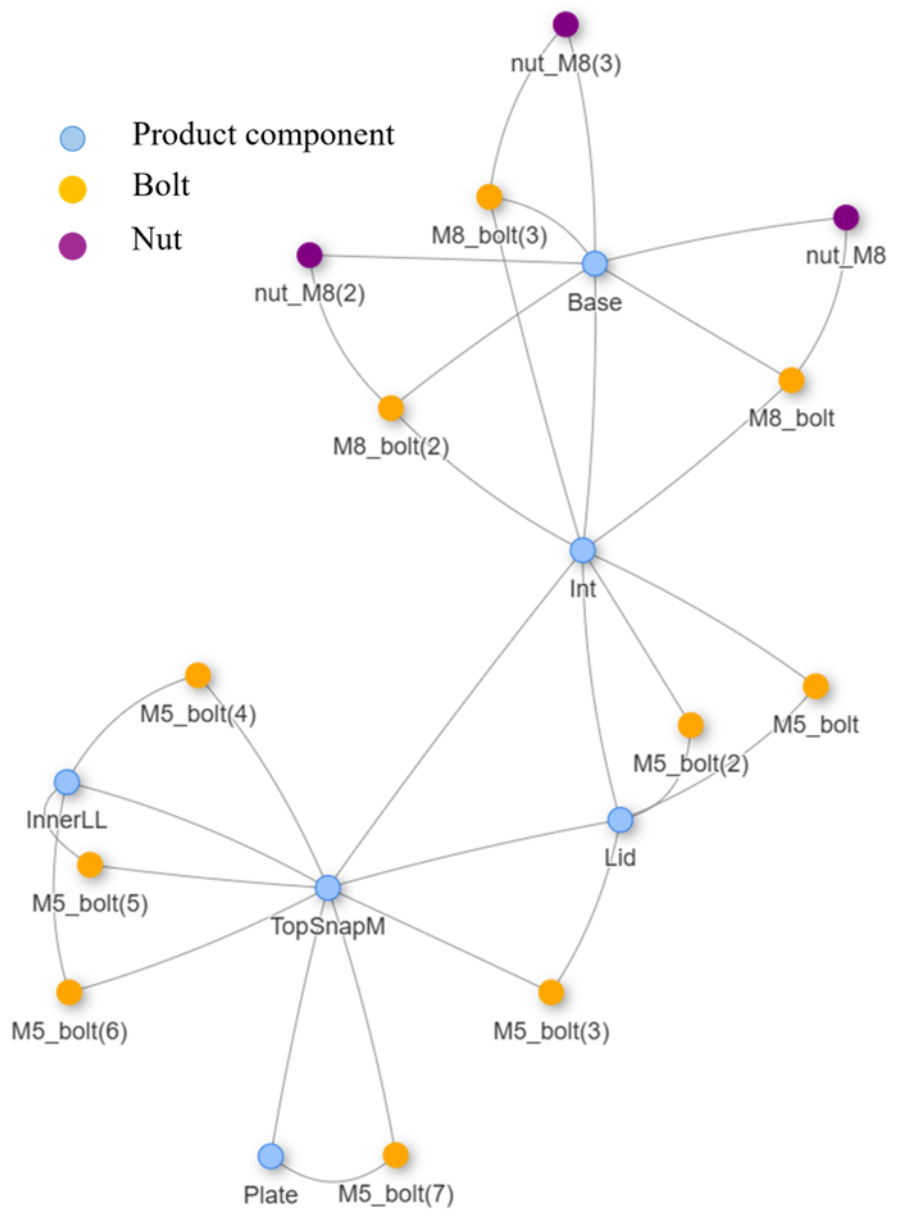

Figure 5. Annotated initial liaison graph of the running example.

The nominal shank diameter is estimated from three polygonal-mesh cross-sections perpendicular to $\boldsymbol{d}_{\mathbf{bolt}}$. The estimated direction and diameter are subsequently used for direction-based fastener grouping, state-dependent tool-access evaluation, and torque selection during virtual removal.

The liaison-graph edges are then acquired from inter-component contact reports generated by the physics engine. Because some components with loose-fit or clearance interfaces may not be in contact in the nominal CAD configuration, direct contact detection may omit mechanically relevant liaisons. After fastener recognition, the recognized bolts and nuts are retained in their fastened states, while the remaining components are allowed to translate with their orientations fixed. The contact graph in the nominal configuration is first recorded. To recover potentially omitted liaisons, a small perturbation acceleration is applied successively along the positive and negative directions of the three Cartesian axes. For each direction, the simulation is advanced for two physics time steps. The nominal contact graph and the contact graphs obtained under the six directional perturbations are accumulated by graph union to construct the initial liaison graph. The perturbation acceleration magnitude and physics time step are reported in Supplementary Table S11.

The resulting liaison graph is annotated using the recognized fastener categories, as shown in Figure 5. Accepted bolt-like and nut-like candidates are represented as orange and purple nodes, respectively, while rejected candidates remain ordinary blue product-component nodes. Their contact edges are retained as structural evidence.

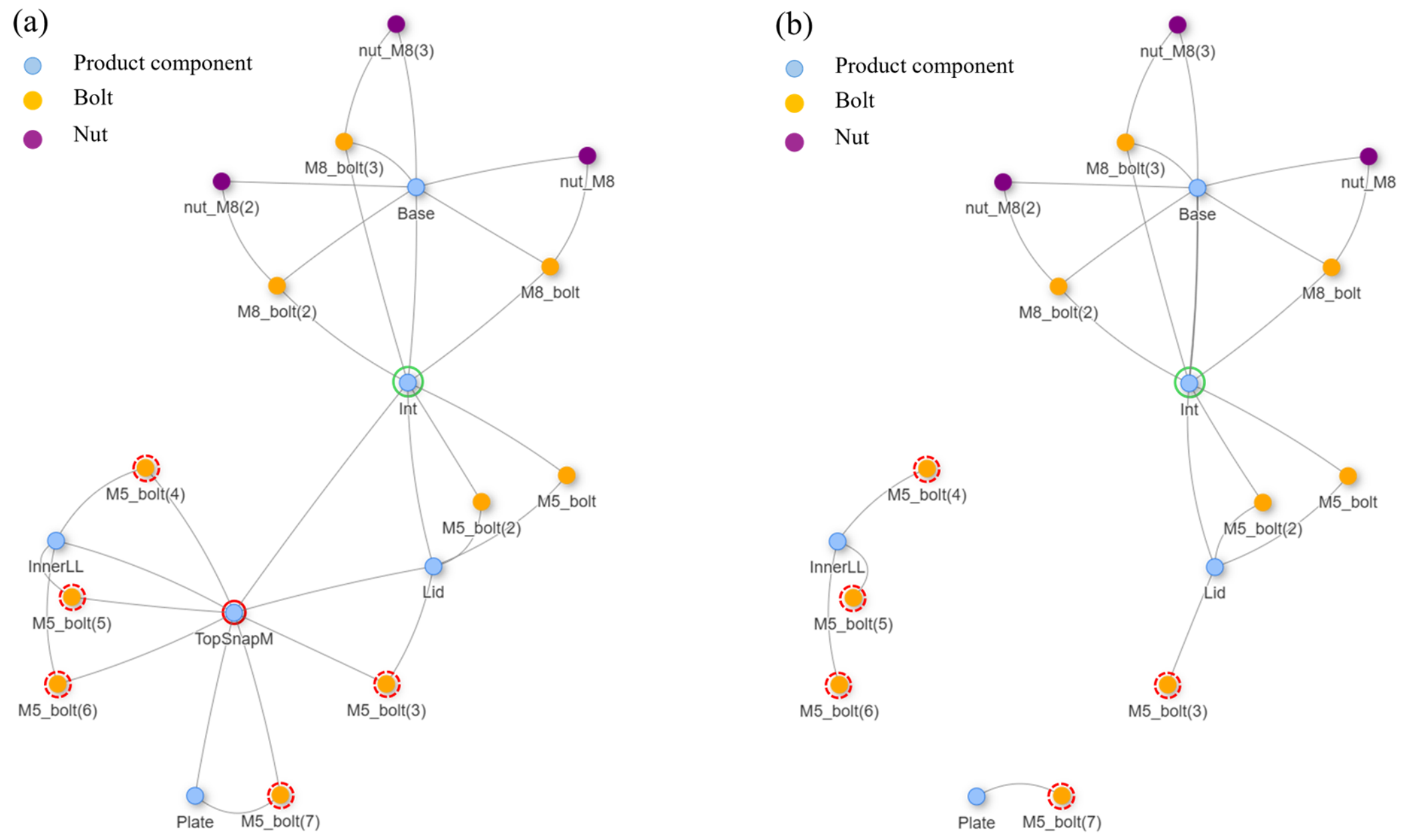

Figure 6. Contact-topology-based grouping of accepted bolt candidates using TopSnapM as the reference component: (a) TopSnapM (red) and its five adjacent bolt candidates (dashed red) and (b) their distribution among the structural regions containing InnerLL, Lid, and Plate after temporary removal of TopSnapM.

### 3.1.2 Fastener-mediated liaison inference

This step identifies contacts constrained by recognized fasteners and associates them with the fastener sets required for their release.

Individual fasteners are not treated as independent operations because an interface may remain constrained until all jointly acting fasteners have been removed. For example, as shown in Figures 2 and 5, the Base–Int interface is fastened by three M8 bolts and three nuts. Releasing this interface requires the corresponding M8 fasteners to be removed as a set. The framework therefore groups jointly acting fasteners into operation-level sets whose complete removal releases the corresponding interface-level constraint.

Fasteners adjacent to the same product component do not necessarily belong to the same removal operation. A component may connect several otherwise separate structural regions, with different fastener groups securing the component to each region. The proposed method identifies these groups by temporarily removing each non-fastener component from the liaison graph and examining how its adjacent fasteners are distributed among the remaining connected components.

Figure 6 illustrates the grouping principle using TopSnapM as the reference component. Five accepted bolt candidates are adjacent to TopSnapM in Figure 6(a) as shown by the dashed red circles:

$$N_F(\text{TopSnapM}) = \{\text{M5_bolt(3)}, \text{M5_bolt(4)}, \text{M5_bolt(5)}, \text{M5_bolt(6)}, \text{M5_bolt(7)}\}.$$

Although all five bolts contact TopSnapM, they do not represent a single fastening operation. Bolts 4, 5, and 6 secure TopSnapM to InnerLL, whereas bolt 3 secures TopSnapM to the region containing Lid, and bolt 7 secures Plate to TopSnapM. After TopSnapM is temporarily removed, the five bolts are distributed among three connected components containing InnerLL, Lid, and Plate, respectively, as shown in Figure 6(b). They are therefore partitioned into the provisional fastener sets

$$S_1 = \{\mathrm{M5_bolt}(4), \mathrm{M5_bolt}(5), \mathrm{M5_bolt}(6)\},$$
$$S_2 = \{\mathrm{M5_bolt}(3)\},$$
$$S_3 = \{\mathrm{M5_bolt}(7)\}.$$

The same procedure is applied to every non-fastener reference component. Let

$$G = (V_C \cup V_F, E) \tag{2}$$

denote the liaison graph, where $V_C$ is the set of non-fastener component nodes, $V_F$ is the set of accepted fastener-candidate nodes, and $E$ is the set of contact edges. For a reference component $r \in V_C$, its adjacent fastener neighborhood is

$$N_F(r) = \{f \in V_F \mid (r, f) \in E\}. \tag{3}$$

The reference component is temporarily removed to form the induced graph

$$G^{-r} = G[(V_C \cup V_F) \setminus \{r\}]. \tag{4}$$

Let

$$\mathcal{C}(G^{-r}) = \{C_1, C_2, \dots, C_k\} \tag{5}$$

denote the node sets of the connected components of $G^{-r}$, where each $C_i$ represents one structural region remaining after removal of $r$. The provisional fastener set associated with $C_i$ is then defined as

$$S_i(r) = N_F(r) \cap C_i,\ S_i(r) \neq \emptyset. \tag{6}$$

Fasteners assigned to different connected components are interpreted as fastening the reference component to different structural regions. If all members of $N_F(r)$ remain in the same connected component, the complete neighborhood is retained as one provisional fastener set. Thus, the number and membership of the provisional sets are inferred from contact topology rather than specified manually or determined solely from fastener geometry. After the provisional bolt sets have been inferred, each recognized nut is paired with the bolt that it contacts and included in the same constraint-release operation.

Each provisional set is subsequently associated with one or more constrained part-to-part interfaces. For each provisional fastener set, the non-fastener components adjacent to its bolt members are examined. If a pair of these components also shares a direct contact edge, that edge is annotated as the fastener-mediated constraint released by complete removal of the set. For example,

$$S_1 = \{\text{M5_bolt(4)}, \text{M5_bolt(5)}, \text{M5_bolt(6)}\}$$

is adjacent to both TopSnapM and InnerLL. Their direct contact edge is therefore associated with $S_1$ as the TopSnapM–InnerLL fastener-mediated constraint.

Applying the procedure to different reference components can produce nested fastener sets. For Lid,

$$N_F(\text{Lid}) = \{\text{M5_bolt}, \text{M5_bolt(2)}, \text{M5_bolt(3)}\}.$$

When Lid is temporarily removed, the three adjacent bolts remain within the same connected region of the remaining graph. The complete neighborhood is therefore retained as the provisional fastener set

$$S_{\text{Lid}} = \{\text{M5_bolt}, \text{M5_bolt(2)}, \text{M5_bolt(3)}\}.$$

In contrast, Int has a larger adjacent fastener neighborhood:

$$N_F(\text{Int}) = \{\text{M8_bolt}, \text{M8_bolt(2)}, \text{M8_bolt(3)}, \text{M5_bolt}, \text{M5_bolt(2)}\}.$$

When Int is temporarily removed, these five bolts are distributed among different connected components of the remaining graph. In the connected component containing Lid, only M5_bolt and M5_bolt(2) remain. Because Lid and Int also share a direct contact edge, these two bolts are associated with the local Lid–Int interface:

$$S_{\text{Lid-Int}} = \{\text{M5_bolt}, \text{M5_bolt(2)}\}.$$

The resulting sets satisfy

$$S_{\text{Lid-Int}} \subset S_{\text{Lid}}. \tag{7}$$

Both sets are retained as independent candidate operations because $S_{\text{Lid}}$ releases Lid from the remaining structure, whereas $S_{\text{Lid-Int}}$ releases the local Lid–Int interface. Thus, nested fastener sets are retained when they correspond to different constraint-release objectives.

Each operation-level fastener set is then partitioned according to the bolt-removal directions acquired in Section 3.1.1. Fasteners associated with the same constraint and having approximately aligned removal directions form a direction group. Each paired nut is assigned to the same direction group as its corresponding bolt. When a set $S$ contains multiple direction groups $S^{(1)}, \ldots, S^{(m)}$,

$$S = \bigcup_{j=1}^{m} S^{(j)}, \tag{8}$$

all groups must be removed before the constraint associated with $S$ is considered released. The possible removal orders of these groups are represented separately because successive workpiece reorientations to facilitate tool access may produce different poses of the resulting released state and, consequently, different subsequent separation paths. The removal order of individual fasteners within each direction group is treated as an internal detail of the constraint-release operation. Individual-fastener permutations are therefore excluded from the present product-level exploration and left to a subsequent lower-level planning stage.

### 3.1.3 Residual-interface probing

Part-to-part constraints are not necessarily mediated by recognized bolts or nuts. Components may also be retained by non-fastener mechanisms, such as snap-fits, latches, or other interlocking geometries. To identify evidence of such constraints, direct part-to-part contact edges that are not associated with an accepted fastener set are treated as residual interfaces and evaluated through local motion probing.

For example, although both Int and TopSnapM are adjacent to recognized bolt candidates, the direct contact between them in Figure 6(a) is not associated with any accepted fastener set that jointly secures the Int–TopSnapM interface. The corresponding edge is therefore evaluated as a residual interface.

A residual interface may represent either a mechanically constraining relation or a passive contact. To distinguish these cases, a local two-body separation test evaluates whether either endpoint unit can move away from the other, which remains fixed. A virtual interference check in the physics simulator evaluates whether the moving unit can be displaced over the probing distance defined below without interfering with the fixed unit.

The probing directions and distance are derived from the surface-informed oriented bounding box of the moving unit. Because assembly and disassembly motions, such as pull-out and guided sliding motions, are often associated with dominant planar interfaces, an OBB based solely on principal component analysis (PCA) of the vertex distribution may not provide operation-relevant directions. The present procedure instead constructs the OBB basis from two dominant, approximately orthogonal surface-normal directions; Appendix B provides the detailed axis-selection procedure. Let $a_1$, $a_2$, and $a_3$ denote the resulting orthonormal axes. The probing-direction set is

$$\boldsymbol{\mathcal{D}}_{\mathbf{OBB}} = \{\pm, a_1, \ \pm a_2 \ \pm a_3\}. \tag{9}$$

These six directions provide practical local separation tests rather than an exhaustive directional search. The probing distance is set to half the length of the longest edge of the surface-informed OBB. If either endpoint unit has at least one overlap-free direction, the residual interface is classified as contact-only. Otherwise, the edge is annotated as an interlock-like constraint.

In addition to these two classifications, some cylindrical contact relations, such as those in pin-connected linkages or link-rod mechanisms, permit relative rotation while remaining structurally connected. For such a relation between components $P$ and $Q$, the framework identifies one or more release-source components $R$ that maintain the relation and records the association

$$R \mapsto (P, Q). \tag{10}$$

The contact-rotatable edge $(P, Q)$ contributes to constraint-based graph connectivity in the same manner as a fastener-mediated constraint edge. When an associated release-source component $R$ is removed, $(P, Q)$ is added to the released-edge set before the updated graph is evaluated. The relation therefore identifies which component releases the cylindrical contact constraint; it does not require the components to be actively rotated during relation acquisition. Appendix D describes the geometric identification of $(P, Q)$ and its association with $R$, while Supplementary Table S11 summarizes the contact-acquisition parameters and common physics settings used throughout the virtual evaluation.

#### 3.1.4 Phase 1 output: Annotated liaison graph

Phase 1 produces an annotated liaison graph containing accepted fastener candidates, operation-level fastener sets, fastener-mediated constraints, interlock-like residual constraints, contact-only edges, and contact-rotatable relations with their release-source associations.

Figure 7 shows the annotated liaison graph for the running example. For explanatory purposes, inferred operation-level fastener sets are enclosed by green dashed outlines. The part-to-part contact edges annotated as fastener-mediated constraints are highlighted in blue, and green arrows link each fastener set to the constraint edge released by its complete removal.

### 3.2 Phase 2: Recursive operation acquisition

In Phase 2, each operation-level fastener set associated with the current assembly state undergoes the two-stage evaluation shown in Figure 1: visibility and tool-access screening, followed by a virtual fastener-removal trial.

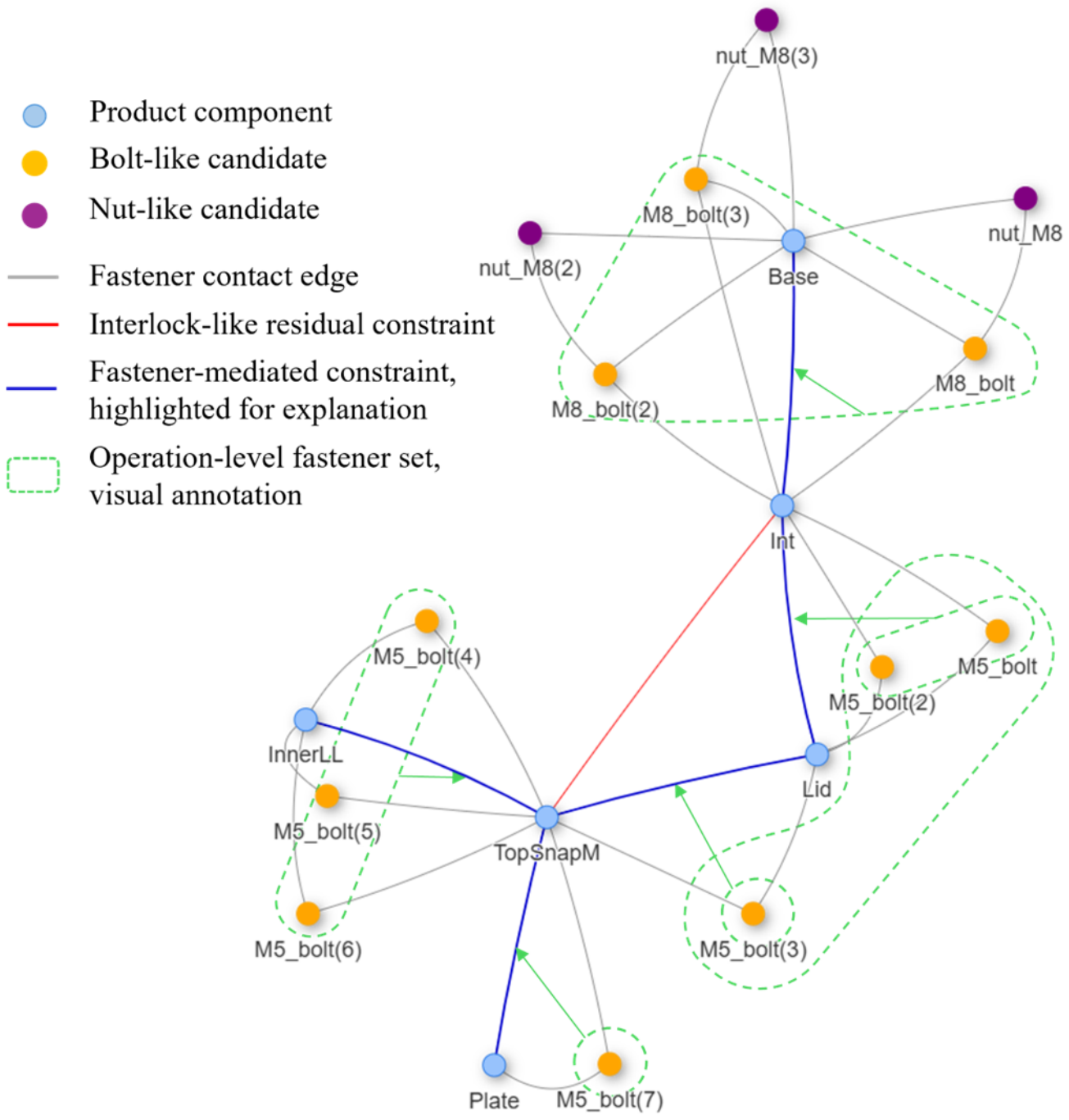

Figure 7. Annotated liaison graph and operation-level fastener sets obtained in Phase 1 for the running example.

### 3.2.1 State-dependent visibility evaluation

The active components in each assembly state are reconstructed in the virtual environment and observed using virtual cameras. Unlike the isolated rendering used for fastener recognition in Section 3.1.1, all active components remain present so that the observations reflect state-dependent occlusion.

The evaluation uses the 12-view icosahedral camera arrangement shown in Figure 3. The camera positions and look-at point were determined by framing the initial full assembly state and were retained throughout subsequent state transitions. Semantic segmentation assigns a component identity to each visible pixel, and a fastener is considered externally visible when its label appears in at least one view. A complete fastener set satisfies the visibility condition only when every required fastener is visible, although its members need not appear in the same view.

### 3.2.2 Tool-access evaluation

A visible fastener set proceeds to state-dependent tool-access evaluation. For each required bolt-like fastener, a cylindrical tool proxy is placed on the head side, with its base positioned 2 mm outward from the center of the head-side OBB face, and aligned with the removal direction acquired in Section 3.1.1, as shown in Figure 8. The proxy diameter is defined as

$$d_{\text{proxy}} = r_{\text{proxy}} l_{\text{diag}}, \tag{11}$$

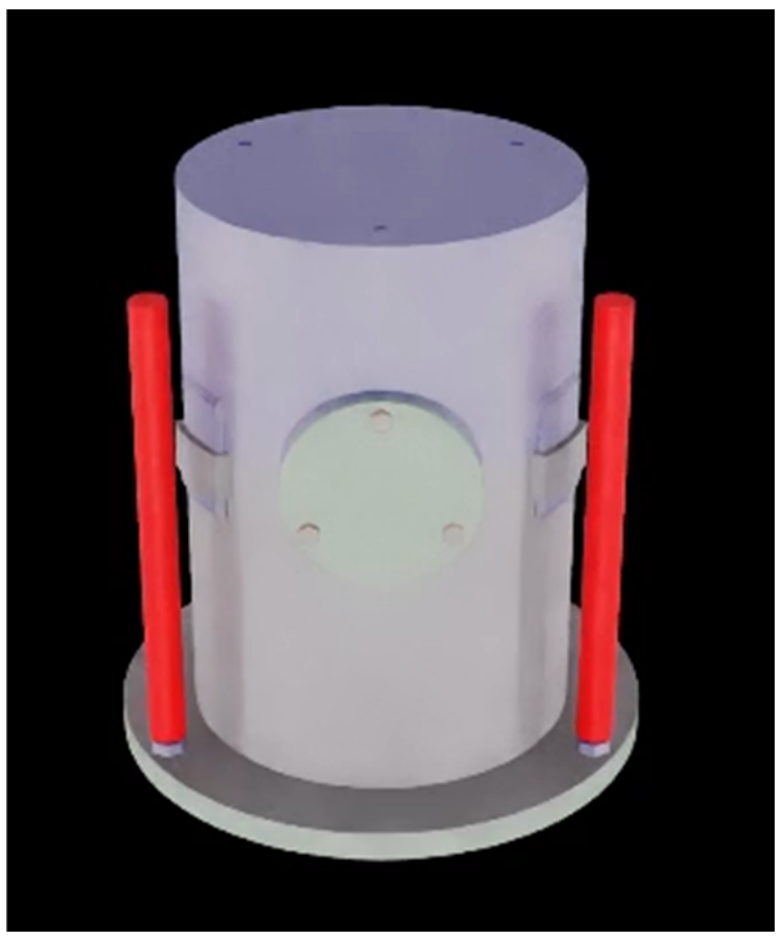

Figure 8. State-dependent tool-access evaluation using cylindrical proxies aligned with the acquired bolt-removal directions. Red cylinders represent the tool-clearance proxies.

where $l_{\mathrm{diag}}$ is the diagonal length of the head-side OBB face as shown in Figure 4 and $r_{\mathrm{proxy}}$ is a configurable scaling ratio. In the reported experiments, $r_{\mathrm{proxy}} = 0.8$, and the proxy length is 200 mm. The proxy diameter therefore scales with the bolt-head dimensions, while both parameters can be adjusted for the assumed tool.

A direction group satisfies the tool-access condition only when every required tool proxy for the group is both visible and collision-free. The proxy represents the required tool-clearance envelope rather than a specific commercial tool and can be replaced by an explicit tool model when tool-specific evaluation is required. The implementation parameters used for visibility and tool-access evaluation are summarized in Supplementary Table S12.

### 3.2.3 Virtual fastener-removal trial

After satisfying the visibility and tool-access conditions, each retained fastener set undergoes a torque-driven removal trial. When a set contains multiple direction groups, the possible group-removal orders are represented separately. Group-level removal results are reused when the corresponding virtual trial conditions are equivalent, rather than repeating the same physics-based removal trial for every permutation. For each group in an evaluated order, the active assembly is rigidly reoriented so that the group's removal direction aligns with a predefined tool-insertion axis in the virtual workspace, such as the positive $z$-axis. This reorientation represents presenting the fasteners in an accessible working orientation for manual or robotic removal. All active components are transformed together to preserve their relative poses and contacts, and the resulting world-coordinate orientation is retained after removal of the group.

For each bolt in the group, a prescribed torque, selected according to the bolt size estimated in Section 3.1.1, is applied about the aligned bolt axis. A small outward axial force is also applied to retain the bolt on the removal side after disengagement and reduce recontact with the assembly. The threaded fastener is simulated under rigid-body contact, friction, and thread-surface interaction.

Removal is accepted when every operated bolt has moved from its initial position by at least its axial length along the estimated removal direction. The trial is terminated if this condition is not reached within the prescribed simulation period.

When multiple direction groups are present, the retained pose after removal of one group becomes the initial pose for removal of the next group in the represented order. The fastener-set removal is accepted only when all groups are successfully removed.

An associated nut is included in the same fastener-removal operation rather than evaluated independently. During bolt removal, the nut is rotationally constrained to represent a structurally fixed nut or one held by an external tool. Nut-side tool access can be evaluated by adding a corresponding tool proxy when required. The fastener-size estimation, virtual-removal parameters, and size-dependent torque values are provided in Supplementary Tables S13 and S14.

**3.2.4 Liaison-graph update after fastener-set removal**

After successful removal of a complete fastener set, its bolt nodes and associated nut nodes are deleted from the virtual scene and liaison graph, together with their incident contact edges. All corresponding fastener-mediated constraint edges identified in Section 3.1.2 are also deleted. For each removed nut, the liaison graph before node deletion is also examined. If two non-fastener components adjacent to the nut share a direct liaison edge, that edge is additionally deleted as part of the same constraint-release operation.

This update represents constraint release, not physical separation. Figure 9(a) shows the M8 fastener set associated with the Base–Int interface and its constraint before removal. After deletion of the fastener nodes and released constraint edge, Figure 9(b) shows Base as one connected component and the remaining active components as the other.

When the updated graph contains multiple connected components, stay-unit and candidate-moving-unit selection proceeds as described in Section 3.2.5. If the updated graph still consists of a single connected component, the fastener removal is recorded as a constraint-release operation without a separation path, and the resulting intermediate state remains available for further exploration.

**3.2.5 Candidate moving-unit selection and initial-direction screening for fastener-mediated releases**

Before full path generation, the framework selects candidate moving units and tests whether each can begin to disengage from the remaining assembly. This screening identifies locally admissible directions for close-fitting pockets, grooves, and sliding guides, for which path generation without an appropriate initial direction may fail.

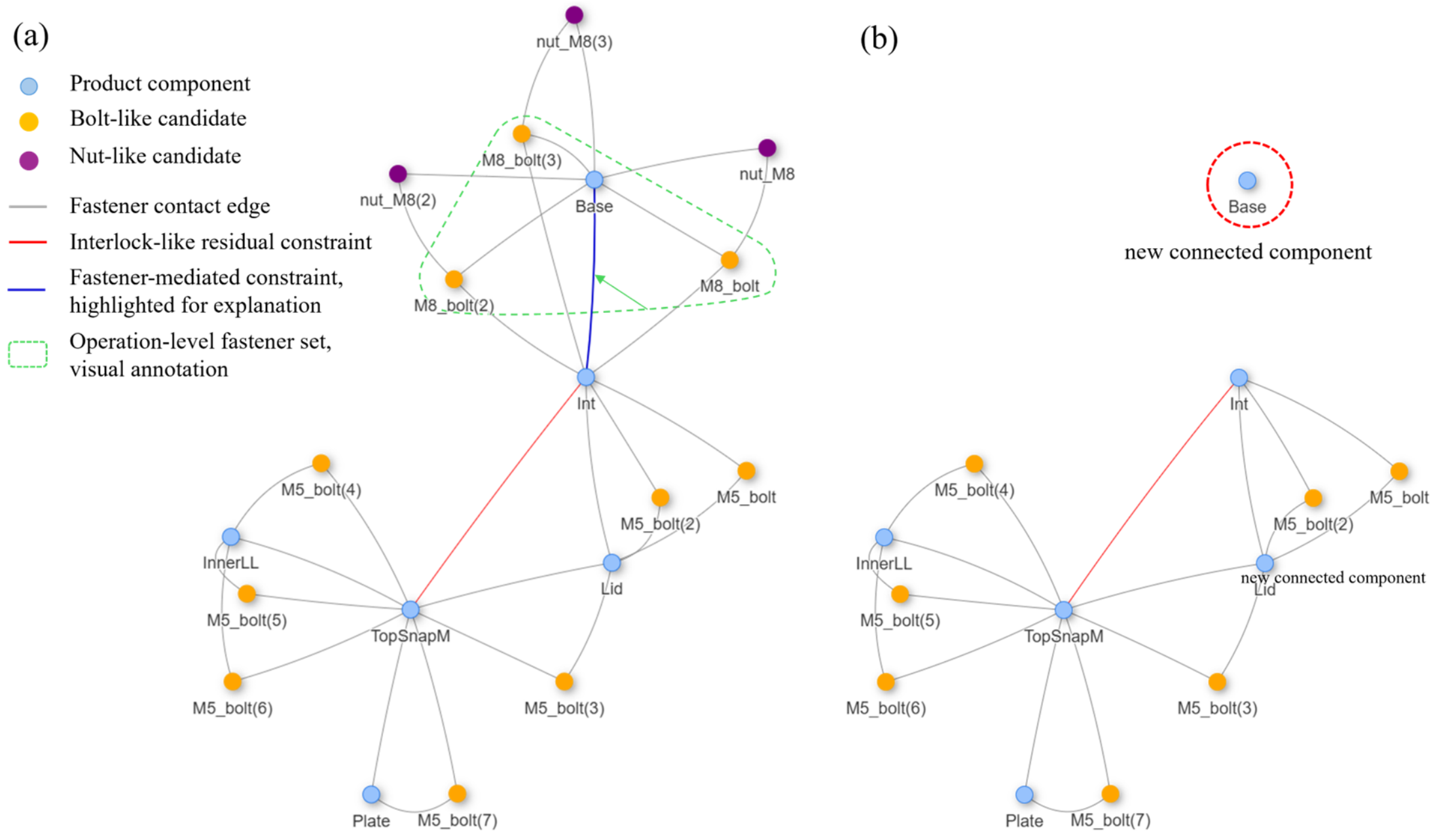

Figure 9. Liaison-graph update after executable removal of the M8 fastener set: (a) selected fastener set and associated Base–Int constraint before removal and (b) resulting connected components after deletion of the fastener nodes and released constraint edge.

When removal of the released constraint edge divides the liaison graph into multiple connected components, one component is selected as the stay unit. If a predefined fixed component is present in the current state, the connected component containing it is selected; otherwise, the component having the largest total mass is used. The other connected components affected by the current release, each comprising either a part or a multi-part subassembly, are evaluated as candidate moving units.

Components associated with contact-only relations require additional consideration because their passive retention may change when an active constraint is released. Such components may remain supported or enclosed by neighboring components, or they may become separable following the constraint release. When an active constraint is released, the corresponding graph update removes contact-only relations associated with the affected components, extending this release through connected contact-only relations to account for the possible loss of contact-based support. Components identified in this way are subsequently screened and validated to determine whether they can actually disengage from the remaining assembly.

For each candidate resulting from a fastener-mediated release, the surface-informed OBB provides the six initial directions defined in Eq. (9). Each direction is evaluated using the swept-clearance test illustrated in Figure 10. For a moving unit $U$, unit direction $\boldsymbol{d}$, and OBB vertices $\boldsymbol{v}_i$, the screening distance is the projected OBB length

$$d_{\text{sweep}}(U, \boldsymbol{d}) = \max_i(\boldsymbol{v}_i \cdot \boldsymbol{d}) - \min_i(\boldsymbol{v}_i \cdot \boldsymbol{d}). \tag{12}$$

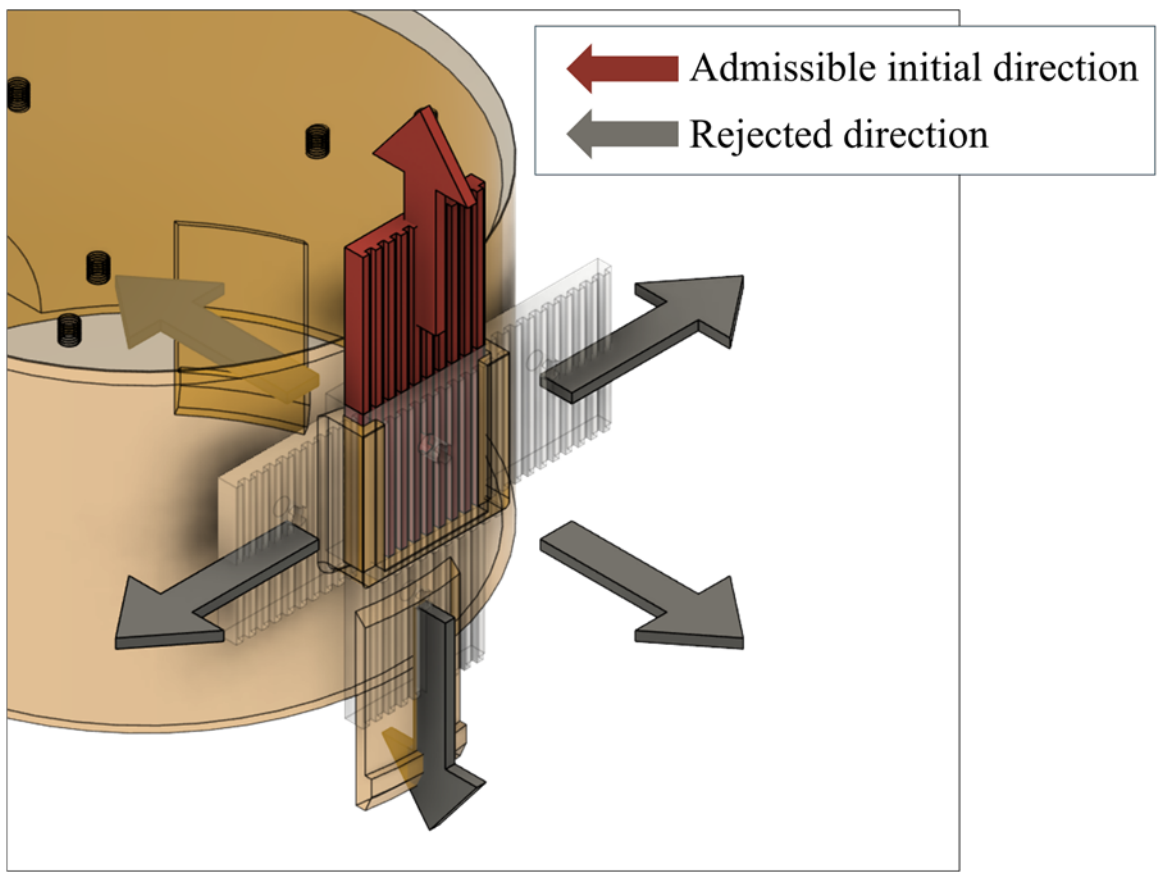

Figure 10. Swept-clearance screening for a guided sliding interface. Five directions derived from the surface-informed OBB are blocked by the surrounding geometry, leaving the red direction as the only admissible initial direction.

A virtual interference check determines whether the moving unit can be displaced over $d_{\text{sweep}}(U, \boldsymbol{d})$ without interfering with the remaining active components. Interference-free directions are retained as locally admissible initial-motion candidates.

In Figure 10, the groove-like structure blocks five OBB directions, leaving only the direction shown in red. When multiple directions remain, they are reduced to a single initial direction. An admissible direction aligned with the released-fastener axis is given the highest priority. If two admissible directions are opposite and collinear, the direction oriented toward the goal position used for subsequent path generation in Section 3.2.7 is selected. Otherwise, the normalized vector sum of the admissible directions is used.

### 3.2.6 Initial-direction extraction for residual interlock-like releases

For a residual interlock-like release, the framework estimates an initial separation direction from contact reactions generated by two opposing loading trials, as illustrated in Figure 11. Interlocking interfaces, including snap-fit-like engagements, contain contact regions that resist local relative motion. Loading these regions produces contact reactions that provide directional evidence for disengagement. Both loading orientations are tested because the orientation producing stronger contact reactions is not known beforehand.

Both endpoint components must satisfy the multi-view visibility criterion in Section 3.2.1. Their interaction is isolated by disabling collisions with the remaining assembly. A residual release proceeds to separation evaluation only when removal of its residual-interface edge divides the current liaison graph into two or more connected components; this graph check follows direction extraction.

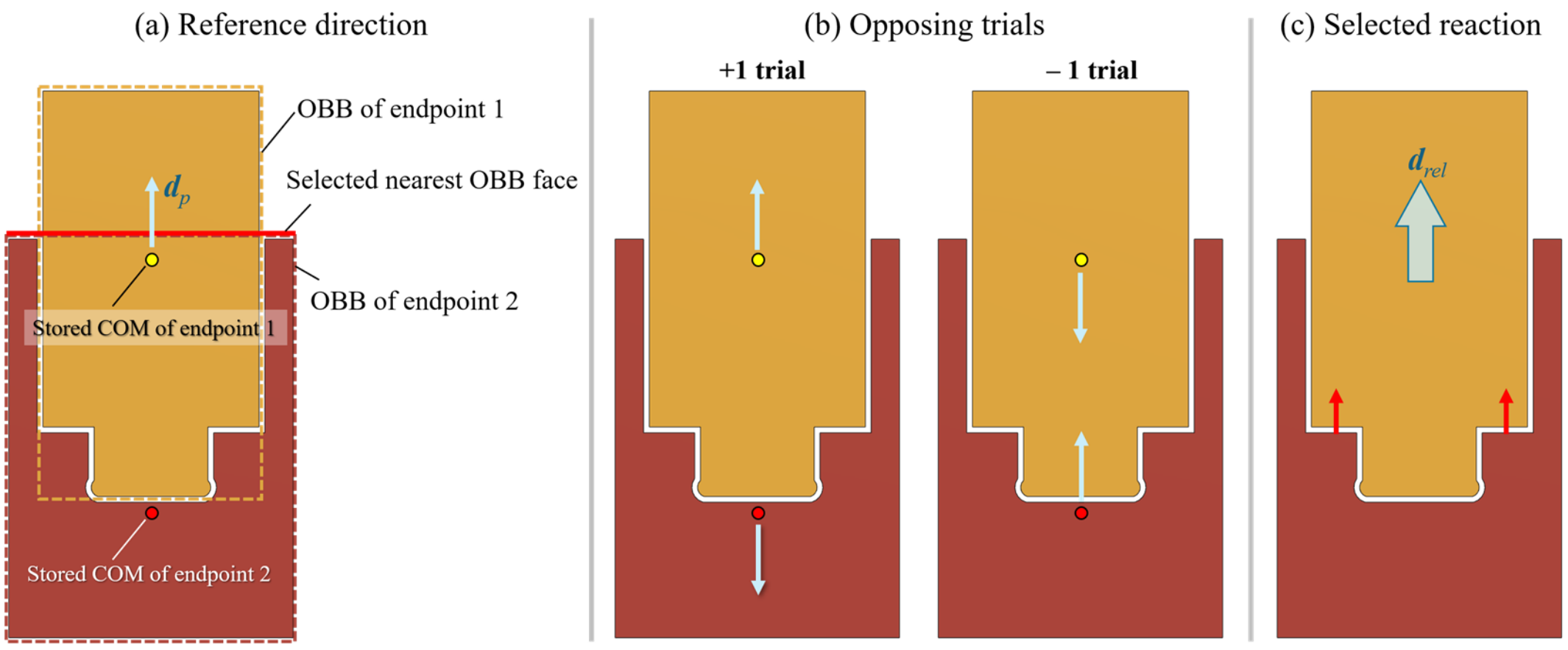

Figure 11. Reaction-based initial-direction estimation for a residual interlock-like release. (a) The stored center-of-mass position of endpoint 1 lies within the surface-informed OBB of endpoint 2. The outward normal of the nearest OBB face defines the reference direction $\boldsymbol{d}_p$. (b) Two trials apply opposing acceleration loads along $\boldsymbol{d}_p$, with the loading signs reversed between trials. (c) When the two candidate directions are nearly opposite and aligned with the line connecting the endpoint OBB centers, their signs are resolved using the endpoint-center direction; otherwise, the trial with the larger accumulated contact contribution is selected. The impulse-weighted contact normals of the selected trial determine the initial separation direction.

To obtain a reference direction that reflects the local containment geometry rather than relying solely on the displacement between the endpoint positions, the reference unit direction, $\boldsymbol{d}_p$, is determined using the endpoint surface-informed OBBs. Each endpoint's stored center-of-mass position is tested for containment within the other endpoint's OBB. For each successful containment test, the face of the containing OBB nearest to the contained position is identified. The outward normal of this face indicates the shortest direction from the contained position to the boundary of the containing OBB. If both containment tests succeed, the two face distances are compared, and the face with the smaller distance is selected. The outward normal of the selected face defines $\boldsymbol{d}_p$ in world coordinates as illustrated in Figure 11(a). If neither endpoint position is contained within the other endpoint's OBB, $\boldsymbol{d}_{\mathrm{p}}$ is instead defined as the normalized vector from the stored center-of-mass position of the first endpoint to that of the second.

The two trials apply opposing acceleration loads of prescribed magnitude $\alpha > 0$ to endpoints 1 and 2, reversing their signs between trials, as illustrated in Figure 11(b):

$$\alpha_1^{(\sigma)} = \sigma \alpha d_p, \qquad \alpha_2^{(\sigma)} = -\alpha_1^{(\sigma)}, \qquad \sigma \in \{+1, -1\}. \tag{13}$$

During each loading trial, the first endpoint is free to translate while its orientation is fixed; the other endpoint remains fixed.

For trial $\sigma$, let $k$ index the reported endpoint-to-endpoint contacts across all sampling steps. Let $\boldsymbol{J}_k^{(\sigma)}$ denote the contact impulse, and let $\boldsymbol{n}_k^{(\sigma)}$ denote the contact normal oriented consistently for endpoint 1. The scalar contribution of contact $k$ is

$$w_k^{(\sigma)} = \left| \boldsymbol{J}_k^{(\sigma)} \cdot \boldsymbol{n}_k^{(\sigma)} \right|.$$

If this normal-impulse magnitude is below the numerical tolerance, the impulse norm is used instead; contributions that remain below the tolerance are discarded. The total contact contribution for the trial is

$$W^{(\sigma)} = \sum_k w_k^{(\sigma)}.$$

The weighted sum of the contact-normal directions is

$$\boldsymbol{r}^{(\sigma)} = \sum_k w_k^{(\sigma)} \boldsymbol{n}_k^{(\sigma)}. \tag{14}$$

Accumulation covers all sampled contacts without separately normalizing each sampling step. When both trials yield nearly opposite candidate directions aligned with the line connecting the endpoint OBB centers, the candidate pointing from the second endpoint's center toward the first endpoint's center is selected to resolve the direction-sign ambiguity. Otherwise, the valid trial with the larger total contact contribution, $W^{(\sigma)}$, is selected. The selected reaction and the resulting reaction-derived direction are illustrated in Figure 11(c):

$$\boldsymbol{d}_{rel} = \frac{\boldsymbol{r}^{(\sigma_*)}}{\|\boldsymbol{r}^{(\sigma_*)}\|}. \tag{15}$$

If the reaction-derived direction is sufficiently aligned with an axis of either endpoint OBB, it is aligned with that axis while preserving its sign. The corresponding direction for endpoint 2 has the opposite sign. Sampling settings are reported in Supplementary Table S16.

Following the reorientation procedure in Section 3.2.3, the active assembly is rigidly rotated so that the reaction-derived direction aligns with the predefined removal axis. The resulting initial direction is then passed to the path-generation and physics-based validation procedures in Section 3.2.7 without applying the swept-clearance screening used for fastener-mediated releases.

### 3.2.7 Path generation and physics-based validation

The moving-unit and initial-direction pairs obtained in Sections 3.2.5 and 3.2.6 are passed to path generation and physics-based validation. A moving unit may be a single part or a multi-part subassembly, while the other active components form the surrounding assembly.

Each pair is evaluated in its corresponding released trial state. For a fastener-mediated release, the validated bolts and associated nuts have been removed, and the corresponding constraint edge has been deleted from the liaison graph. For a residual interlock-like release, the selected residual edge is treated as temporarily released without removing any component.

Because elastic deformation and latch disengagement are not modeled explicitly, the release mechanism of the selected residual interface is not simulated directly. Instead, only the stay-unit endpoint component of the selected residual edge is excluded from collision checking during the corresponding separation trial. Collision checking remains enabled for every other component of the stay unit and surrounding assembly. Internal collision relationships within the moving unit are also preserved.

**FIM-based candidate-path generation**

The Fast Iterative Method (FIM) is used for deterministic, GPU-compatible candidate-path generation during recursive state exploration [42]. The moving-unit centroid serves as the representative point, and a predefined goal position in the virtual workspace defines the endpoint.

The candidate path begins with a straight displacement along the initial direction obtained in Section 3.2.5 or 3.2.6 over the projected OBB length $d_{\text{sweep}}(U, \boldsymbol{d})$ defined in Eq. (12). FIM then generates the remaining path from the displaced position to the goal. Direct discretization at a coarse resolution may eliminate narrow but geometrically feasible passages and consequently prevent path generation. The workspace is therefore initially discretized using a fine occupancy grid with a spacing of 1 mm to preserve narrow free-space regions and obstacle-boundary geometry. This spacing is increased by factors of two only when necessary to keep the occupancy grid below 300 million cells. The FIM grid initially adopts the resulting occupancy-grid spacing and is further coarsened by factors of two until its size does not exceed 20 million cells. Appendix C provides further details of the FIM-based path-generation procedure.

The resulting centroid path is only an object-level reference and does not establish an executable separation motion. Executability is evaluated using the complete moving-unit geometry and physical interactions through ghost-guided validation.

**Ghost-guided physics validation**

Final acceptance of a candidate path is determined through ghost-guided physics validation. The FIM path is resampled at spatial intervals approximately equal to the FIM grid spacing $h$, and a collision-disabled kinematic ghost representing the moving unit advances sequentially through the resulting path points. The moving unit remains a dynamic rigid body and retains collision interactions with the surrounding active components, which are held fixed during validation, as illustrated in Figure 12. For a residual-interface release, only the endpoint component that does not belong to the moving unit is excluded from collision checking against the moving unit; collision interactions with all other active components remain enabled.

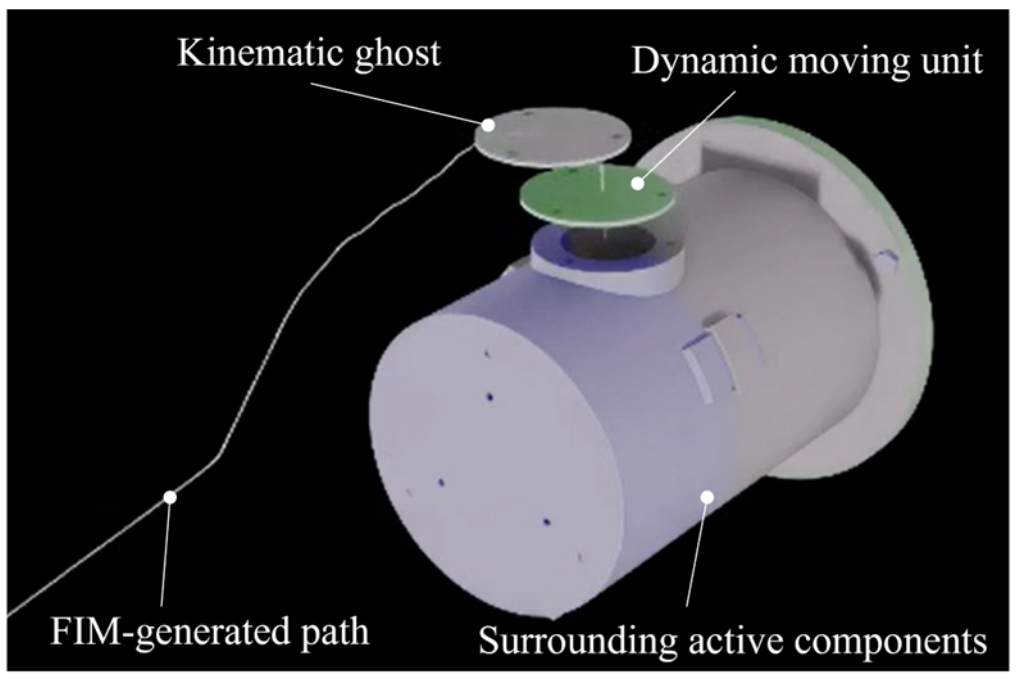

Figure 12. Ghost-guided physics validation along an FIM-generated reference path.

The ghost follows the positional path while retaining the initial moving-unit orientation. A velocity-based proportional-derivative controller guides the dynamic moving unit toward each ghost position. Linear and angular velocity commands are bounded, while exact orientation tracking is not required, allowing contact-induced translation and rotation of the moving unit. The controller gains, velocity limits, and per-point tracking limits are summarized in Supplementary Table S15.

During validation, friction between the moving unit and the surrounding assembly is temporarily set to zero. In close-fitting guides, small contact-induced orientation deviations can cause frictional wedging and prevent path following. Friction is therefore disabled to isolate geometric separability from friction-dependent locking behavior. Accordingly, the complete moving-unit geometry must follow the reference path without violating any collision interaction that remains enabled.

At each path point, the moving unit is considered to have reached the ghost when the positional tracking error satisfies

$$d_{\text{track}} \leq h. \tag{16}$$

If the moving unit does not satisfy this criterion within the configured simulation-step limit, the separation trial is rejected. Otherwise, the ghost advances to the next path point. A separation transition is accepted when the moving unit successfully tracks the complete reference path. The resulting state, reference path, and simulation-derived operation time are then recorded.

### 3.3 Knowledge output and T-AOG construction

Successful operations and their resulting states are organized into a Transformed AND/OR Graph (T-AOG), following the representation introduced by Koc et al. [25]. Unlike a conventional AND/OR graph, a T-AOG introduces disassembly-task nodes, referred to here as operation nodes, to explicitly represent the operations producing subsequent part or subassembly states.

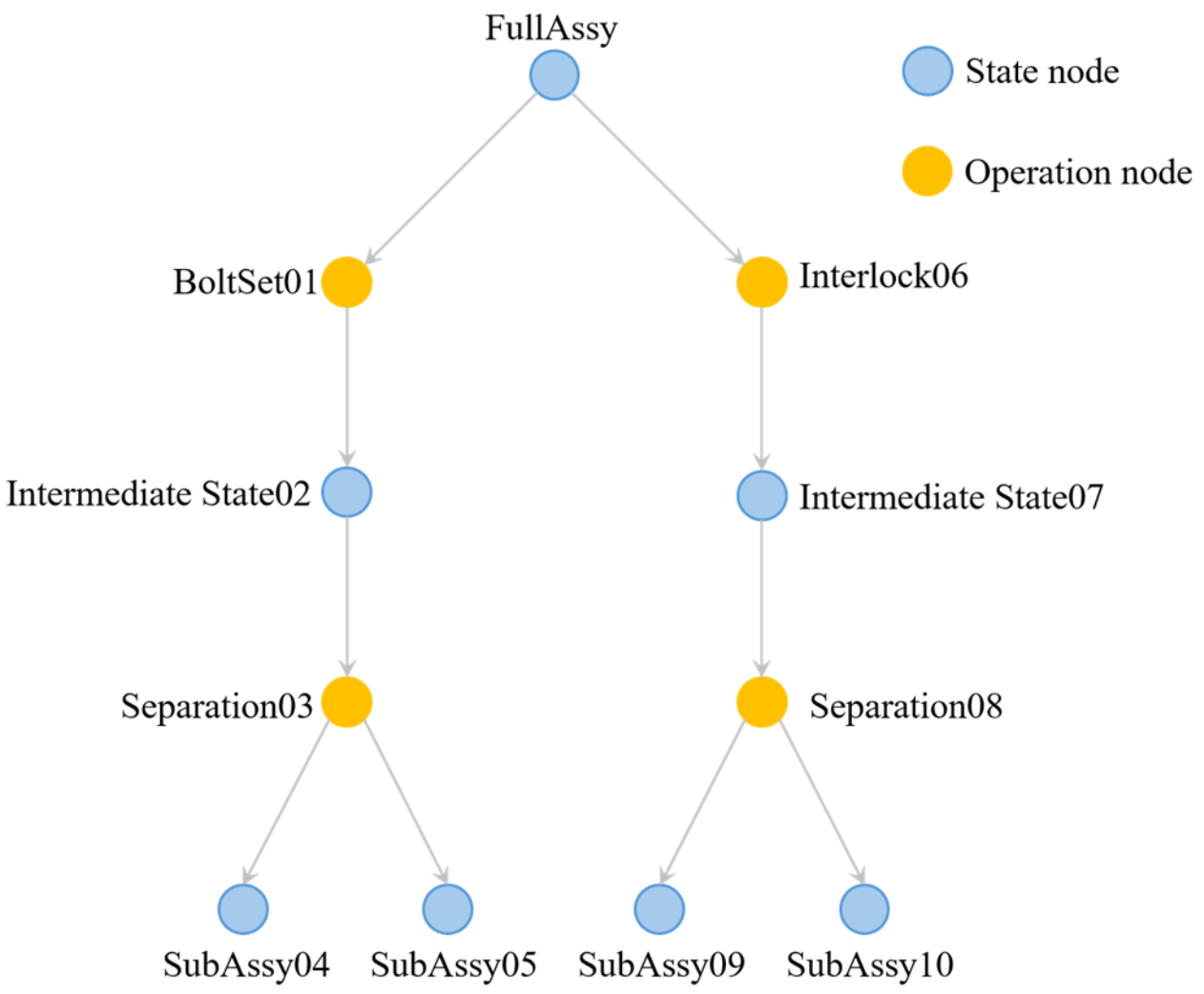

Figure 13. T-AOG representation of operation-level virtual disassembly knowledge. Blue state nodes and orange operation nodes function semantically as OR and AND nodes, respectively, including when they have only one outgoing edge.

Figure 13 illustrates the resulting T-AOG representation, in which state and operation nodes function semantically as OR and AND nodes, respectively. A state node represents an assembly, subassembly, or separated part in a specific configuration. Alternative operations executable from that state form outgoing branches. An operation node represents either constraint release or physical separation and connects its source state to the state or states produced by the operation.

The T-AOG preserves the state-dependent dependencies acquired through the recursive exploration in Algorithm 1. In particular, fastener removal and physical separation are represented as distinct operations with a released intermediate state between them.

For operation-level fastener sets partitioned into multiple direction groups, each feasible group-removal order is recorded as a distinct T-AOG route. For example, let $A$ and $B$ denote two groups of fasteners having different removal directions. The orders $A \rightarrow B$ and $B \rightarrow A$ are retained separately. Each route retains the workpiece pose resulting from its group-removal order for subsequent separation evaluation.

Following a fastener-set removal, validated separations of different moving units are recorded as alternative outgoing operations from the released state. After each separation, the remaining moving-unit candidates are reevaluated under the updated obstacle configuration. Feasible moving-unit separation orders are thereby represented as alternative T-AOG routes, with equivalent resulting states reused.

Table 1. Principal node attributes recorded in the T-AOG.

| Category | Node type | Stored attributes |
|---|---|---|
| State | Assembly/Subassembly/Part state | component part set, subassembly mass, pose |
| Operation | Fastener-set removal operation, including associated nuts | fastener type and size, number of fasteners, number of cameras from which each fastener is visible, removal direction, fastener total mass |
| Operation | Interlock-interface handling operation | residual interface type, detected visibility evidence, initial release direction |
| Operation | Separation transition | validated separation direction |

Table 2. Principal edge attributes recorded in the T-AOG.

| Edge type | Meaning | Stored attributes (when applicable) |
|---|---|---|
| State → Operation | Operation selected and executed from the source state | simulation-derived fastener-removal time, removed fastener mass |
| Operation → State | State produced by the operation | simulation-derived moving-unit separation time, transported mass, validated path trajectory, moving unit, and resulting component membership |

Child subassembly states produced by validated separation are compared with the state nodes already registered in the T-AOG. Single-part states containing the same component are merged without pose comparison because they represent terminal branch states. For multi-part states associated with validated separation paths, equivalence requires the same component set, an orientation difference of less than $5°$, and a dot product greater than 0.98 between the normalized initial separation directions. An equivalent existing node is reused; otherwise, a new state node is registered and added to the exploration queue. Released intermediate states are retained separately because they encode the parent-specific constraint-release history and must not be merged across different parent states.

**State- and operation-node attributes**

Table 1 summarizes the principal node attributes. State nodes record component part set, subassembly mass, and pose. Operation nodes store the fastener, interface, or moving-unit information required to define the corresponding operation.

**Transition-edge attributes**

Table 2 summarizes the principal attributes stored on state-to-operation and operation-to-state edges. State-to-operation edges specify operations executable from a source state and store the source-state pose used for the corresponding operation trial. Operation-to-state edges represent the resulting states and store transition-specific outcomes, including the validated separation path, simulation-derived moving-unit separation time, transported mass, and resulting component membership.

Table 3. Computational environment.

| Item | Setting |
|---|---|
| Cloud platform | AWS |
| Instance type | g6e.2xlarge |
| GPU | NVIDIA L40S Tensor Core GPU, 48 GB memory |
| CPU | AMD EPYC 7R13, 8 vCPUs |
| System memory | 64 GB |
| OS | Ubuntu 20.04 |
| Simulator | NVIDIA Isaac Sim 5.1.0 |
| Programming environment | Python 3.11 |

## 4. Results

The proposed framework was evaluated using the controlled toy assembly introduced in Section 3 and an OpenManipulator Chain gripper. The former served as the running example for examining fastener-set inference and operation-state transitions, whereas the latter provided a denser electromechanical test assembly. Because many publicly available assembly datasets are restricted to non-commercial research, the second test assembly was selected from a model distributed under a license permitting the present use.

### 4.1 Experimental setup, test assemblies, and Phase 1 verification

The experiments were conducted in the physics-based virtual-disassembly environment implemented using NVIDIA Isaac Sim 5.1.0 [43]. Isaac Sim was selected because its built-in rigid-body contact capabilities permit simulation of interactions between threaded geometries, whose feasibility and computational scalability within NVIDIA's robotics-simulation ecosystem had previously been demonstrated in Factory [36]. The framework was executed in Python 3.11 on an AWS g6e.2xlarge instance. Table 3 summarizes the computational environment.

The OpenManipulator Chain gripper shown in Figure 14 was reconstructed from component models distributed through Thingiverse under the Creative Commons Attribution 4.0 International license [44]. The evaluation was limited to the gripper; the complete manipulator shown in the inset was reconstructed only for visual context and was not included in the evaluated virtual assembly. The RUBBER_PAD components were excluded because they are attached with adhesive tape and are outside the scope of the rigid-body physical model, as discussed in Section 5.3.1. The gripper contains repeated left-right mechanisms, co-fastened components, link structures, and close-fitting sliding guides, providing a denser electromechanical test assembly for evaluating fastener-set inference, tool accessibility, constraint release, and subsequent physical separation.

Table 4 summarizes the composition, dimensions, and evaluation characteristics of the two test assemblies. The OpenManipulator Chain gripper comprised 18 non-fastener components, 24 bolts, and 12 nuts, giving 54 components in total.

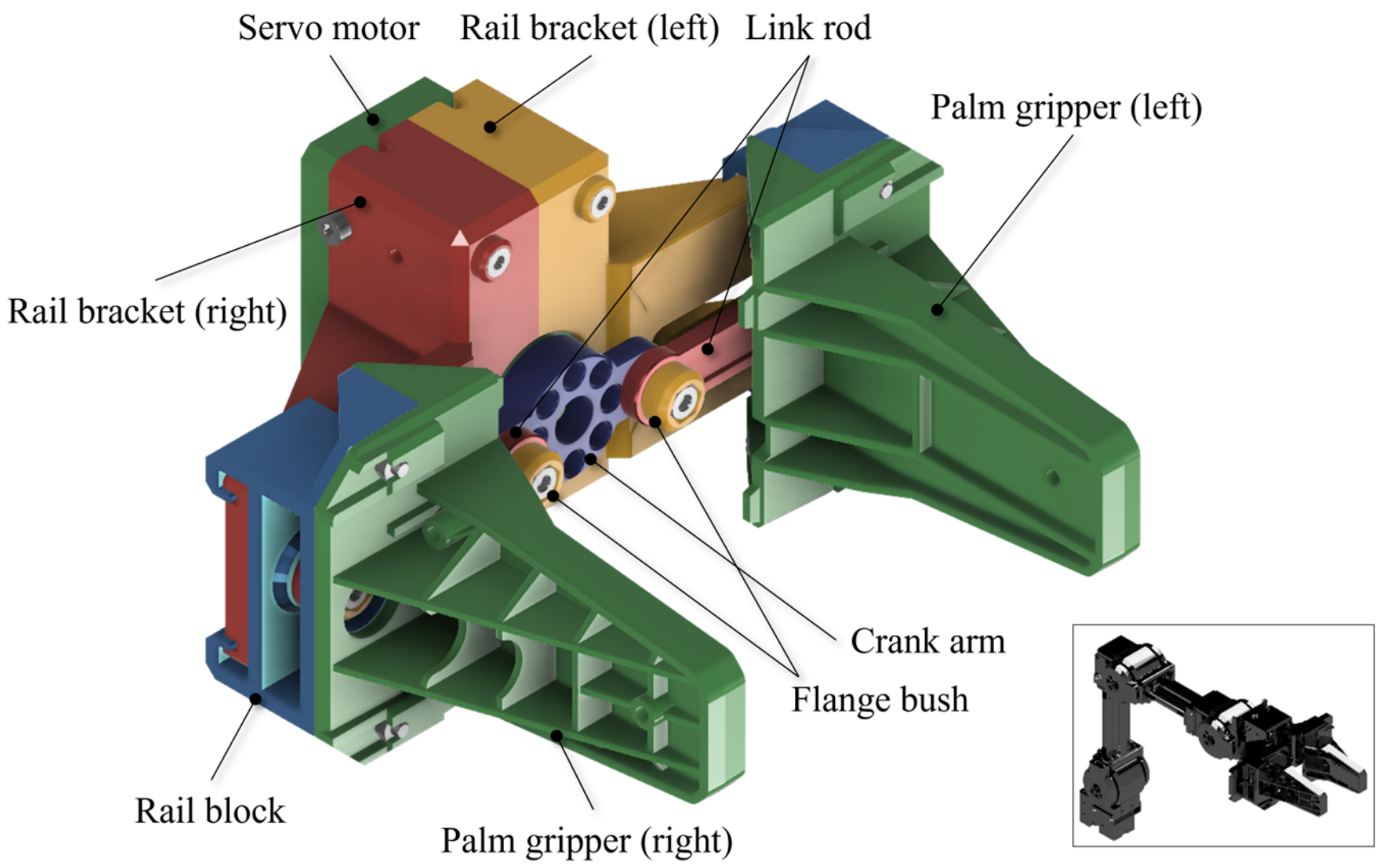

Figure 14. OpenManipulator Chain gripper assembly evaluated in the virtual-disassembly experiment. The inset shows the complete manipulator for context; only the gripper was included in the evaluated assembly. OpenManipulator Chain CAD models provided by ROBOTIS under CC BY 4.0.

Table 4. Characteristics and evaluation roles of the two test assemblies.

| Assembly | Components | Bolts/Nuts/ Non-fasteners | Dimensions, $L_x \times L_y \times L_z$ (mm) | Characteristics |
|---|---|---|---|---|
| Toy assembly | 19 | 10/3/6 | 220×220×256 | fastener-mediated interfaces, interlock-like residual interfaces, sliding part |
| OpenManipulator Chain gripper | 54 | 24/12/18 | 138×69×119 | sliding parts, contact-only relations, co-fastened parts, link rods |

**The toy assembly contained three M8 bolts, three M8 nuts, and seven M5 bolts. The OpenManipulator Chain gripper contained four M3 bolts, four M3 nuts, sixteen M2.5 bolts, eight M2.5 nuts, and four M2 bolts.*

Table 5. Processing time and GPU memory usage.

| Assembly | Processing time | Approximate observed GPU memory usage |
|---|---|---|
| Toy assembly | Approx. 60 min | Approx. 20 GB |
| OpenManipulator Chain gripper | Approx. 2.5 h | Approx. 21 GB |

The OpenManipulator Chain virtual assembly was reconstructed from the available STL component models. Local geometric artifacts introduced by the STL representation were corrected to restore the intended geometry. In particular, faceted holes were reconstructed as cylindrical surfaces, and small fillets were locally introduced where STL-derived geometry caused unintended interference. No fastener-interface associations, precedence relations, operation sequences, or removal paths were specified manually. The servo motor was designated as the fixed component. In states containing the servo motor, its connected component formed the stay unit after each liaison-graph division; otherwise, the largest-mass component was selected. Table 5 reports the end-to-end processing time from initial evidence acquisition through T-AOG generation and the approximate peak GPU-memory usage for each assembly.

Table 6. Recognition results for the evaluated assemblies after class-attribute consistency checking.

| Assembly | Bolts accepted | Nuts accepted | Non-fasteners rejected | False fastener acceptance |
|---|---|---|---|---|
| Controlled toy assembly | 10/10 | 3/3 | 6/6 | 0 |
| OpenManipulator Chain gripper | 24/24 | 12/12 | 18/18 | 0 |

The Phase 1 outputs were compared with the known component identities and represented assembly structures of the two test assemblies. Table 6 summarizes the final fastener-recognition results after class-attribute consistency checking. All 10 bolts and 3 nuts in the controlled toy assembly and all 24 bolts and 12 nuts in the OpenManipulator Chain gripper were retained as fastener candidates. Some non-fastener components initially received fastener-class predictions, but these predictions were rejected because their geometric attributes were inconsistent with the predicted classes. Consequently, no non-fastener component was retained as fastener evidence in either assembly. Component-family results and representative rejected predictions are provided in Supplementary Tables S1–S3.

The accepted fasteners were subsequently grouped into operation-level sets, and the inferred memberships and interface associations were compared with the represented assembly structures. The toy assembly produced six candidate sets, including nested sets with distinct release objectives. The OpenManipulator Chain gripper produced local sets corresponding to the palm-gripper, servo-connected, and link-mechanism interfaces, together with a nested twelve-bolt set spanning the three servo-connected interfaces. No incorrect set membership or interface association was identified. Set memberships, associated nuts, interface associations, and nested-set relations are provided in Supplementary Tables S4 and S5.

The acquired residual and contact-rotatable relations were likewise compared with the represented structures. In the toy assembly, the framework identified the TopSnapM-Int contact as an interlock-like residual constraint, as illustrated in Figure 7. In the OpenManipulator Chain gripper, it identified contact-rotatable relations between the left- and right-side link rods and flange bushes and associated them with the corresponding bolt-based release sources. The annotated liaison graph and structural roles of these relations are presented in Section 4.3. Detailed records are provided in Supplementary Tables S6 and S7.

## 4.2 Results on the toy assembly

### 4.2.1 Exhaustive verification of T-AOG branching

The toy assembly was designed as a controlled model to facilitate explanation and understanding of the framework processes and to enable verification of the soundness of the T-AOG branching generated by the framework. For this purpose, the assembly contains the following two deliberately introduced conditions. The verification separately examined a tool-accessibility condition and a precedence condition for residual-interface release.

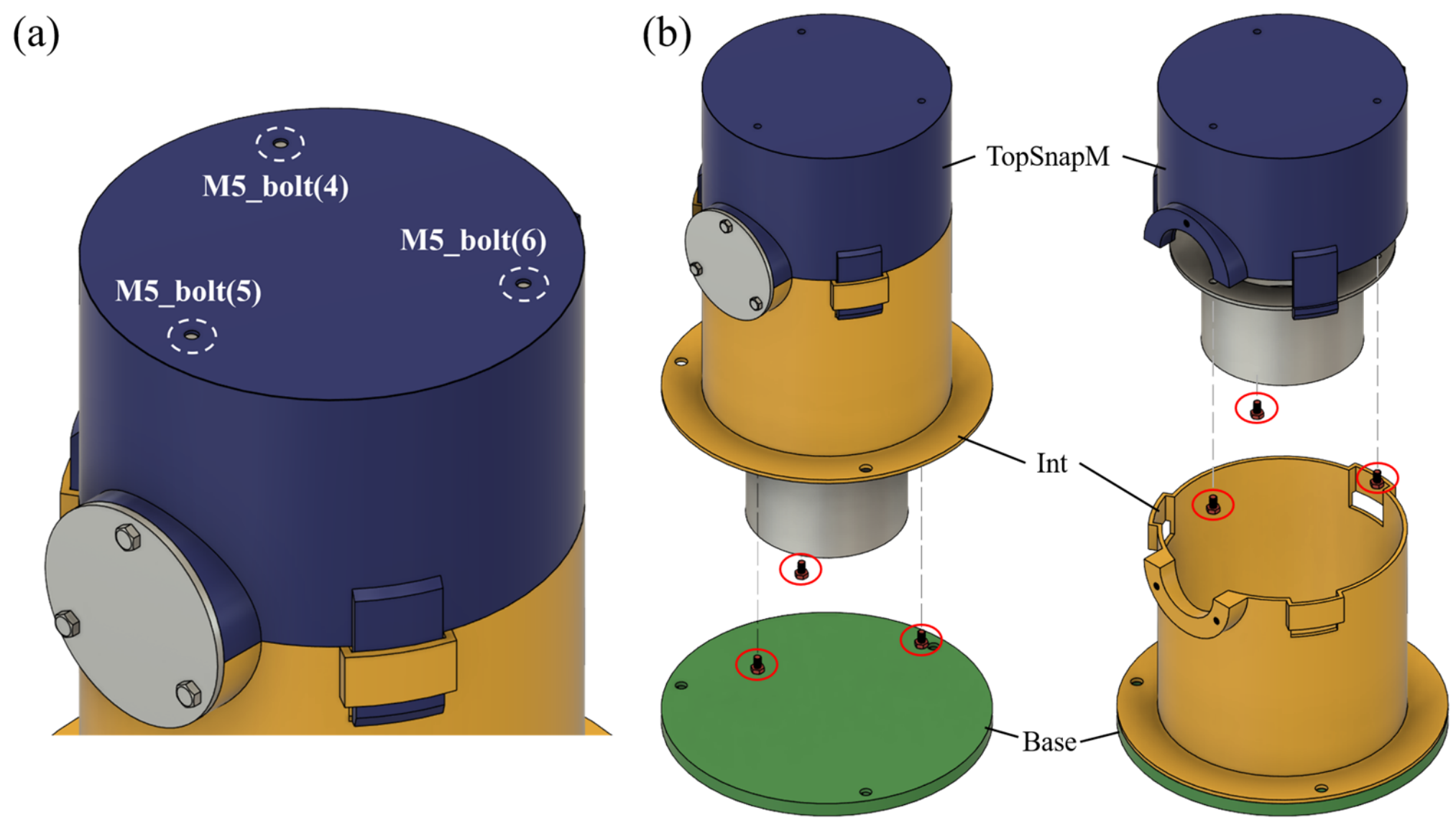

Figure 15. Visibility and tool accessibility of the three-bolt set in the controlled toy assembly. (a) External visibility of the threaded tips of M5_bolt(4), M5_bolt(5), and M5_bolt(6) through the corresponding through-holes. (b) Bolt-head-side tool access enabled by release of either the Int-Base contact or the TopSnapM-Int contact.

The three-bolt set comprising M5_bolt(4), M5_bolt(5), and M5_bolt(6) was used to verify that external visibility and tool-proxy accessibility were distinguished during operation generation. As shown in Figure 15(a), the threaded tips of the three bolts remain externally visible through the corresponding through-holes. However, their bolt heads cannot be accessed until either the Int-Base contact or the TopSnapM-Int contact has been released, as shown in Figure 15(b). The framework first identifies the three-bolt set as externally visible and then evaluates whether the bolt heads are accessible to the corresponding tool proxies. The runtime tool-accessibility results were compared with the logically expected accessibility of the three-bolt set at every applicable T-AOG state node.

A separate precedence condition was introduced for release of the TopSnapM-Int contact. As shown in Figure 16, this contact can be released after removal of the two-bolt set comprising M5_bolt and M5_bolt(2) in Figure 16(a), the single-bolt set comprising M5_bolt(3) in Figure 16(b), or the combined three-bolt set in Figure 16(c). Removal of the combined set satisfies both prerequisite conditions because it includes the fasteners in the two smaller sets. The generated T-AOG was examined to confirm that no TopSnapM-Int release branch appeared before at least one prerequisite had been satisfied and that the branch appeared whenever the required components were present and at least one prerequisite had been satisfied.

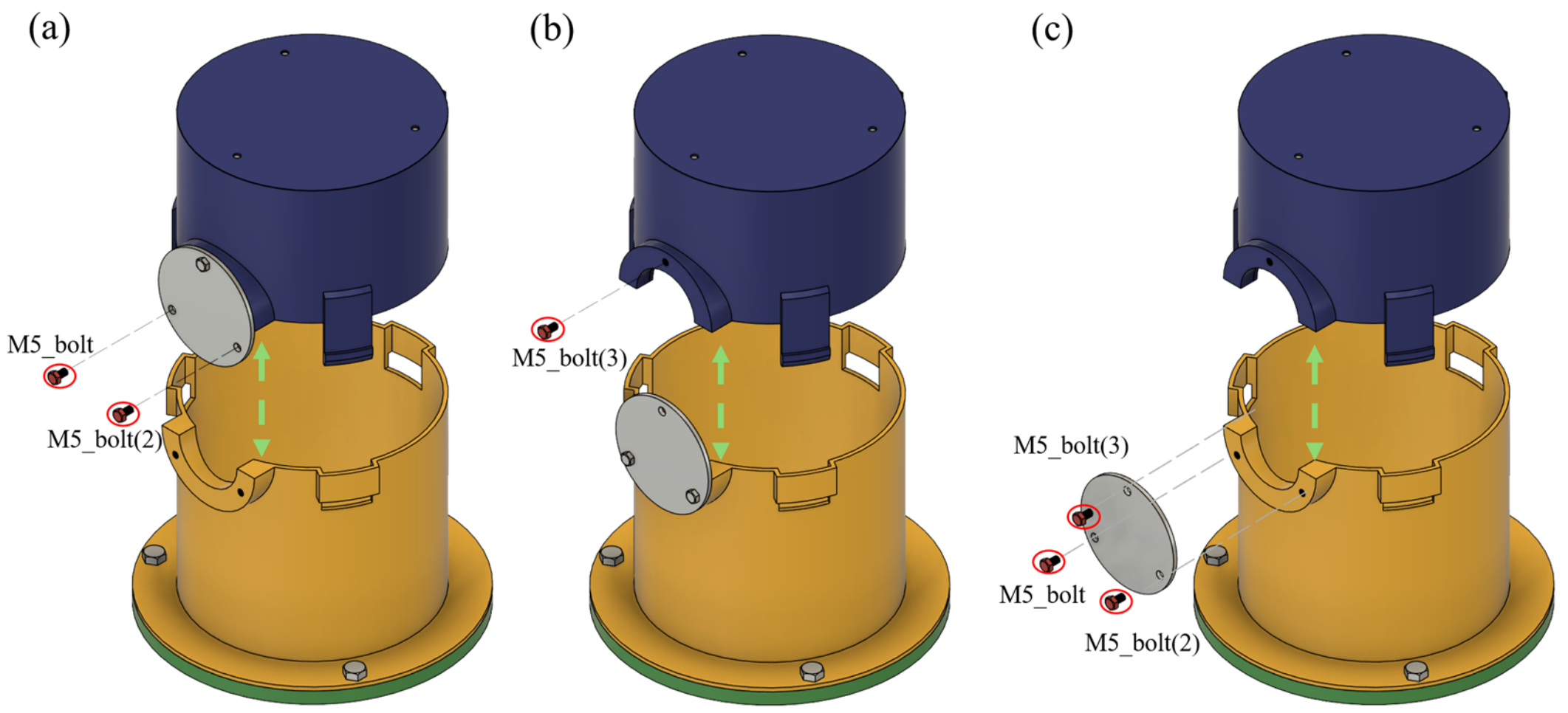

Figure 16. Fastener sets associated with release of the TopSnapM-Int contact in the controlled toy assembly: (a) the two-bolt subset comprising M5_bolt and M5_bolt(2), (b) the single-bolt subset comprising M5_bolt(3), and (c) their combined three-bolt set.

Figure 17 shows the full T-AOG generated for the controlled toy assembly. Although the graph is too dense for detailed node-level inspection, it provides an overview of the branching structure examined in this verification. Each applicable state node and its outgoing operation nodes were independently checked against the initial liaison graph, the contacts released through the preceding operation history, and the runtime tool-proxy accessibility results. The expected branches derived from these conditions were then compared with those recorded in the generated T-AOG.

As shown in Table 7, the target of the operation for removing M5_bolt(4), M5_bolt(5), and M5_bolt(6) was present in 37 of the 90 evaluated state nodes. The operation was generated in all 26 states in which bolt-head-side tool access was available and correctly suppressed in all 11 states in which the threaded tips remained visible but complete tool access was unavailable. The target of the TopSnapM–Int release operation was present in 47 of the 90 evaluated state nodes. Its branch was generated in all 39 states in which either prerequisite fastener set had been removed and correctly suppressed in all eight states in which neither prerequisite had been satisfied. No violations were detected. The state-level verification details are provided in Supplementary Table S8, while the complete state-history-level audit across all seven operation identities is reported in Supplementary Table S9.

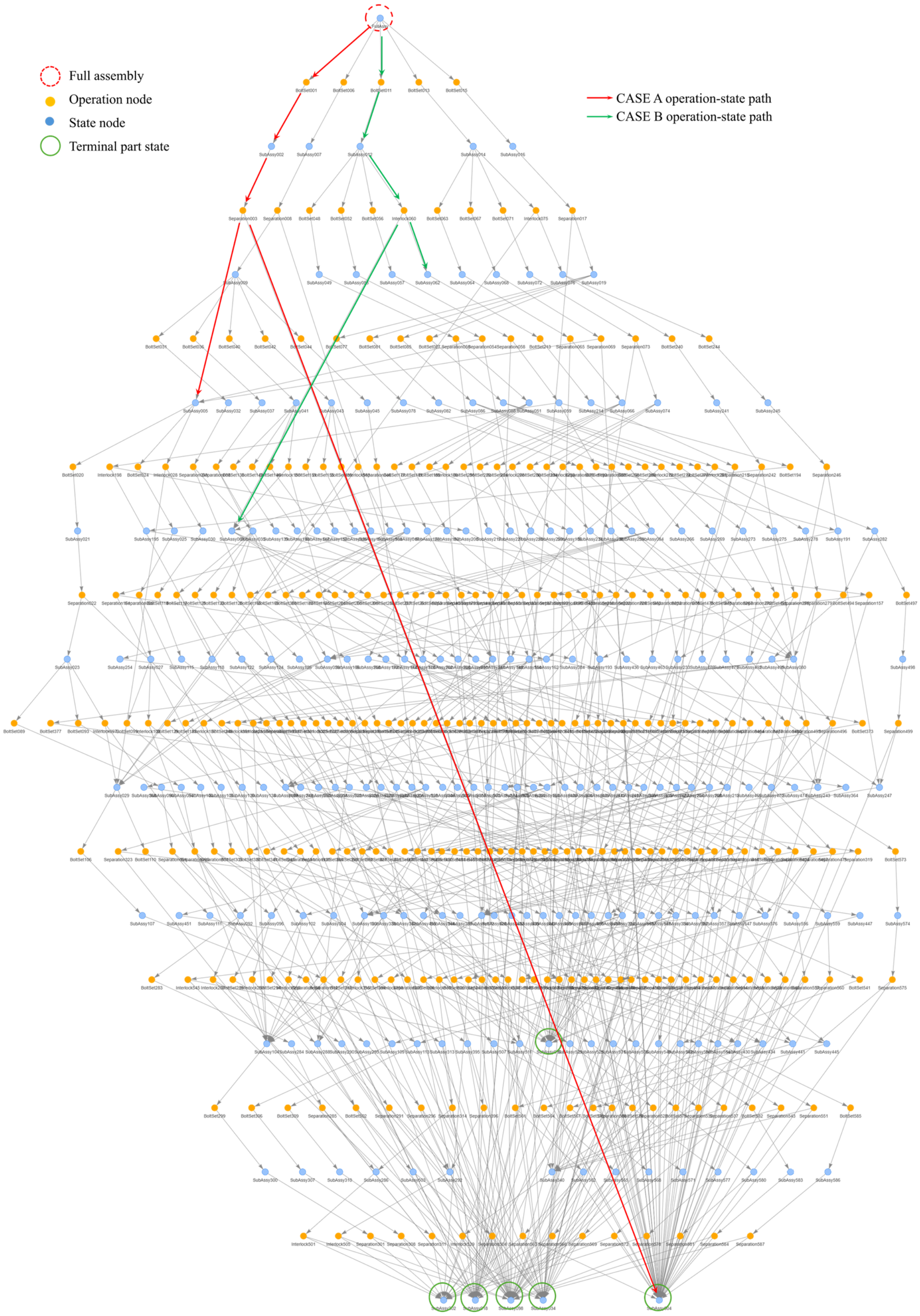

Figure 17. T-AOG generated for the toy assembly under the configured exploration procedure. The red and green paths correspond to Cases A and B, whose state transitions are shown in Figure 18.

Table 7. Verification of controlled branching conditions in the toy-model T-AOG.

| Controlled operation | Evaluated state nodes | Target-present state nodes | Correctly generated | Correctly suppressed | Violations |
|---|---|---|---|---|---|
| Three-bolt removal based on bolt-head-side tool accessibility, Figure 15 | 90 | 37 | 26 | 11 | 0 |
| TopSnapM-Int release based on preceding fastener-set removal, Figure 16 | 90 | 47 | 39 | 8 | 0 |

*Evaluated state nodes are the reachable state nodes whose reconstructed liaison graph has a single connected component, regardless of whether the target of the corresponding operation is present. Each state node is counted once, irrespective of the number of operation histories reaching it.*

### 4.2.2 Representative physics-based outcomes of nested fastener-set operations

Section 4.2.1 confirmed that the two controlled conditions were correctly reflected in the branching of the toy-model T-AOG. This subsection examines the different physical outcomes recorded along two representative branches. The grouping procedure identified the nested fastener sets $S_{\mathrm{Lid}}$ and $S_{\mathrm{Lid-Int}}$, which satisfy the inclusion relation in Eq. (7). These sets correspond to the combined three-bolt set in Figure 16(c) and the two-bolt subset in Figure 16(a), respectively.

Two representative paths are highlighted in the T-AOG shown in Figure 17. The red path denotes Case A, initiated by removal of the combined three-bolt set $S_{\mathrm{Lid}}$, whereas the green path denotes Case B, initiated by removal of the two-bolt subset $S_{\mathrm{Lid-Int}}$. Figure 18 shows the corresponding state transitions.

In Case A, shown in Figure 18(a), removal of $S_{\mathrm{Lid}}$ released the constraint associated with Lid, and the subsequent physical separation of Lid was successfully validated. As shown in Figure 18(a2), the assembly was reoriented to align the bolt-removal direction with the predefined tool-insertion axis before the fastener-removal operation. The T-AOG therefore records fastener-set removal and physical separation as distinct operations.

In Case B, shown in Figure 18(b), removal of $S_{\mathrm{Lid-Int}}$ released the local fastener-mediated constraint but did not divide the active liaison graph into separable units. The resulting connected intermediate state was therefore recorded without a separation transition. As shown in Figure 18(b3), the assembly was then reoriented to align the extraction direction of the interlock-like TopSnapM-Int interface with the predefined removal axis. From this state, the TopSnapM-Int residual interface was released as a separate constraint-release operation, after which the subsequent physical separation was successfully validated.

The two cases show that nested fastener sets can produce different physical outcomes despite sharing constituent fasteners. Removal of the three-bolt superset directly enabled physical separation, whereas removal of the two-bolt subset produced a connected intermediate state requiring an additional constraint-release operation.

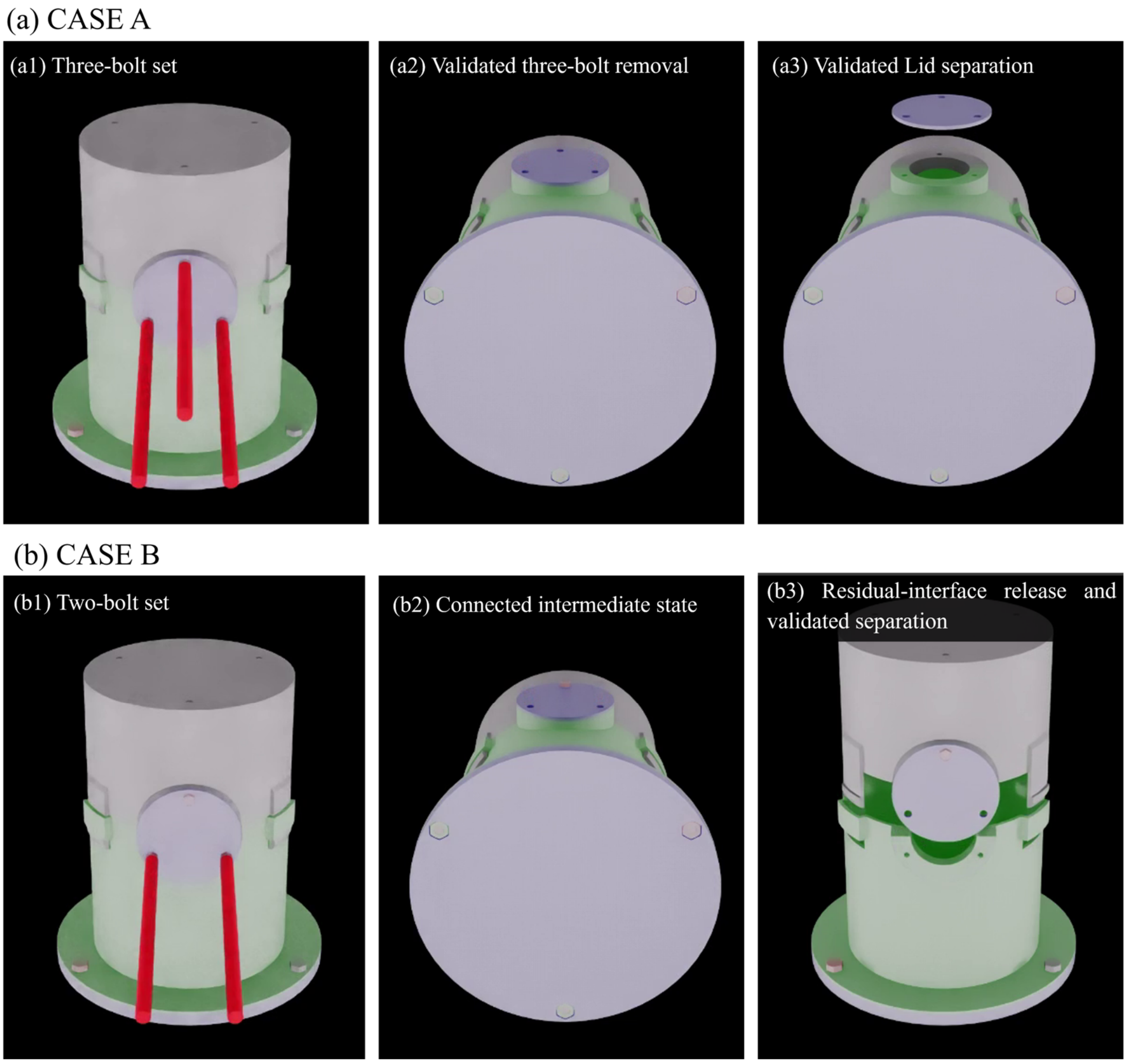

Figure 18. State transitions along the representative T-AOG paths in Figure 17: (a) Case A, removal of the combined three-bolt set, and (b) Case B, removal of the two-bolt subset.

### 4.3 Application to the OpenManipulator Chain gripper

The toy assembly demonstrated the fundamental operation-acquisition principles of the framework under controlled structural conditions. The second experiment examined whether the same processing pipeline could be applied to a denser product-derived electromechanical assembly containing configurations that were absent from, or simplified in, the toy assembly.

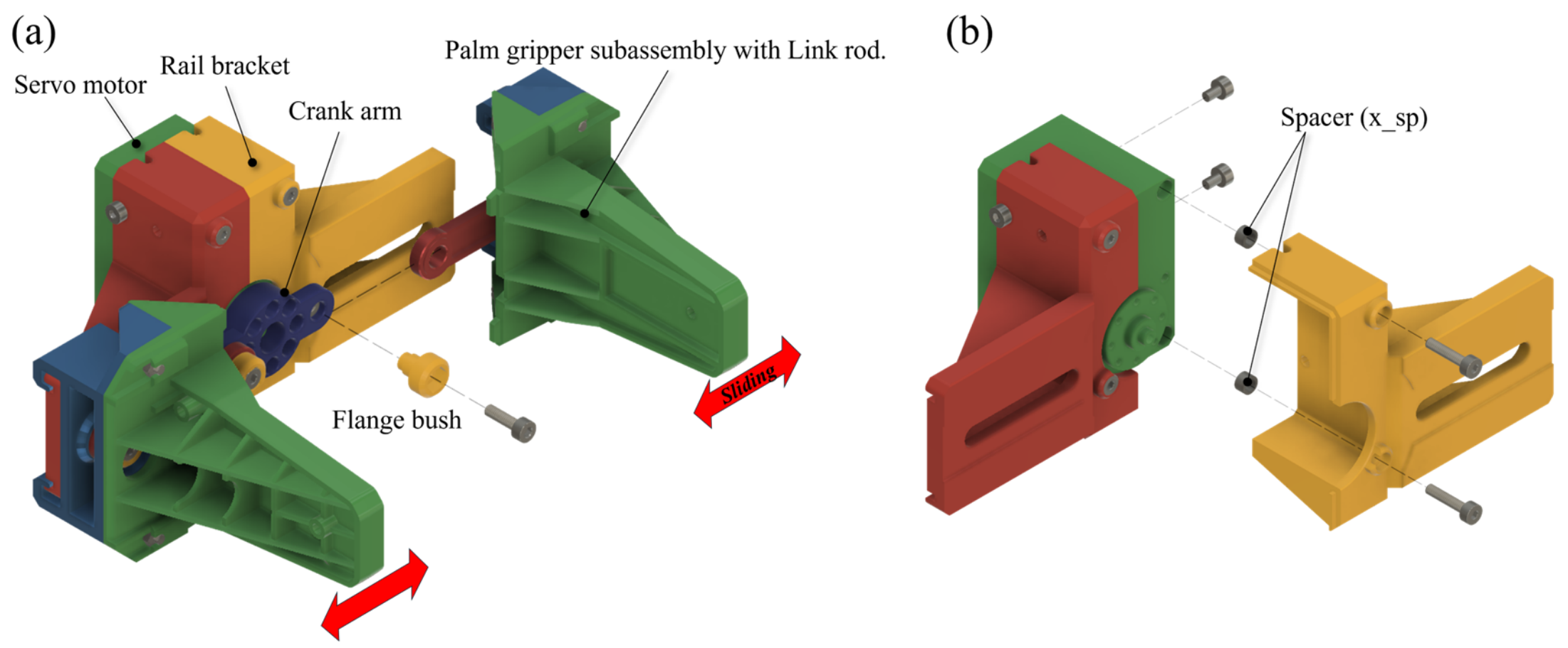

Figure 19. Representative structures of the OpenManipulator Chain gripper: (a) sliding interface between the rail bracket and a palm-gripper subassembly and (b) two x_sp spacers co-fastened with the corresponding rail bracket and servo-motor structure.

As shown in Figure 19(a), the gripper contains two approximately symmetric palm-gripper subassemblies that slide along their corresponding rail brackets. On each side, two x_sp spacers are co-fastened with the corresponding rail bracket and servo-motor structure, as illustrated in Figure 19(b). The assembly also contains compact fastening arrangements, link rods, flange bushes, and co-fastened components.

**4.3.1 Annotated liaison graph**

The framework completed fastener recognition, liaison-graph construction, and relation annotation for the gripper assembly. The resulting graph in Figure 20 preserves the central servo-motor structure, crank arm, repeated left-right mechanisms, and rail-guide contacts required for operation acquisition.

A representative contact-only configuration was identified around the four x_sp spacers. On each side, two spacers lie between a rail bracket and the servo-motor structure, and these components are jointly clamped by a common fastener set. Because the spacers are not independently fastened to either neighbor, their contacts were represented as contact-only rather than separate fastener-mediated constraints.

The link mechanisms were also preserved. Residual contacts that restrict direct translation but permit local rotation were examined for cylindrical compatibility and annotated as contact-rotatable when the corresponding geometric conditions were satisfied. Appendix D describes the identification procedure. Each link rod retained its relations with the flange bushes, fastening bolts, and neighboring components on both sides of the gripper.

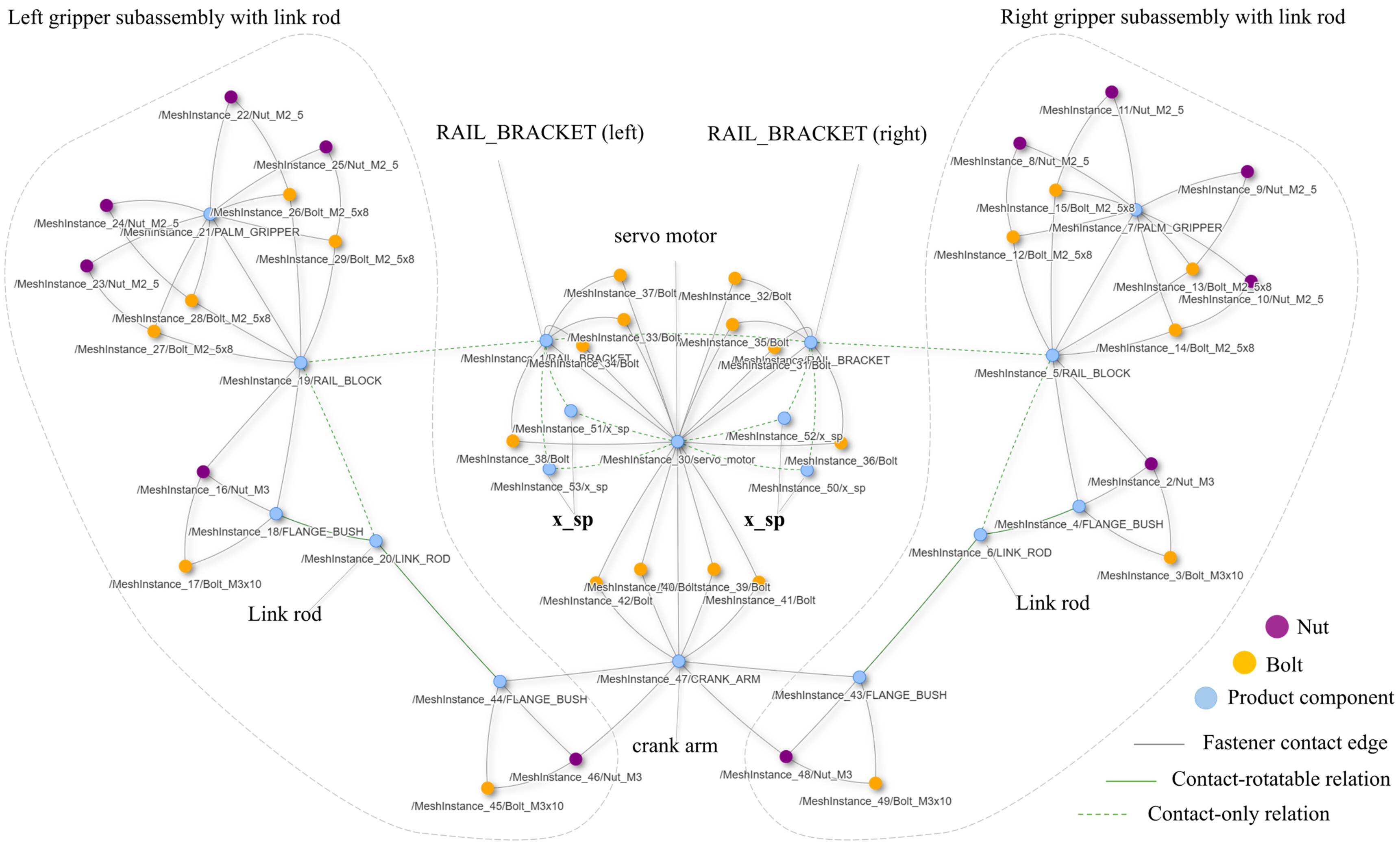

Figure 20. Annotated liaison graph acquired from the OpenManipulator Chain gripper. The graph preserves the central servo-motor structure, crank arm, repeated left-right mechanisms, and rail-guide contacts; node and edge categories are defined in the legend.

Because some component pairs that move relative to each other are separated by clearances as small as approximately 0.15 mm, the acquired contact-only relations can vary slightly among simulation runs. The fastener and contact-rotatable relations remained unchanged. Under the contact-only handling described in Section 3.2.5, the observed variation did not affect the resulting T-AOG.

#### 4.3.2 T-AOG generation and representative validated disassembly cases

Recursive virtual-disassembly exploration generated the T-AOG shown in Figure 21. Because of its size, the figure emphasizes the global branching structure and its principal accessibility bottleneck rather than individual nodes and edges.

The upper branches primarily represent removal of the surrounding gripper-side structures and converge where the crank-arm fasteners become accessible. Before this point, the palm-gripper structures obstructed access by the assumed tool proxy to the crank-arm fasteners. Their removal enables crank-arm release, after which decomposition of the servo-motor-containing subassembly proceeds in the lower branches. This state-dependent precedence was not specified manually, although the acquired order remains conditional on the assumed tool-proxy geometry.

Two representative branches were selected to examine multi-directional fastener removal, co-fastened components, and close-fitting guided motion. Figure 22 shows the corresponding operation and separation transitions.

**Case 1: Multi-directional fastener removal and separation of co-fastened components**

**Case 1.1: Removal of orthogonal direction groups**

The common fastener set comprised two direction groups, $A$ and $B$, each containing two bolts. Their removal directions differed by approximately 90°. Figure 22(a1) shows the tool-access evaluation for both direction groups. Figures 22(a2)–(a4) illustrate the $A \rightarrow B$ removal order: group $A$ was removed (a2), the assembly was subsequently reoriented for removal of group $B$ (a3), and group $B$ was removed (a4). The reverse order, $B \rightarrow A$, was retained as a separate T-AOG route.

**Case 1.2: Separation of the co-fastened x_sp spacer components**

Removal of the fastener set disconnected the rail bracket from the stay unit in the constraint graph and triggered reevaluation of the two x_sp spacers, which were geometrically retained but had only contact-only relations with neighboring structures. The graph update produced the rail bracket, two x_sp spacers, and the stay unit containing the servo motor; the first three were evaluated as candidate moving units.

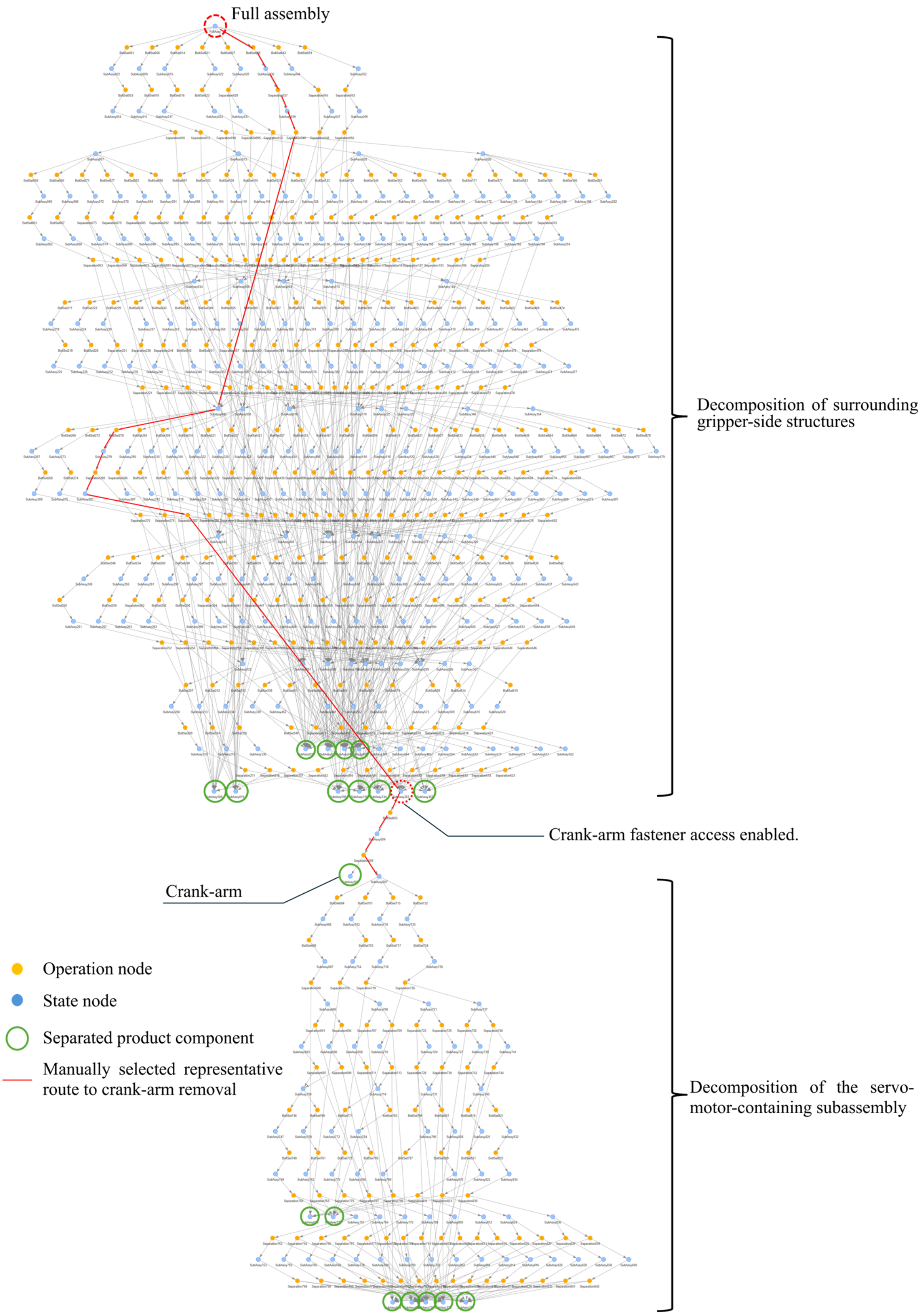

Figure 21. T-AOG acquired from the OpenManipulator Chain gripper. The root-oriented layout shows early decomposition of the gripper-side structures, convergence at the crank-arm-access bottleneck, and subsequent decomposition of the servo-motor-containing subassembly.

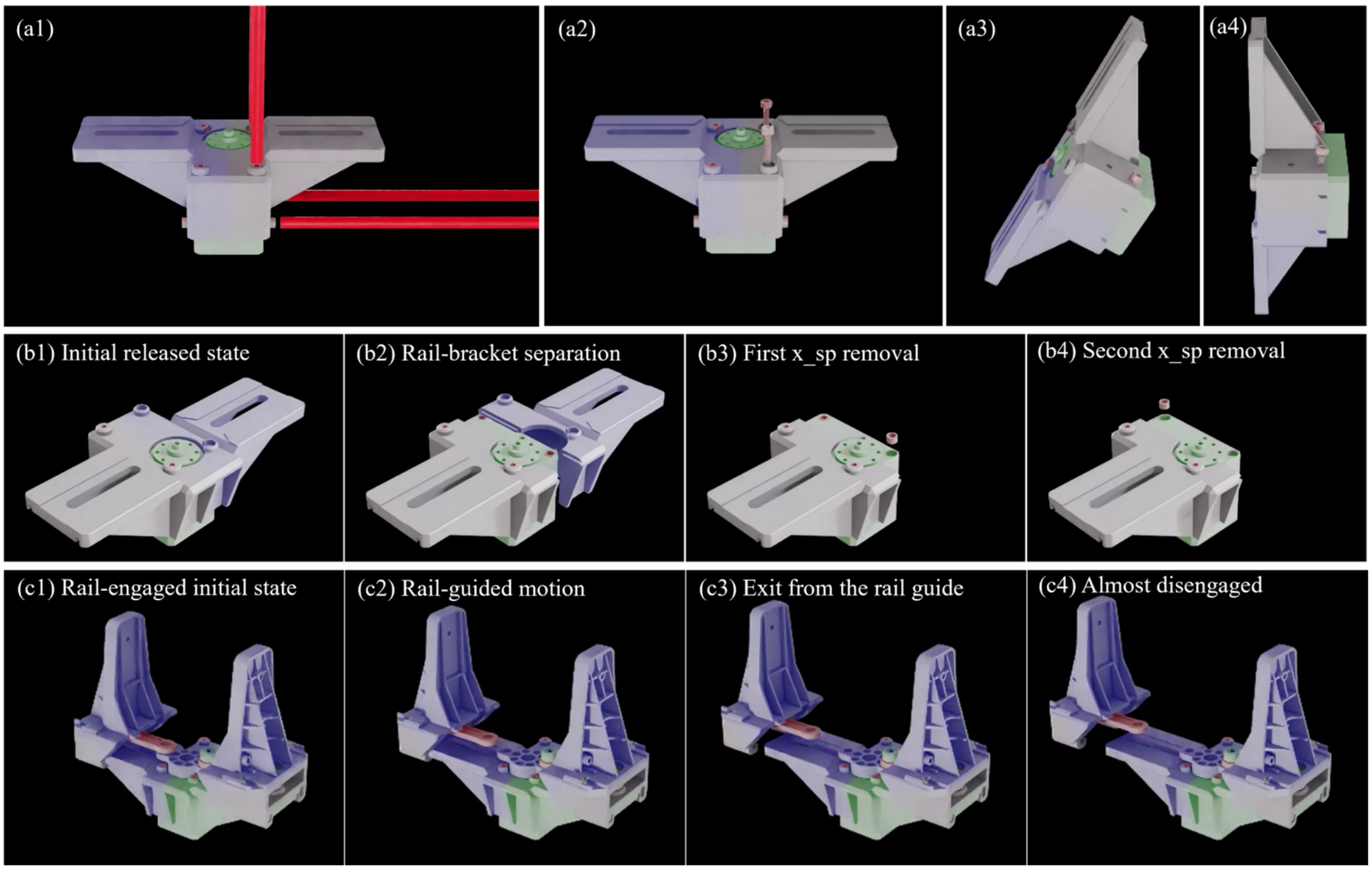

**Figure 22.** Representative operation and separation transitions for the OpenManipulator Chain gripper: (a) removal of two fastener direction groups differing by approximately 90°: (a1) tool-access evaluation, (a2) removal of the first two-bolt group, (a3) workpiece reorientation, and (a4) removal of the second group; (b) state-dependent separation of co-fastened components: rail-bracket separation followed by individual removal of the two x_sp spacers; and (c) guided disengagement of a palm-gripper subassembly through rail engagement, rail-guided motion, exit from the guide, and near-complete disengagement.

As shown in Figures 22(b1)–(b4), initial-motion screening showed that the rail bracket had to be separated before either spacer became removable. The spacers were then reevaluated and validated individually. Both removal orders of the two x_sp spacers were retained as alternative routes in the T-AOG. Thus, a single constraint release generated multiple moving-unit candidates whose separation order depended on the resulting assembly state.

**Case 2: Guided removal of the palm-gripper subassembly**

Figure 22(c) shows disengagement of a palm-gripper subassembly from its close-fitting rail interface. The lateral clearance measured from the CAD geometry was 0.1 mm on each side of the rail, or 0.2 mm in total. Initial-motion screening rejected arbitrary removal directions and identified an admissible direction aligned with the rail. This direction was used for candidate-path generation, and complete removal was validated through ghost-guided physics simulation.

Inspection of the representative transitions in Figure 22 against the represented assembly geometry confirmed that the acquired moving units, prerequisite operations, and separation directions were consistent with the rail-guide, co-fastened spacer, and crank-arm-access structures.

### 4.3.3 Example of simulation-derived operation durations

To demonstrate the operation-duration attributes recorded in the acquired T-AOG, a representative route from the initial gripper state to crank-arm removal was manually selected. The selected route is highlighted in red in Figure 21. It was selected for illustration and was not obtained through shortest-path or minimum-duration optimization.

Figure 23 shows the nine assembly states along the selected route, from the initial state to the state after crank-arm removal. Each state is shown using top and oblique views. The labels between successive states identify the corresponding fastener-removal or moving-unit-separation operation and its attributed duration.

For each direction group $g$ in a fastener set, the $n_{f,g}$ fasteners in that group were removed concurrently in a single physics trial. Let $t_{\mathrm{trial},g}$ denote the duration measured for that trial. Because the sequence-evaluation model treats the fasteners as individual removal tasks, the duration attributed to the complete fastener-removal operation was calculated as

$$t_{\mathrm{attr}} = \sum_{g \in G} n_{f,g} \; t_{\mathrm{trial},g}, \tag{17}$$

where $G$ is the set of direction groups in the fastener set. The fastener-removal durations shown in Figure 23 are these operation-level attributed durations. In contrast, the durations of moving-unit separation operations were obtained directly from the corresponding ghost-guided physics simulations. The attributed fastener-removal and simulated separation durations along the selected route totaled 19.87 s. The total route duration is therefore expressed as

$$T_{\mathrm{route}} = 19.87 \text{ s} + t_{\mathrm{setup}} + t_{\mathrm{change}}, \tag{18}$$

where $t_{\mathrm{setup}}$ and $t_{\mathrm{change}}$ denote the prescribed initial tool-setup and tool-change times, respectively. Because the current framework does not simulate tool setup and exchange time in the physics simulation, these auxiliary times are assigned during subsequent sequence evaluation. The need for a tool change can nevertheless be identified when successive fastener-removal operations require different tool specifications based on the estimated fastener sizes.

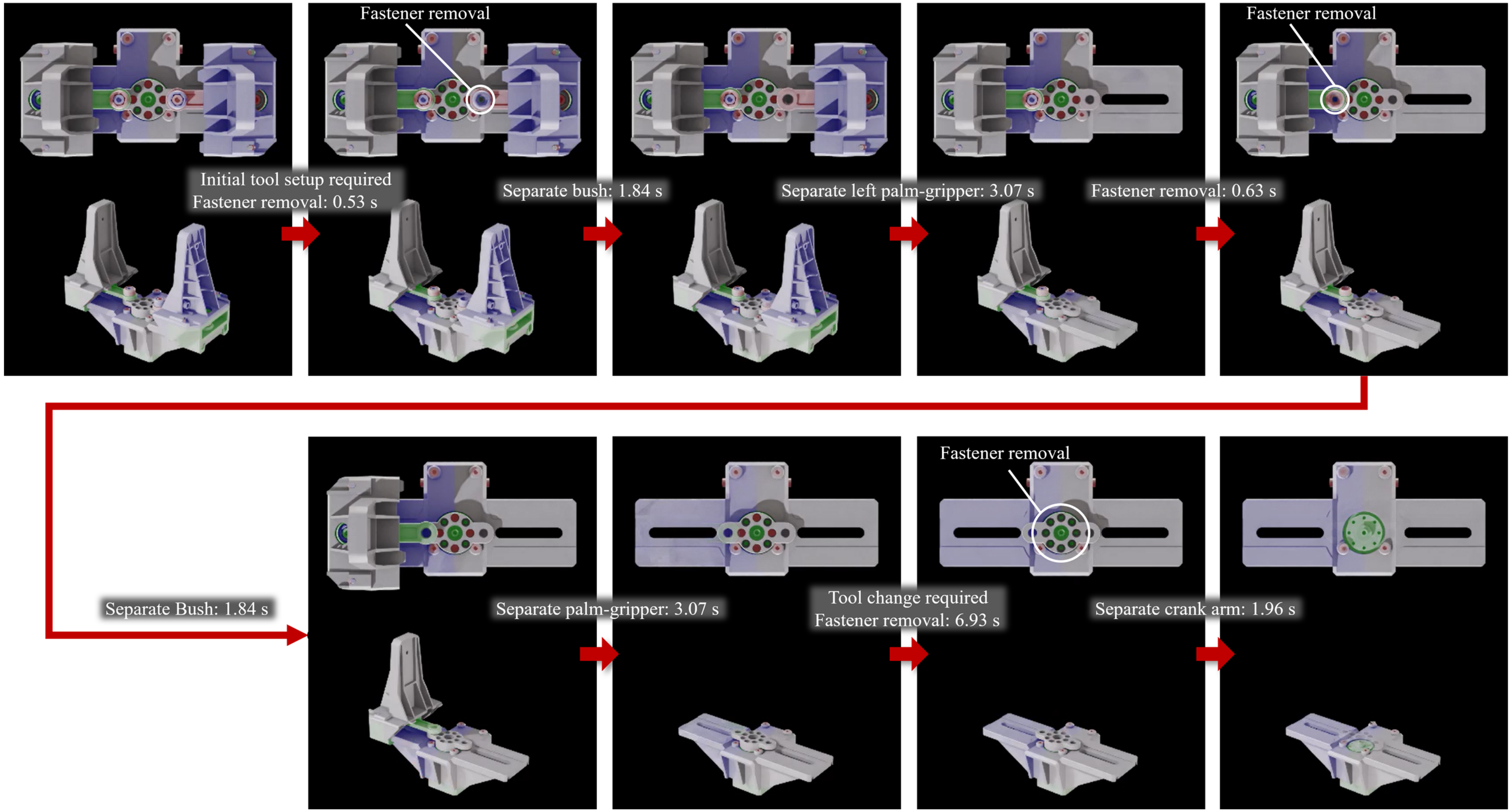

Figure 23. Operation-state sequence along the representative route from the initial gripper state to crank-arm removal. Each state is shown in top and oblique views, and transition labels indicate the corresponding operation durations.

# 5. Discussion

## 5.1 Interpretation of the acquired operation-level knowledge

The proposed framework represents disassembly as executable state transitions rather than only as product decomposition. In particular, fastener removal and physical separation are treated as distinct operations.

Conventional T-AOG representations may model a disassembly operation as a direct transition from a subassembly state to its decomposed child states：

$$\text{subassembly state} \rightarrow \text{disassembly operation} \rightarrow \text{child states.}$$

When the operation is interpreted as fastener removal, this representation implicitly assumes that removing the fastener immediately produces the decomposed child states. The fastener-centric representation instead introduces an explicit intermediate state:

$$\begin{gathered}\text{subassembly state} \rightarrow \text{fastener-removal operation} \rightarrow \text{released intermediate state} \\ \rightarrow \text{separation operation} \rightarrow \text{child states.}\end{gathered}$$

Because the units in the released intermediate state must be separated sequentially and may permit different separation orders, intermediate states preserve the partially separated configurations required to represent the corresponding operation sequences, as illustrated in Figure 22(b).

Explicit intermediate states increase the number of T-AOG nodes, but do not necessarily increase the number of alternative disassembly decisions. Instead, many nodes refine an existing branch by representing the physically meaningful state between constraint release and separation. This increases operational fidelity while preserving alternative executable branches.

## 5.2 State-dependent branch pruning and computational screening

Recursive exploration evaluates every operation-level fastener set identified in each assembly state but does not enumerate arbitrary orders of individual fasteners. Grouping fasteners that jointly release the same constraint therefore defines a mechanically meaningful operation space before state exploration. Within this operation space, state-dependent screening is applied at two levels: before fastener removal and before separation-path generation.

Table 8. State-dependent pruning of fastener-set candidate evaluations.

| Evaluation stage | OpenManipulator Chain gripper |
|---|---|
| State-dependent fastener-set candidate evaluations | 216 |
| Evaluations pruned by visibility | 28 |
| Evaluations pruned by tool accessibility | 83 |
| Evaluations pruned by torque-driven removal | 0 |
| Evaluations accepted as executable fastener-set operations | 105 |
| Total evaluations pruned | 111 |
| Overall pruning rate | 51.4% |

*Each evaluation represents an operation-level fastener set evaluated in a particular assembly state rather than an individual fastener. The same set is counted again when reevaluated under different surrounding-component conditions.*

Table 9. Computational screening of state-dependent moving-unit evaluations.

| Evaluation stage | OpenManipulator Chain gripper |
|---|---|
| State-dependent moving-unit candidate evaluations | 993 |
| Rejected by initial-motion screening | 736 |
| Passed initial-motion screening and proceeded to path generation | 257 |
| Evaluations prevented from proceeding to path generation | 74.1% |

*Candidate evaluations were accumulated across repeated screening rounds. A remaining unit was counted again when reevaluated after another unit had been separated.*

As discussed in Section 4.2, the controlled toy assembly was used to verify that operation branches were correctly generated or suppressed according to the specified precedence and tool-accessibility conditions. Because these conditions were deliberately introduced for verification, the resulting rejection counts are not interpreted as practical measures of branch pruning or computational screening. The quantitative analysis therefore focuses on the product-derived OpenManipulator Chain gripper.

Table 8 summarizes the fastener-set candidate evaluations rejected at each state-dependent screening stage. The framework rejected 111 of the 216 fastener-set candidate evaluations for the OpenManipulator Chain gripper, corresponding to an overall pruning rate of 51.4%. Tool accessibility was the principal rejection criterion, while no candidate evaluation failed during torque-driven removal. Tool accessibility can also induce state-dependent operation precedence, as removing obstructing components can enable previously inaccessible fastener operations. As in physical disassembly, these accessibility decisions depend on tool dimensions, represented here by the configured tool proxies.

Table 9 summarizes the cumulative state-dependent moving-unit evaluations performed during initial-motion screening, including reevaluation of remaining units after successful separation operations. Of the 993 evaluations, 736 were rejected before path generation, while 257 proceeded to path generation.

Thus, initial-motion screening prevented 74.1% of the moving-unit evaluations from reaching the more computationally demanding path-generation stage. The large number of evaluations partly reflects the structure of the OpenManipulator Chain gripper and the state-dependent reevaluation of contact-only components. In particular, the gripper contains four co-fastened spacers represented by contact-only relations. These spacers are reconsidered as candidate moving units when the edge state of an adjacent node changes, but are rejected by initial-motion screening while the rail bracket continues to block their removal.

The two screening levels serve different roles. Visibility and tool-accessibility evaluation suppress infeasible fastener-removal branches before they are added to the T-AOG, whereas initial-motion screening prevents graph-derived but currently blocked moving-unit candidates from proceeding to the more computationally demanding path-generation and ghost-validation stages. Fastener-set grouping defines the operation space, and the subsequent state-dependent evaluations control both T-AOG branching and the computational workload of physical validation.

### 5.3 Limitations

The current implementation is limited by the fidelity and physical scope of the virtual model, the configured evaluation conditions, the range of separation motions generated and validated, and its computational cost and scalability.

#### 5.3.1 Scope and fidelity of the physical model

The fidelity of the acquired contact relations depends on the collision representation and contact-generation settings used in the physics simulator. The framework uses a signed distance field (SDF) representation to preserve concave and finely detailed geometries, such as threaded holes and mating-thread contacts, that cannot be adequately represented by convex approximations. Insufficient SDF resolution or an inappropriate contact-generation distance may omit relevant contact regions or incorrectly classify physical clearances as contacts. The SDF resolutions and contact-generation settings used in this study are reported in Supplementary Table S11.

The framework assumes rigid components. Flexible cables, wire harnesses, rubber parts, compliant clips, and adhesive joints may deform, peel, wrap around neighboring components, or change their contact conditions during removal and therefore cannot be represented adequately by fixed-shape rigid bodies.

Residual interfaces are therefore represented at an operation-oriented level without identifying the underlying mechanism as a specific snap-fit, latch, compliant clip, or elastic mechanism. The corresponding operation is treated as an abstract constraint-release event, implemented by temporarily disabling collision between the moving unit and the stay-unit endpoint component of the selected residual edge without reproducing deformation, release force, or detailed disengagement behavior. Physics-based validation applies only to the subsequent separation motion under this assumed released condition, not to the release action itself.

Consistent with the design-for-disassembly objective stated in the Introduction, the present framework focuses on acquiring disassembly knowledge from nominal CAD models during product design. The virtual assembly therefore does not account for end-of-life deviations such as wear, deformation, corrosion, fracture, missing parts, or seized interfaces. Such conditions can alter accessibility, required effort, and operation feasibility and have been explicitly incorporated into robotic disassembly-sequence models in previous work [7]. Extending the framework to inspected end-of-life products would require observed condition information to update the nominal CAD-derived state.

### 5.3.2 Scope of motion generation and validation

The initial-motion screening targets locally translational separation associated with dominant interfaces. The surface-informed OBB provides operation-relevant directions for common pull-out and sliding motions without requiring exhaustive directional sampling. However, the current method does not explicitly generate rotational or coupled translational–rotational release motions, such as twist-and-pull operations or the disengagement of rotational locking mechanisms. Components requiring such motions may therefore not proceed to path generation under the current screening procedure.

During ghost-guided validation, friction between the moving unit and the surrounding assembly is temporarily set to zero to distinguish geometric path feasibility from frictional wedging in close-fitting interfaces. The validation therefore does not evaluate friction-dependent forces or locking behavior that may occur during physical execution.

The operation durations reported in Figure 23 demonstrate that the framework can automatically acquire simulation-derived durations and aggregate them along alternative T-AOG routes. These durations inevitably depend on the configured motion speeds, physics timestep, controller parameters, and other simulation inputs. Their sensitivity to these settings and their correspondence with physical task times were not evaluated in the present study. Calibration against observed manual or robotic operations would therefore be required before the acquired durations could be used as practical disassembly-time estimates. Nevertheless, durations obtained under consistent simulation conditions may support future relative comparisons of alternative routes or product designs before and after modification.

The current framework does not include gravity in the virtual evaluation. The stay unit is maintained by fixing its position and orientation, without evaluating whether the prescribed pose can be realized by a feasible fixture or support arrangement. Consequently, the framework does not assess the gravity-dependent stability of the stay unit after separation or of the moving unit at its designated placement location. Previous physics-based assembly-planning research, such as ASAP, has incorporated gravity into sequence generation by evaluating subassembly stability under limited holding and support conditions [38]. Such evaluation will become important when extending the present framework toward robotic manipulation. Because the physics simulator used in this study can directly apply gravitational loading, gravity-aware evaluation of fixture feasibility and post-separation stability is a natural subject for future work.

### 5.3.3 Computational cost and scalability

Rendering, semantic observation, geometric evaluation, FIM-based path generation, and physics validation use GPU acceleration. GPU computation time and memory capacity therefore constrain the scale and geometric resolution of the assemblies that can be evaluated.

The two assemblies exhibited similar GPU-memory usage despite their different component counts. The toy assembly occupied a larger workspace, whereas the OpenManipulator Chain gripper contained more components within a smaller spatial extent, as shown in Table 4. Both occupancy-grid construction and FIM-based path generation require spatial discretization. The memory required by each stage therefore depends on workspace volume and discretization resolution rather than on component count alone. Processing time is also affected by repeated scene reconstruction, including component activation and deactivation, pose restoration, and moving-subassembly construction. Reusing previously evaluated paths and geometric data reduces recomputation but increases system-memory consumption.

The experiments establish computational feasibility for the two evaluated assemblies but do not establish processing-time or memory scaling for substantially larger products. Scalability therefore involves a trade-off among GPU memory, system memory, spatial resolution, and recomputation time.

## 6. Conclusion

This paper presented a fastener-centric virtual-disassembly framework that converts CAD-derived virtual assemblies into operation-level executable disassembly knowledge. The framework treats fastener removal as a state-dependent constraint-release operation distinct from the subsequent physical separation of a part or subassembly. Fasteners that jointly release the same constraint are grouped into an operation-level set, and their removability is evaluated using visibility, tool accessibility, and torque-driven virtual-removal trials. Graph-based structural reasoning, path generation, and physics-based validation are then used to acquire the resulting separation operations. The acquired operations, intermediate states, dependencies, validated paths, and simulation-derived attributes are organized into a reusable Transformed AND/OR Graph.

The controlled toy-model verification confirmed that the specified precedence and tool-accessibility conditions were correctly reflected in the generated T-AOG branches. The OpenManipulator Chain gripper extended the evaluation to a denser product-derived assembly containing co-fastened components, repeated mechanisms, graph-derived moving units, and close-fitting sliding interfaces. The gripper results indicate that structural and operational constraints can substantially reduce the number of candidate operations passed to detailed physical validation.

By demonstrating that product-specific, state-dependent disassembly knowledge can be acquired from nominal CAD data, the framework provides a computational basis for evaluating disassemblability before physical products are manufactured.

The primary next step is to derive task-specific selective and complete disassembly sequences from the acquired T-AOG. Alternative sequences could be compared using operation time, tool and direction changes, moving-unit mass, path length, component value, and material information, where available. Such sequence-dependent analysis would allow operations that dominate disassembly effort or restrict access to target components to be identified and traced back to fastening arrangements, tool-access conditions, joining mechanisms, and component layouts. The resulting analysis could support redesign recommendations such as fastener reduction and provide inputs to methods that derive component influence scores from disassembly-sequence outcomes and CAD-based constraint graphs [45]. Deriving and comparing task-specific sequences would therefore extend the present framework from executable-knowledge acquisition toward quantitative DfD evaluation and design feedback.

In parallel, extending the framework toward robotic disassembly will require broader mechanism and motion models. Mechanism-specific actions and tools for connector and snap-fit release [46], coupled translational-rotational motion generation, and cross-stage fixture and support-stability evaluation [47] would expand the range of operations that can be physically evaluated. Together with task-specific sequence analysis, these extensions would allow the acquired T-AOG to support broader quantitative DfD evaluation and robotic-disassembly planning before manufacturing.

## CRediT authorship contribution statement

Shiki Iwase: Conceptualization, Methodology, Software, Validation, Formal analysis, Investigation,

Data curation, Visualization, Writing – original draft, Writing – review & editing.

## Declaration of competing interest

The author is named as an inventor on a pending patent application related to this work. The patent application is assigned to Toyota Central R&D Labs., Inc. The author declares no other known competing financial interests or personal relationships that could have appeared to influence the work reported in this paper.

## Funding

This research received no specific grant from funding agencies in the public, commercial, or not-for-profit sectors.

## Data availability

The principal data and evaluation settings supporting the reported results are provided in the article and its Supplementary Material. The original OpenManipulator Chain geometry is publicly available from the source cited in the article under the CC BY 4.0 license. The toy-assembly model and research software developed for this study are proprietary and are not publicly available. Detailed simulation outputs beyond those reported in the article and Supplementary Material are also not publicly available because they are subject to the intellectual-property and confidentiality restrictions of the author's employer.

## Declaration of generative AI and AI-assisted technologies in the manuscript preparation process

During the preparation of this work, the author used Microsoft 365 Copilot and OpenAI tools and models, including ChatGPT, Codex, and GPT-6.1 Sol, to support language editing, improve the clarity and organization of the manuscript, assist in the initial conceptualization of two explanatory figures, and check internal consistency among the main text, equations, algorithms, figures, and Supplementary Material. The author critically reviewed, verified, and edited all AI-assisted output and takes full responsibility for the content of the published article.

## Appendix

### Appendix A. Multi-view attribute combination for fastener recognition

The fastener-recognition network follows the multi-view feature-extraction and aggregation paradigm of MVCNN [41] and uses attention-based aggregation to account for differences in the informativeness of individual views [48]. As shown in Figure A1, a shared ResNet-34 [49] backbone supports geometric-attribute prediction and fastener-class prediction.

Training is performed in two stages. During pretraining, individual rendered views are used to train the shared backbone and attribute-prediction heads. During fine-tuning, each sample consists of 12 rendered views of the same geometry. The pretrained backbone and attribute heads are jointly optimized with an attention module and a softmax-based classification head for three target classes: hex-head bolts, socket-head cap screws, and nuts. The attention module combines the view-wise feature representations for class prediction. For attribute prediction, the sigmoid probability $p_{a,v}$ of attribute $a$ in view $v$ is combined across the 12 views using noisy-OR aggregation:

$$p_a = 1 - \prod_{v=1}^{12} \left(1 - p_{a,v}\right). \tag{A.1}$$

### A.1 Training and validation data

All training images were generated synthetically in NVIDIA Isaac Sim. As shown in Figure A2, attribute pretraining used 106,140 single-view images representing 20 three-dimensional geometric categories with variations in dimensions, thicknesses, fillets, and other category-specific parameters. Camera directions were sampled randomly, excluding elevation angles between $-30°$ and $30°$ to reduce nearly edge-on observations. The images were divided into 80% training and 20% validation subsets.

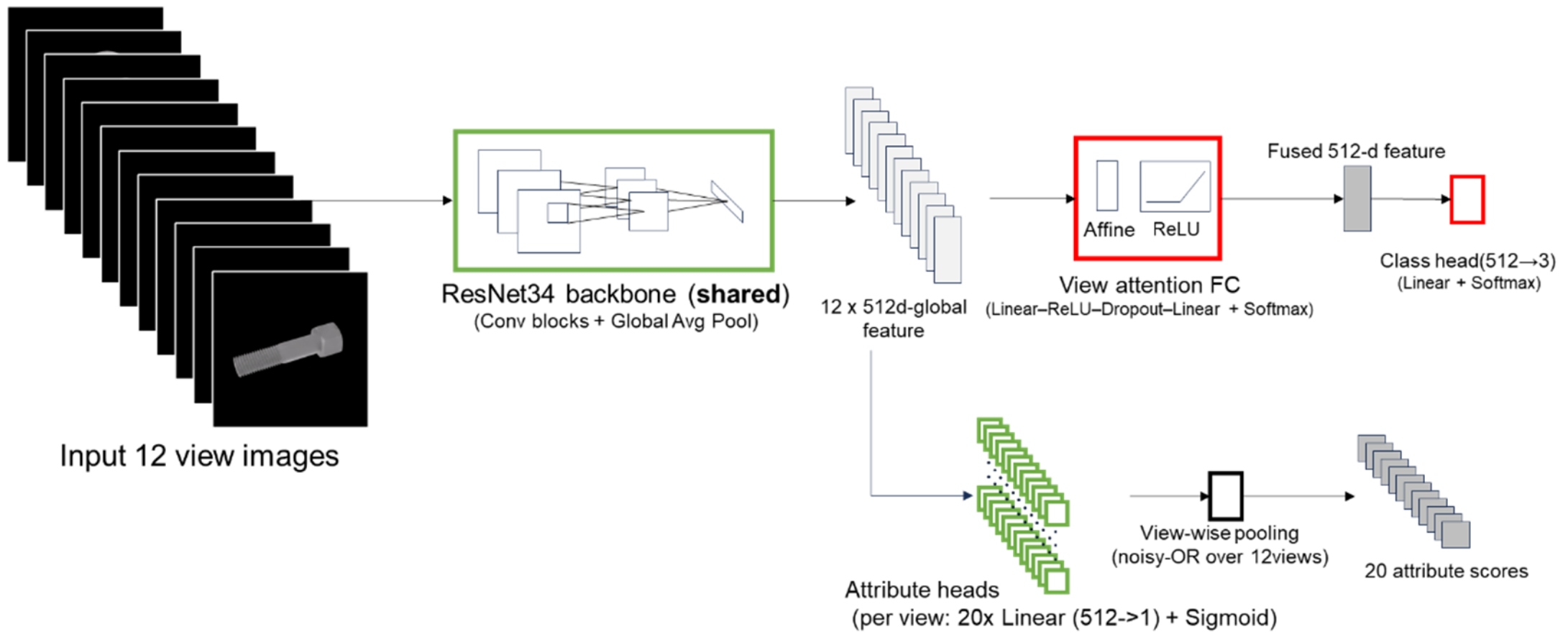

Figure A1. Multi-view network architecture.

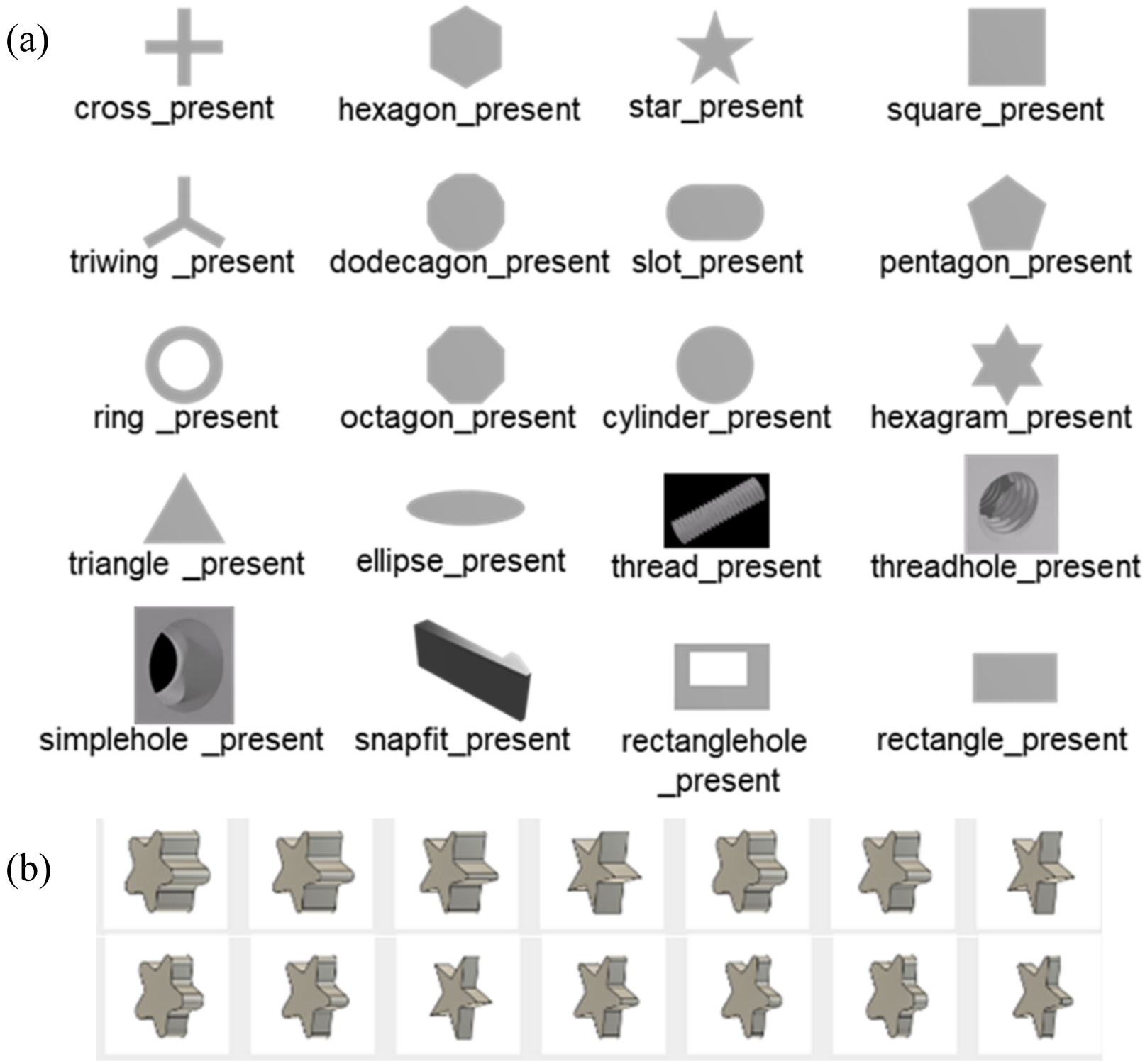

Figure A2. Three-dimensional geometric categories used for attribute pretraining. (a) Representative instances of the 20 basic geometric categories. (b) Example geometric variations generated for the star_present attribute by varying thickness, fillet geometry, and other shape parameters.

Twelve-view fine-tuning used 84,960 images forming 7,080 samples. Of these, 3,420 samples represented the three target fastener classes and provided both class and attribute supervision. The remaining 3,660 samples represented attribute-specific geometries and provided attribute supervision only. The samples were divided at the sample level into 80% training and 20% validation subsets, ensuring that all 12 views of each geometry remained in the same subset. Fine-tuning and inference used the same 12 fixed camera positions shown in Figure 3. Both pretraining and fine-tuning used the AdamW optimizer; the optimization settings, batch sizes, and data split are summarized in Supplementary Table S10.

The epoch with the highest macro-average precision across all 20 attributes was selected, yielding a validation macro-AP of 0.9935. The three attributes used in the final consistency check each achieved an AP of 1.0000. Attribute-specific decision thresholds were selected to maximize validation F1, as listed in Table A1.

Table A1. Decision thresholds for the three attributes used in fastener-class consistency checking.

| Attribute | Decision threshold | Validation F1 |
|---|---|---|
| External hexagonal head feature present (hexagon_present) | 0.660 | 1.000 |
| External thread present (thread_present) | 0.920 | 1.000 |
| Threaded hole present (threadhole_present) | 0.350 | 1.000 |

## A.2 Fastener-class prediction and consistency checking

The class-prediction head achieved an accuracy of 1.000 on the validation subset of the synthetic fastener-class samples. A predicted class is retained only when its thresholded attributes satisfy the corresponding logical definition in Table A2.

Table A2. Logical attribute combinations used for fastener-class assignment.

| Attribute | Hex-head bolt | Socket-head cap screw | Nut |
|---|---|---|---|
| External hexagonal head feature present (hexagon_present) | ✓ | × | ✓ |
| External thread present (thread_present) | ✓ | ✓ | × |
| Threaded hole present (threadhole_present) | × | × | ✓ |

A check mark indicates that the attribute must be present, whereas a cross indicates that it must be absent. Predictions inconsistent with these definitions are rejected, and the corresponding components remain ordinary product-component nodes in the liaison graph. This consistency check provides an operational rejection rule for components that do not match the three target fastener classes, rather than assigning every product component to a fastener class. Component-family recognition results for the two evaluated assemblies and representative non-fastener predictions rejected by attribute-consistency filtering are provided in Supplementary Tables S1–S3.

## Appendix B. Surface-informed OBB axis selection

The surface-informed OBB basis is constructed from area-weighted directions of geometrically verified planar surface groups.

### B.1 Extraction of dominant surface directions

The procedure is applied to all polygonal meshes comprising the evaluated component or moving unit. For each valid polygonal face, a unit normal $\boldsymbol{n}_f$ and area $A_f$ are calculated using Newell's method [50]. Because $\boldsymbol{n}_f$ and $-\boldsymbol{n}_f$ represent the same undirected plane orientation, their signs are ignored during grouping.

Faces are examined in descending order of area, with each unassigned face defining a fixed seed plane. A face is assigned to the seed-plane group only when its normal is sufficiently aligned with the seed normal and all its vertices lie within the plane-distance tolerance:

$$\left|\boldsymbol{n}_f \cdot \boldsymbol{n}_s\right| \geq \cos\theta_{\text{plane}}, \qquad \max_{\boldsymbol{v}\in V_f}\left|(\boldsymbol{v} - \boldsymbol{p}_s) \cdot \boldsymbol{n}_s\right| \leq \varepsilon_{\text{plane}}, \tag{B.1}$$

where $\boldsymbol{n}_s$ and $\boldsymbol{p}_s$ are the seed normal and a point on the seed face, and $V_f$ is the set of face vertices. The plane-distance tolerance is an absolute physical distance of $\varepsilon_{\text{plane}}$ = 0.1 mm, independent of the mesh bounding-box diagonal. Mesh transformations and the scene's length units are taken into account when applying this tolerance in local coordinates. The seed plane remains fixed during grouping, preventing the group from progressively following a curved surface.

Each planar group retains its seed normal and the sum of its face areas. Groups with approximately parallel normals are then aggregated into surface-direction clusters, including groups on spatially separate parallel planes. A group direction $\boldsymbol{n}_g$ is assigned to a cluster with representative direction $\boldsymbol{d}_k$ when $|\boldsymbol{n}_g \cdot \boldsymbol{d}_k| \geq \cos\theta_{\text{cluster}}$.

Within each direction cluster $C_k$, normal signs are aligned with the current representative direction and accumulated using planar-group-area weighting:

$$\boldsymbol{s}_k = \sum_{g\in C_k} A_g \tilde{\boldsymbol{n}}_g, \; \boldsymbol{d}_k = \frac{\boldsymbol{s}_k}{\| \boldsymbol{s}_k \|}, \; A_k = \sum_{g\in C_k} A_g\,, \tag{B.2}$$

where $A_g$ is the accumulated face area of planar group $g$ and $\tilde{\boldsymbol{n}}_g$ is its sign-aligned seed normal. For each mesh, only the clusters with the largest accumulated areas are retained, up to the limit specified in Table B1. The retained cluster directions are transformed into a common world coordinate system and clustered again to obtain surface directions for the complete component or moving unit.

Table B1. Principal parameters used for surface-informed OBB construction.

| Parameter | Value | Purpose |
|---|---|---|
| Seed-plane normal tolerance, $\theta_{\text{plane}}$ | 1° | Maximum undirected angular difference between a face normal and the seed normal |
| Plane-distance tolerance, $\varepsilon_{\text{plane}}$ | 0.1 mm | Maximum physical distance of any face vertex from the seed plane, accounting for mesh transformations and scene length units |
| Surface-direction clustering tolerance, $\theta_{\text{cluster}}$ | 8° | Maximum undirected angular difference used to aggregate verified planar-group directions |
| Maximum retained clusters per mesh | 8 | Number of dominant local surface-direction clusters retained before global clustering |
| Orthogonality tolerance, $\theta_{\text{ortho}}$ | 12° | Maximum permitted deviation from exact orthogonality |
| Minimum less-dominant-direction area ratio, $r_{\text{second}}$ | 0.02 | Minimum surface-area fraction represented by the less dominant direction of a candidate pair |
| Minimum combined pair-area ratio, $r_{\text{pair}}$ | 0.25 | Minimum total surface-area fraction represented by the selected direction pair |

## B.2 Selection of two approximately orthogonal directions

Let $\boldsymbol{d}_i$ and $\boldsymbol{d}_j$ denote two global cluster directions with accumulated areas $A_i$ and $A_j$. The clusters are sorted in descending order of accumulated area, so that $i < j$ implies $A_i \geq A_j$. Each pair $i < j$ is then examined as a candidate for defining the OBB orientation. A pair is considered approximately orthogonal when

$$|\boldsymbol{d}_i \cdot \boldsymbol{d}_j| \leq \sin \theta_{\text{ortho}}, \tag{B.3}$$

where $\theta_{\text{ortho}}$ is the permitted deviation from exact orthogonality.

To prevent a minor local surface, such as a narrow chamfer, from being selected as the second surface direction, the less dominant direction must satisfy

$$\frac{A_j}{A_{\text{total}}} \geq r_{\text{second}}, \tag{B.4}$$

where $A_{\text{total}}$ is the total evaluated face area.

Each admissible pair is scored as

$$s_{ij} = (A_i + A_j)(1 - |\boldsymbol{d}_i \cdot \boldsymbol{d}_j|), \tag{B.5}$$

which favors pairs representing large surface areas and approaching exact orthogonality. The highest-scoring pair is retained only when

$$\frac{A_i + A_j}{A_{\text{total}}} \geq r_{\text{pair}}. \tag{B.6}$$

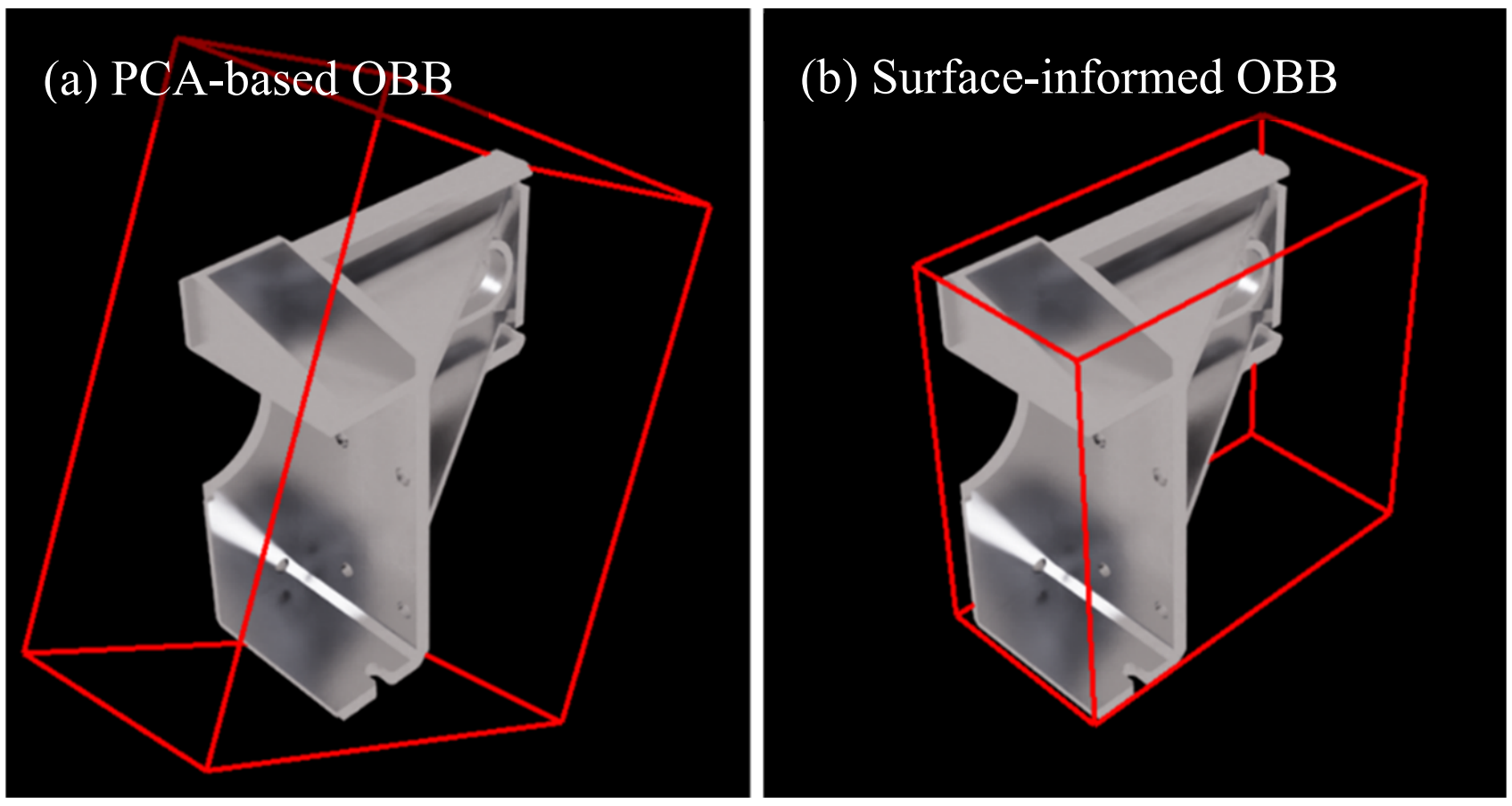

Figure B1. Comparison of (a) a conventional vertex-based PCA OBB and (b) the surface-informed OBB used in this study.

If no pair satisfies these conditions, the procedure falls back to a principal component analysis (PCA)-based OBB.

### B.3 Construction of the OBB basis

For the selected pair $(\boldsymbol{d}_i, \boldsymbol{d}_j)$, the first OBB axis is set to $\boldsymbol{a}_1 = \boldsymbol{d}_i$. The second selected direction $\boldsymbol{d}_j$ is orthogonalized with respect to $\boldsymbol{a}_1$, and the third axis is obtained from their cross product:

$$\boldsymbol{a}_2 = \frac{\boldsymbol{d}_j - (\boldsymbol{d}_j \cdot \boldsymbol{a}_1)\boldsymbol{a}_1}{\|\boldsymbol{d}_j - (\boldsymbol{d}_j \cdot \boldsymbol{a}_1)\boldsymbol{a}_1\|}, \ \boldsymbol{a}_3 = \frac{\boldsymbol{a}_1 \times \boldsymbol{a}_2}{\|\boldsymbol{a}_1 \times \boldsymbol{a}_2\|}. \tag{B.7}$$

The resulting orthonormal basis is used to construct an OBB enclosing all vertices of the component or moving unit. Its positive and negative axes provide the six candidate directions defined in Eq. (9) for local probing and initial-motion screening.

Figure B1 illustrates the practical difference between a conventional vertex-based PCA OBB and the surface-informed OBB. In this example, the global vertex distribution causes the PCA axes to become oblique to the dominant planar surfaces. The surface-informed procedure instead derives its basis from dominant, approximately orthogonal surface-normal directions and therefore aligns the OBB axes with the principal planar interfaces. The resulting axes provide operation-relevant directions for the residual-interface probing in Section 3.1.3 and the initial-motion screening in Section 3.2.5.

## Appendix C. FIM propagation-speed field and path extraction

### C.1 Propagation-speed field

The occupancy grid covers a world-coordinate axis-aligned region containing the current assembly geometry, the FIM start and goal points, and the endpoint of the prescribed initial straight displacement. Let $r_{\text{band}}$ denote the radius of the bounding sphere enclosing the surface-informed OBB of the moving unit. The region is expanded by $r_{\text{band}}$, supplemented by a two-cell boundary margin, aligned with the voxel grid, and discretized using the uniform three-dimensional grid defined in Section 3.2.7.

Let $\text{occ}(\boldsymbol{x})$ denote the occupancy state of grid cell $\boldsymbol{x}$, where $\text{occ}(\boldsymbol{x}) = 0$ indicates free space. For each free cell, $\text{SDF}(\boldsymbol{x})$ gives the distance to the surrounding obstacle geometry. The normalized clearance is

$$\phi(\boldsymbol{x}) = \text{clip}\left(\frac{\text{SDF}(\boldsymbol{x})}{r_{\text{band}}}, 0, 1\right). \tag{C.1}$$

Here, $\text{clip}(z, 0, 1)$limits $z$ to the interval $[0, 1]$. Consequently, $\phi(\boldsymbol{x}) = 0$ at an obstacle boundary and reaches $1$ when the obstacle distance is equal to or greater than $r_{\text{band}}$. Although the propagation speed could be defined directly from the normalized clearance, the intermediate openness measure is introduced to separate the clearance-to-openness mapping from the preference assigned to open regions during path generation. The normalized clearance is mapped to an openness measure:

$$O(\boldsymbol{x}) = O_{\min} + (1 - O_{\min})\phi(\boldsymbol{x})^{\gamma}, \tag{C.2}$$

where $O_{\min} > 0$ retains low-clearance free cells as traversable, and $\gamma$ controls the dependence on clearance. The FIM propagation speed in free space is then

$$F(\boldsymbol{x}) = s_{\min} + (1 - s_{\min})O(\boldsymbol{x})^{k_O}, \ \text{occ}(\boldsymbol{x}) = 0. \tag{C.3}$$

The artificial speed $F(\boldsymbol{x})$ does not represent the physical velocity of the moving unit. The lower bound $s_{\min} > 0$ makes low-clearance cells less preferable without excluding them before physics-based validation, while $k_O$controls the preference for open regions. Occupied cells are excluded from propagation.

The implementation uses $O_{\min} = 0.01$, $\gamma = 1.0$, and $s_{\min} = 0.1$. Two speed fields are evaluated using $k_O \in \{2, 20\}$. The smaller exponent produces a gradual clearance preference, whereas the larger exponent more strongly favors open regions.

## C.2 Traversal cost and arrival-time field

The inverse of the propagation speed defines the local traversal cost per unit distance:

$$w(\boldsymbol{x}) = \frac{1}{F(\boldsymbol{x})}. \tag{C.4}$$

The minimum arrival-time field satisfies

$$\| \nabla T(\boldsymbol{x}) \| = w(\boldsymbol{x}) = \frac{1}{F(\boldsymbol{x})}, \; T(\boldsymbol{x}_{\text{goal}}) = 0. \tag{C.5}$$

Here, $T(\boldsymbol{x})$ represents the minimum accumulated traversal cost from the goal to $\boldsymbol{x}$ under the artificial propagation-speed field $F(\boldsymbol{x})$. FIM numerically solves this Eikonal equation on the three-dimensional grid [42].

## C.3 Path extraction and selection

For each value of $k_O$, a candidate path is extracted from the start point by following decreasing values of $T(\boldsymbol{x})$ until the goal is reached. If paths are successfully generated using both $k_O = 2$ and $k_O = 20$, the path obtained with $k_O = 20$ is selected. If only one path is successfully generated, that path is selected.

The selected FIM segment is concatenated with the prescribed initial straight-line displacement described in Section 3.2.7. The resulting centroid path is used as a reference for subsequent ghost-guided physics validation using the complete moving-unit geometry.

## C.4 Reuse of previously evaluated paths

To reduce repeated path generation and validation, the framework reuses a previously evaluated path when the current path-generation problem is sufficiently similar to an already evaluated case. Path reuse requires that the moving units contain the same component set, that the rotation distance between their stored poses be less than $5°$, and that the dot product between their normalized initial separation directions exceed 0.98. The surrounding obstacle configurations must additionally exhibit sufficient spatial overlap.

Because the workspace is represented by a discrete FIM grid, similarity of the surrounding obstacle configurations is evaluated using their occupied grid cells. Let $\mathcal{O}_{\text{ref}}$ denote the set of obstacle-occupied cells in the reference configuration and $\mathcal{O}_{\text{cur}}$ denote the corresponding set in the current configuration. The obstacle-overlap ratio is defined as

$$\rho_{\text{obs}} = \frac{|\mathcal{O}_{\text{ref}} \cap \mathcal{O}_{\text{cur}}|}{|\mathcal{O}_{\text{ref}}|}. \tag{C.6}$$

A previously evaluated path is considered reusable when

$$\rho_{\text{obs}} > 0.8. \tag{C.7}$$

Thus, at least 80% of the obstacle-occupied grid cells used for the original path generation must remain occupied at the same grid locations in the current configuration. The ratio is defined asymmetrically with respect to the reference configuration because disassembly state transitions generally remove components and thereby reduce the occupied obstacle region. The criterion therefore evaluates how much of the obstacle configuration used for the original path generation remains at the same grid locations in the current state. The threshold is configurable and provides an approximate criterion for path reuse rather than a guarantee of exact geometric equivalence.

## Appendix D. Identification and release association of cylindrical contact-rotatable relations

This appendix details the geometric identification and release-source association of the contact-rotatable relations introduced in Section 3.1.3. The association between a release-source component $R$ and a contact-rotatable edge $(P, Q)$ follows Eq. (10). Figure D1 summarizes the adopted surface-based method.

### D.1 Identification of cylindrical contact-rotatable relations

For each candidate contact edge between parts, connected cylindrical patches are identified independently on its incident components $P$ and $Q$ using shared triangle edges, surface normals, and cylindrical consistency, as shown in Figure D1(a,b). Let $\mathcal{T}_X$ denote an identified connected cylindrical patch on $X \in \{P, Q\}$. The unit triangle normals of $\mathcal{T}_X$, shown in Figure D1(b), form the rows of

$$\boldsymbol{N}_X = \begin{bmatrix} (\boldsymbol{n}_1^X)^T \\ (\boldsymbol{n}_2^X)^T \\ \vdots \\ (\boldsymbol{n}_{N_X}^X)^T \end{bmatrix}. \tag{D.1}$$

An initial axis direction is obtained from the right singular vector associated with the smallest singular value of $N_X$, equivalently,

$$\boldsymbol{a}_X^{(0)} = \underset{\|\boldsymbol{b}\|=1}{\arg\min} \parallel \boldsymbol{N}_X \boldsymbol{b} \parallel. \tag{D.2}$$

Here, $\boldsymbol{b} \in \mathbb{R}^3$ is a unit vector representing a trial axis direction. The distinct vertices of $\mathcal{T}_X$ are projected onto the circle-fitting plane $\Pi_X^{\text{fit}}$, which is normal to $\boldsymbol{a}_X^{(0)}$ and passes through their mean position. A circle is fitted to the projected vertices by linear least squares, and its center and radius initialize an independent cylinder fit. The axis direction, axis reference point, and radius may subsequently be refined against the vertex radial residuals. Nearly planar patches are rejected based on their normal-direction spread. The remaining patches must yield a nondegenerate circle fit and satisfy the normal-axis perpendicularity, radial-consistency, and radial-normal alignment criteria listed in Table D1.

Each accepted patch therefore retains its own fitted axis line and radius, as illustrated in Figure D1(c):

$$L_X(s) = \boldsymbol{c}_X + s\boldsymbol{a}_X,\ r_X > 0,\ X \in \{P, Q\}. \tag{D.3}$$

Oppositely facing patches are compared, with $B$ denoting the inward-facing bore and $S$ the outward-facing solid. After sign alignment, their mean direction is

$$\boldsymbol{a} = \frac{\boldsymbol{a_B} + \tilde{\boldsymbol{a}}_{\boldsymbol{S}}}{\| \boldsymbol{a_B} + \tilde{\boldsymbol{a}}_{\boldsymbol{S}} \|}, \quad \tilde{\boldsymbol{a}}_{\boldsymbol{S}} = \begin{cases} \boldsymbol{a_S}, & \boldsymbol{a_B} \cdot \boldsymbol{a_S} \geq 0, \\ -\boldsymbol{a_S}, & \boldsymbol{a_B} \cdot \boldsymbol{a_S} < 0. \end{cases} \tag{D.4}$$

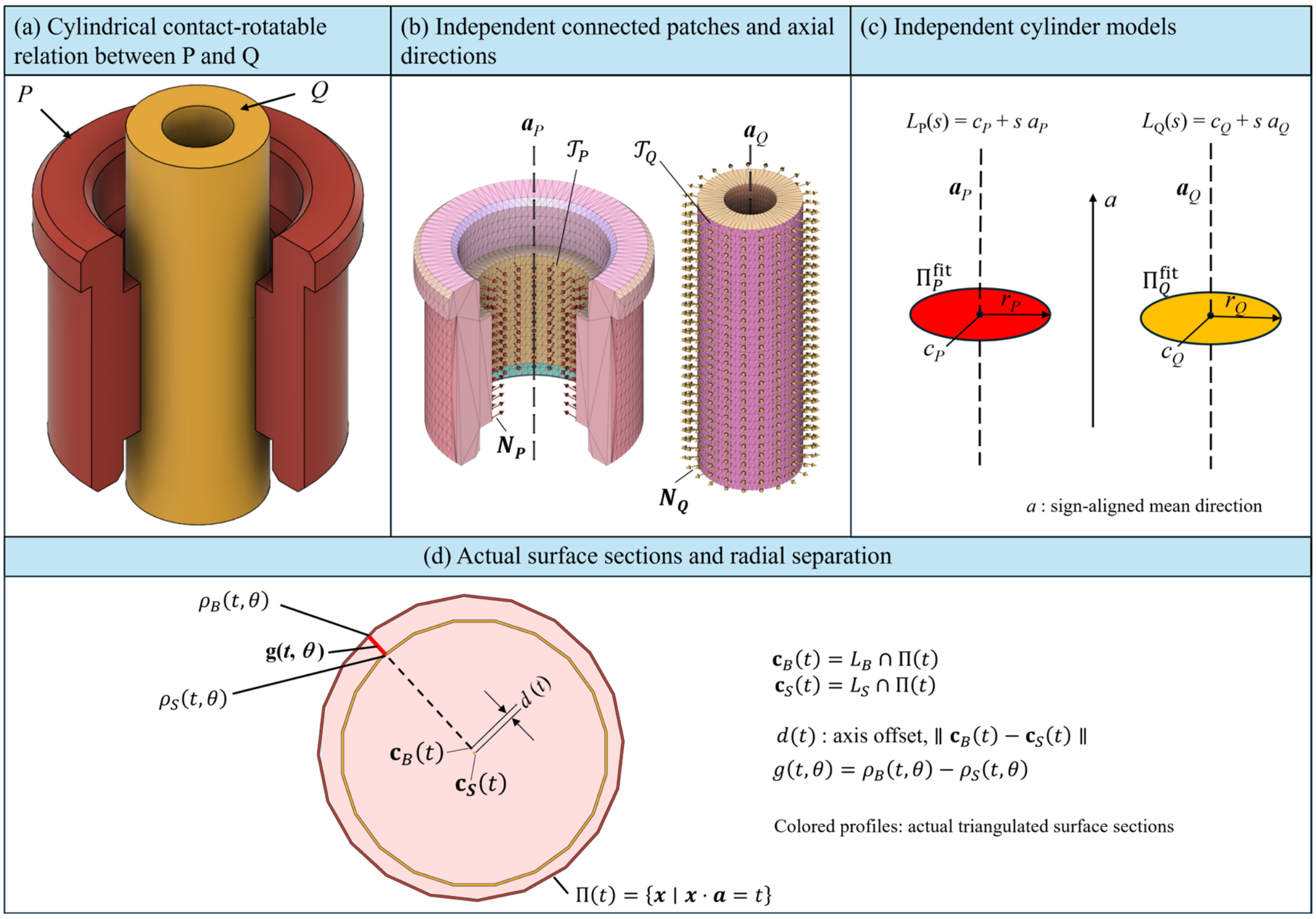

Figure D1. Surface-based identification of a cylindrical contact-rotatable relation between components $P$ and $Q$: (a) incident cylindrical interface; (b) independently identified connected cylindrical patches and their fitted axis directions; (c) independent cylinder models, including the respective axis reference points, axis directions, and fitted radii; and (d) actual triangulated surface sections in a common comparison plane, showing the transverse offset between the independently fitted axis lines and the radial surface separation $g(t,\theta)$ along a common ray. The comparison direction, finite common axial range, and axial discretization are defined in the text. Axis offsets and surface gaps may be enlarged schematically for visibility. In the illustrated configuration, $P = B$ and $Q = S$; the normal arrows in panel (b) represent the triangle normals forming the rows of $\boldsymbol{N}_P$ and $\boldsymbol{N}_Q$, respectively. An initial conceptual layout for this explanatory figure was generated with assistance from Microsoft 365 Copilot using GPT-6.1 Sol. The author subsequently redrew, restructured, and verified the final figure.

The vector $a$ defines a common comparison direction, not a unique common axis line. Projecting both patches onto $a$ gives their axial extents $[\ell_X, h_X]$ and the finite common range

$$[\ell, h] = \left[\max(\ell_P, \ell_Q), \min(h_P, h_Q)\right].$$

For $t \in [\ell, h]$, let

$$\Pi(t) = \{\boldsymbol{x} \mid \boldsymbol{x} \cdot \boldsymbol{a} = t\}$$

be the common comparison plane. The intersection of each independently fitted axis line with this plane is $\boldsymbol{c}_X(t)$, and the maximum transverse offset between the two lines over the common range is

$$\boldsymbol{c}_X(t) = \boldsymbol{c}_X + \frac{t - \boldsymbol{c}_X \cdot \boldsymbol{a}}{\boldsymbol{a}_X \cdot \boldsymbol{a}} \boldsymbol{a}_X,\ d_{\text{env}} = \max_{t \in \{\ell, h\}} \| \boldsymbol{c}_B(t) - \boldsymbol{c}_S(t) \|. \tag{D.5}$$

Using the maximum transverse axis offset $d_{\text{env}}$ defined in Eq. (D.5), the fitting-resolution allowance $\eta$, and the surface-gap threshold $\tau$ listed in Table D1, a resolved common range and compatible fitted models require

$$h - \ell > 2\eta,\ \eta < \tau,\ r_B - r_S - d_{\text{env}} \geq -\eta,\ \mid r_B - r_S \mid - d_{\text{env}} \leq \tau + \eta. \tag{D.6}$$

No explicit inter-axis angle threshold is imposed. Instead, axis offset and inclination affect the finite-range screening and are also assessed on sampled cross sections of the original triangulated patches $\mathcal{T}_B$ and $\mathcal{T}_S$.

In each comparison plane, angular rays originate at the bore-axis position $c_B(t)$, as shown in Figure D1(d). For each direction $\theta$ supported by both surface sections, the nearest positive intersections define the radial positions $\rho_B(t, \theta)$ and $\rho_S(t, \theta)$ from the same origin. Their signed radial surface separation is

$$g(t, \theta) = \rho_B(t, \theta) - \rho_S(t, \theta). \tag{D.7}$$

The common range is partitioned at projected mesh-vertex positions. Radial gaps are evaluated at representative cross sections within the resolved intervals and weighted proportionally to the corresponding interval lengths. For nonempty retained support, acceptance requires

$$\min g \geq -\eta,\ Q_{0.90}^{w}(g) \leq \tau, \tag{D.8}$$

where $Q_{0.90}^{w}(g)$ denotes the weighted 90th percentile of the sampled radial gaps.

For an accepted relation, the midpoint coordinate $t_c = (\ell + h)/2$ and the corresponding bore-axis position $\boldsymbol{p}_a = \boldsymbol{c}_B(t_c)$ define the representative line $\boldsymbol{p}_a + s\boldsymbol{a}$. The registered axial range is

$$I_{PQ} = [\ell - t_c, h - t_c]. \tag{D.9}$$

The relation also retains the range length, mean fitted radius, and weighted mean angular coverage. The representative line and overall range are subsequently used for release-source association; section-supported subintervals are not propagated.

### D.2 Association with release-source components

For each accepted edge $(P, Q)$, release-source candidates $R$ are examined in increasing liaison-graph distance from $P$ and $Q$, up to two edges, excluding traversal across the target edge. Fastener-mediated constraint edges and contact-rotatable edges are traversable. Nut candidates are excluded, and all geometrically acceptable candidates at the first successful distance are retained.

For bolt-like and thread-like candidates, the previously estimated longitudinal axis must satisfy the axis-alignment and axis-line proximity criteria defined in Table D1 with respect to the registered line $\boldsymbol{p}_a + s\boldsymbol{a}$. Thread-envelope regions are then identified using criteria appropriate to the inclined thread-flank geometry.

For other candidates, cylindrical surface regions are verified relative to the registered line using normal-axis perpendicularity, radial consistency, radial-normal alignment, and angular coverage.

For either candidate type, the accepted triangles are projected onto $\boldsymbol{a}$ relative to $\boldsymbol{p}_a$. Their projected extents are combined into continuous intervals without bridging disjoint axial gaps. One such interval,

$$J_R = [\alpha_R, \beta_R],$$

must fully contain the registered overall range:

$$\alpha_R \leq \ell - t_c,\ h - t_c \leq \beta_R. \tag{D.10}$$

No axial expansion is applied. The containment test uses the overall range in Eq. (D.9), rather than the axial intervals at which individual surface sections were evaluated during detection.

The association is based on liaison-graph proximity and geometric consistency of the corresponding surface regions. It does not independently evaluate joint load capacity, frictional retention, or dynamic stability.

### D.3 Principal parameters

Table D1 lists the principal parameters of the adopted surface method and release-source association. Lengths are expressed in millimeters. CV and RMS denote coefficient of variation and root mean square, respectively. Independent-patch radial variation is normalized by the fitted radius, whereas general release-candidate variation is normalized by the mean radial distance.

Table D1. Principal parameters for cylindrical surface identification and release-source association

| Parameter | Value | Purpose |
|---|---|---|
| Surface radial-gap threshold $\tau$ | 0.05 mm | Maximum weighted 90th-percentile supported surface gap; also spatial broad-phase expansion |
| Minimum independent patch size | 6 triangles | Cylinder fitting |
| Planar-direction rejection | RMS spread < 0.05 and 90th-percentile angular deviation < $8°$ | Reject nearly uniform patch normals or common section directions |
| Axis-perpendicularity tolerance | Mean absolute unit-normal projection $\leq 0.15$ | Independent patch fitting; growth uses a per-face bound of 0.15 |
| Cylindrical radial consistency | Radius CV $\leq 0.08$ | Independent patch acceptance; connected growth bounds vertex residuals by 0.08 times seed radius |
| Radial-normal alignment | Median absolute alignment $\geq$ cos20° | Independent patch acceptance |
| Fit uncertainty $\eta$ | Maximum of $10^{-6}$ mm and summed maximum radial residuals | Intersection allowance and resolution of axial ranges/slabs, whose width must exceed $2\eta$ |
| Surface-section sampling | 512 uniform angular rays; 3 axial stations at 0.1, 0.5, 0.9 of each resolved slab | Actual shared-section gap evaluation |
| Gap aggregation | Weighted 90th percentile; weight $\Delta t/3$ per angular sample | Final clearance criterion |
| Maximum release-source distance | 2 graph edges | Candidate search |
| Thread-like axis consistency | 12°; 0.3 mm | Longitudinal direction and maximum head/tip distance from registered line |
| Thread-envelope selection | Projected normal magnitude > 0.15; absolute radial alignment $\geq$ cos20° | Select envelope triangles; radial distance must exceed $10^{-6}$ mm |
| General release-cylinder consistency | Normal-axis projection $\leq 0.15$; radius CV $\leq 0.08$; median radial alignment $\geq$ cos20° | Smooth cylindrical support about the registered line |
| General release radial clustering | Maximum of 0.001 mm and 0.08 times median radial distance | Separate radial groups |
| Release-source support | At least 24 samples and $240°$ angular coverage | Both thread-envelope and general cylindrical candidates |
| Release-source axial containment | One continuous interval; zero expansion | Full containment of the overall $I_{PQ}$ |

# Supplementary Material

*Recognition results, structural verification records, training configuration, and reproducibility settings*

**Scope.** This document consolidates the supplementary evidence and implementation settings for the virtual-disassembly knowledge-acquisition study. Result tables are presented first, followed by network-training conditions and numerical settings not fully specified in the main text or Appendices A-D. Duplicate drafts, editable inventories, and parameters already reported in the appendices have been omitted.

## S1. Consistency-filtered fastener-recognition results

These are application results for the two evaluated assemblies, not held-out estimates of standalone network accuracy. A component was retained as a fastener only when the predicted class satisfied the required attribute conditions; otherwise, its final label was UNKNOWN.

**Table S1. Component-family results in the controlled toy assembly.**

| Component name in the liaison graph | Original CAD name | Component specification | Count | Known type | Final label | Outcome |
|---|---|---|---|---|---|---|
| M8_bolt, M8_bolt(2), M8_bolt(3) | M8_bolt | M8 x 12 mm | 3 | Bolt | bolt_like_hex | 3 / 3 retained |
| M5_bolt, …, M5_bolt(6) | Bolt | M5 x 8 mm | 6 | Bolt | bolt_like_hex | 6 / 6 retained |
| M5_bolt(7) | Bolt_M5x8 | M5 x 8 mm cap screw | 1 | Bolt | bolt_like_cap | 1 / 1 retained |
| nut_M8 | nut_M8 | M8 nut | 3 | Nut | nut_like | 3 / 3 retained |
| Base, Int, TopSnapM, InnerLL, Lid, Plate | | | 6 | Non-fastener | UNKNOWN | 6 / 6 rejected |

**Table S2. Component-family results in the OpenManipulator Chain gripper.**

| Component name in the liaison graph | Original CAD name | Component specification | Count | Known type | Final label | Outcome |
|---|---|---|---|---|---|---|
| Bolt | Bolt | M2 x 4 mm cap screw | 4 | Bolt | bolt_like_cap | 4 / 4 retained |
| Bolt | Bolt | M2.5 x 4 mm cap screw | 4 | Bolt | bolt_like_cap | 4 / 4 retained |
| Bolt | Bolt | M2.5 x 12 mm cap screw | 4 | Bolt | bolt_like_cap | 4 / 4 retained |
| Bolt_M2_5x8 | Bolt_M2_5x8 | M2.5 x 8 mm cap screw | 8 | Bolt | bolt_like_cap | 8 / 8 retained |
| Bolt_M3x10 | Bolt_M3x10 | M3 x 10 mm cap screw | 4 | Bolt | bolt_like_cap | 4 / 4 retained |
| Nut_M2_5 | Nut_M2_5 | M2.5 nut | 8 | Nut | nut_like | 8 / 8 retained |
| Nut_M3 | Nut_M3 | M3 nut | 4 | Nut | nut_like | 4 / 4 retained |
| RAIL_BLOCK | RAIL_BLOCK | | 2 | Non-fastener | UNKNOWN | 2 / 2 rejected |
| RAIL_BRACKET | RAIL_BRACKET | | 2 | Non-fastener | UNKNOWN | 2 / 2 rejected |
| FLANGE_BUSH | FLANGE_BUSH | | 4 | Non-fastener | UNKNOWN | 4 / 4 rejected |

| Component name in the liaison graph | Original CAD name | Component specification | Count | Known type | Final label | Outcome |
|---|---|---|---|---|---|---|
| LINK_ROD | LINK_ROD | | 2 | Non-fastener | UNKNOWN | 2 / 2 rejected |
| PALM_GRIPPER | PALM_GRIPPER | | 2 | Non-fastener | UNKNOWN | 2 / 2 rejected |
| servo_motor | servo_motor | | 1 | Non-fastener | UNKNOWN | 1 / 1 rejected |
| CRANK_ARM | CRANK_ARM | | 1 | Non-fastener | UNKNOWN | 1 / 1 rejected |
| x_sp | x_sp | | 4 | Non-fastener | UNKNOWN | 4 / 4 rejected |

**Table S3. Representative non-fastener predictions rejected by attribute-consistency filtering.**

| Assembly | Component | Raw class prediction | Raw attribute prediction | Consistency-filter reason | Final label |
|---|---|---|---|---|---|
| Toy | Base | nut_like | None | Required attributes threadhole_present and hexagon_present were not predicted | UNKNOWN |
| Toy | Plate | nut_like | None | Required attributes threadhole_present and hexagon_present were not predicted | UNKNOWN |
| OpenManipulator Chain | LINK_ROD | nut_like | thread_present | Required attributes threadhole_present and hexagon_present were not predicted, while thread_present was inconsistent with nut_like | UNKNOWN |
| OpenManipulator Chain | CRANK_ARM | nut_like | thread_present | Required attributes threadhole_present and hexagon_present were not predicted, while thread_present was inconsistent with nut_like | UNKNOWN |
| OpenManipulator Chain | x_sp spacer | nut_like | None | Required attributes threadhole_present and hexagon_present were not predicted | UNKNOWN |

**Summary.** All 13 known fasteners in the controlled toy assembly and all 36 known fasteners in the OpenManipulator Chain gripper were retained after attribute-consistency filtering. All 24 known non-fastener components were excluded from the final fastener set as UNKNOWN.

## S2. Inferred fastener sets and interface associations

The tables report the final operation-level bolt sets. Contacting nuts are paired with their bolts and included in the same constraint-release operation; nut-contact edges removed during scene updating are not treated as separate part-to-part release objectives. Contact-rotatable release-source associations are listed in Table S7.

**Table S4. Inferred fastener sets and associated interfaces in the controlled toy assembly.**

| Set ID | Membership | Component name in the liaison graph | Paired nuts | Associated interface edge(s) | Nested-set note |
|---|---|---|---|---|---|
| T1 | 3 M5_bolt | M5_bolt(4); M5_bolt(5); M5_bolt(6) | None | TopSnapM–InnerLL | Local set |
| T2 | 3 M5_bolt | M5_bolt; M5_bolt(2); M5_bolt(3) | None | Int–Lid; Lid–TopSnapM | Superset containing T4 and T5 |
| T3 | 3 M8_bolt | M8_bolt, M8_bolt(2), M8_bolt(3) | nut_M8, nut_M8(2), nut_M8(3) | Base–Int | Three one-to-one bolt–nut pairs |
| T4 | 2 M5_bolt | M5_bolt; M5_bolt(2) | None | Int–Lid | T4 subset of T2 |
| T5 | 1 M5_bolt | M5_bolt(3) | None | Lid–TopSnapM | T5 subset of T2 |
| T6 | 1 M5_bolt | M5_bolt(7) | None | TopSnapM–Plate | Local set |

**Table S5. Inferred fastener sets and associated interfaces in the OpenManipulator Chain gripper.**

| Set ID | Membership | Bolt component paths | Paired nuts | Associated interface edge(s) | Nested-set note |
|---|---|---|---|---|---|
| O1 | 12 bolts (8 M2.5 and 4 M2) | MeshInstance_31/Bolt<br>MeshInstance_32/Bolt<br>MeshInstance_33/Bolt<br>MeshInstance_34/Bolt<br>MeshInstance_35/Bolt<br>MeshInstance_36/Bolt<br>MeshInstance_37/Bolt<br>MeshInstance_38/Bolt<br>MeshInstance_39/Bolt<br>MeshInstance_40/Bolt<br>MeshInstance_41/Bolt<br>MeshInstance_42/Bolt | None | MeshInstance/RAIL_BRACKET - MeshInstance_30/servo_motor<br>MeshInstance_1/RAIL_BRACKET - MeshInstance_30/servo_motor<br>MeshInstance_30/servo_motor - MeshInstance_47/CRANK_ARM | Nested superset spanning O4-O6 |
| O2 | 4 M2.5 bolts | MeshInstance_12/Bolt_M2_5x8<br>MeshInstance_13/Bolt_M2_5x8<br>MeshInstance_14/Bolt_M2_5x8<br>MeshInstance_15/Bolt_M2_5x8 | MeshInstance_8/Nut_M2_5<br>MeshInstance_9/Nut_M2_5<br>MeshInstance_10/Nut_M2_5<br>MeshInstance_11/Nut_M2_5 | MeshInstance_5/RAIL_BLOCK - MeshInstance_7/PALM_GRIPPER | Four bolt-nut pairs |
| O3 | 4 M2.5 bolts | MeshInstance_26/Bolt_M2_5x8<br>MeshInstance_27/Bolt_M2_5x8<br>MeshInstance_28/Bolt_M2_5x8<br>MeshInstance_29/Bolt_M2_5x8 | MeshInstance_22/Nut_M2_5<br>MeshInstance_23/Nut_M2_5<br>MeshInstance_24/Nut_M2_5<br>MeshInstance_25/Nut_M2_5 | MeshInstance_19/RAIL_BLOCK - MeshInstance_21/PALM_GRIPPER | Four bolt-nut pairs |
| O4 | 4 M2.5 bolts | MeshInstance_31/Bolt<br>MeshInstance_32/Bolt<br>MeshInstance_35/Bolt<br>MeshInstance_36/Bolt | None | MeshInstance/RAIL_BRACKET - MeshInstance_30/servo_motor | O4 subset of O1 |
| O5 | 4 M2.5 bolts | MeshInstance_33/Bolt<br>MeshInstance_34/Bolt<br>MeshInstance_37/Bolt<br>MeshInstance_38/Bolt | None | MeshInstance_1/RAIL_BRACKET - MeshInstance_30/servo_motor | O5 subset of O1 |
| O6 | 4 M2 bolts | MeshInstance_39/Bolt<br>MeshInstance_40/Bolt<br>MeshInstance_41/Bolt<br>MeshInstance_42/Bolt | None | MeshInstance_30/servo_motor - MeshInstance_47/CRANK_ARM | O6 subset of O1 |

| Set ID | Membership | Bolt component paths | Paired nuts | Associated interface edge(s) | Nested-set note |
|---|---|---|---|---|---|
| O7 | 1 M3 bolt | MeshInstance_17/Bolt_M3x10 | MeshInstance_16/Nut_M3 | MeshInstance_18/FLANGE_BUSH - MeshInstance_20/LINK_ROD (contact-rotatable) | Single-bolt operation |
| O8 | 1 M3 bolt | MeshInstance_45/Bolt_M3x10 | MeshInstance_46/Nut_M3 | MeshInstance_44/FLANGE_BUSH - MeshInstance_20/LINK_ROD (contact-rotatable) | Single-bolt operation |
| O9 | 1 M3 bolt | MeshInstance_49/Bolt_M3x10 | MeshInstance_48/Nut_M3 | MeshInstance_43/FLANGE_BUSH - MeshInstance_6/LINK_ROD (contact-rotatable) | Single-bolt operation |
| O10 | 1 M3 bolt | MeshInstance_3/Bolt_M3x10 | MeshInstance_2/Nut_M3 | MeshInstance_4/FLANGE_BUSH - MeshInstance_6/LINK_ROD (contact-rotatable) | Single-bolt operation |

Note: Associated interface edge(s) lists the interfaces inferred during Phase 1. Additional direct liaison edges associated with paired nuts are identified from the current liaison graph and removed during virtual fastener removal.

## S3. Verification of residual and contact-rotatable relations

The records compare automatically acquired relation annotations with the represented assembly structures. The toy-assembly record concerns an interlock-like residual interface; the gripper records concern release-source associations for contact-rotatable relations.

**Table S6. Residual-interface relation identified in the controlled toy assembly.**

| Relation | State | Component P | Component Q | Acquired relation | Selected release direction | Graph result after edge release | Verification result |
|---|---|---|---|---|---|---|---|
| R1 | FullAssy | TopSnapM | Int | Interlock-like residual constraint | [0, 0, 1] (approximately +Z) | Connected after releasing TopSnapM–Int | Accepted relation; no graph division in FullAssy |

**Table S7. Contact-rotatable release-source associations in the OpenManipulator Chain gripper.**

| Release source R | Constrained component P | Constrained component Q | Interpretation |
|---|---|---|---|
| MeshInstance_45/Bolt_M3x10 | MeshInstance_20/LINK_ROD | MeshInstance_44/FLANGE_BUSH | Left-side cylindrical relation |
| MeshInstance_49/Bolt_M3x10 | MeshInstance_6/LINK_ROD | MeshInstance_43/FLANGE_BUSH | Right-side cylindrical relation |
| MeshInstance_17/Bolt_M3x10 | MeshInstance_18/FLANGE_BUSH | MeshInstance_20/LINK_ROD | Left-side cylindrical relation |
| MeshInstance_3/Bolt_M3x10 | MeshInstance_4/FLANGE_BUSH | MeshInstance_6/LINK_ROD | Right-side cylindrical relation |

## S4. Exhaustive verification of toy-model constraint-release branching

The toy assembly was used as a controlled model for exhaustive verification of the state-dependent constraint-release branches recorded in the generated T-AOG. The verification combined an independent audit of the annotated liaison structure and operation rules with runtime tool-accessibility records for the three-bolt operation highlighted in the main text.

Seven constraint-release operations were included in the exhaustive audit: the six fastener-set removals T1–T6 defined in Table S4 and the TopSnapM–Int residual-interface release R1 defined in Table S6. Geometrically overlapping fastener sets were retained as distinct operation identities when they released different component-level liaison edges.

**Table S8. Detailed verification of the controlled branching conditions.**

| Controlled operation | Evidence and eligibility condition | Evaluated state nodes | Target-present state nodes | Correctly generated | Correctly suppressed | Violations |
|---|---|---|---|---|---|---|
| T1 removal: M5_bolt(4), M5_bolt(5), and M5_bolt(6) | The threaded tips remained externally visible. The operation was eligible only when all three bolt heads were accessible to their tool proxies after release of either the Int-Base contact or the TopSnapM-Int contact. | 90 | 37 | 26 | 11 | 0 |
| R1 release: TopSnapM–Int residual interface | Int and TopSnapM were present, and either the two-bolt subset, the single-bolt subset, or their combined three-bolt set had been removed. | 90 | 47 | 39 | 8 | 0 |

**Note:** Evaluated state nodes are the reachable state nodes whose reconstructed liaison graph has a single connected component, regardless of whether the target of the corresponding operation is present. Each state node is counted once, irrespective of the number of operation histories reaching it.

**Tool-accessibility cross-check.** The three-bolt operation target was present in 37 of the 90 evaluated state nodes. All three bolt heads were tool-accessible in 26 states, and the corresponding operation branch was correctly generated in all 26 states. Tool access was denied in 11 states, and the operation was correctly suppressed in all 11 states. No operation was generated when complete bolt-head-side tool access was unavailable, and no eligible operation was omitted.

**State-history audit procedure.** For each state and each reachable operation history, the active liaison graph was reconstructed from the raw liaison graph by removing components absent from the state, their incident edges, and the liaison edges released by preceding operations. Connected components were recomputed without using separation child nodes. States whose active liaison graph consisted of one connected component were evaluated for subsequent constraint-release operations; multi-component intermediate states awaiting separation processing were excluded from the missing-operation check.

The acquired T-AOG contained 219 state nodes, comprising 90 states with one connected component in their active liaison graphs and 129 intermediate states with multiple connected components in their active liaison graphs. Alternative operation orders produced 246 reachable state-history combinations with one connected component in the active liaison graph. Evaluating all seven operation identities gave 1,722 state-history-operation evaluations.

**Table S9. Operation-wise results of the exhaustive state-history verification.**

| Operation ID | Correctly generated | Correctly suppressed | Unexpected | Missing | Precision | Recall |
|---|---|---|---|---|---|---|
| T2 | 8 | 238 | 0 | 0 | 1.000 | 1.000 |
| T3 | 28 | 218 | 0 | 0 | 1.000 | 1.000 |
| T1 | 51 | 195 | 0 | 0 | 1.000 | 1.000 |
| T4 | 36 | 210 | 0 | 0 | 1.000 | 1.000 |
| T5 | 35 | 211 | 0 | 0 | 1.000 | 1.000 |
| T6 | 47 | 199 | 0 | 0 | 1.000 | 1.000 |
| R1 | 52 | 194 | 0 | 0 | 1.000 | 1.000 |

**Note:** Each operation identity was evaluated for all 246 reachable state-history combinations whose reconstructed liaison graph had a single connected component. Correctly suppressed includes both state-history combinations in which the operation target was absent and those in which the target was present but the eligibility condition was not satisfied.

No missing, unexpected, or duplicate state-operation pair was detected. No operation was generated when its required components were absent or when only part of a fastener set remained. The independently reconstructed connected-component classification also agreed with the T-AOG separation-child structure for all 219 states.

**Scope of the result.** This verification establishes soundness and completeness of state-dependent constraint-release branching with respect to the annotated toy-model liaison structure and specified operation rules. It does not establish global completeness with respect to unmodeled release mechanisms, alternative tool configurations, compliant behavior, or separation motions outside the configured motion-generation scope.

## S5. Multi-view fastener-recognition training configuration

The table reports the optimizer, batch sizes, data split, and module-specific learning rates used for attribute pretraining and twelve-view fine-tuning.

**Table S10. Training configuration for the multi-view fastener-recognition network.**

| Training stage | Setting | Module or sample unit | Value |
|---|---|---|---|
| Attribute pretraining | Optimizer | All trainable modules | AdamW |
| Attribute pretraining | Batch size | Single-view samples | 192 |
| Attribute pretraining | Learning rate | ResNet-34 backbone: conv1 and layers 1-4 | 1 x 10^-4 |
| Attribute pretraining | Learning rate | Attribute-prediction heads | 3 x 10^-5 |
| Twelve-view fine-tuning | Optimizer | All trainable modules | AdamW |
| Twelve-view fine-tuning | Batch size | Twelve-view samples | 16 |
| Twelve-view fine-tuning | Learning rate | ResNet-34 conv1 | 1 x 10^-6 |
| Twelve-view fine-tuning | Learning rate | ResNet-34 layer1 | 1 x 10^-6 |
| Twelve-view fine-tuning | Learning rate | ResNet-34 layer2 | 3 x 10^-5 |
| Twelve-view fine-tuning | Learning rate | ResNet-34 layer3 | 3 x 10^-5 |
| Twelve-view fine-tuning | Learning rate | ResNet-34 layer4 | 1 x 10^-4 |
| Twelve-view fine-tuning | Learning rate | Attention fully connected module | 1 x 10^-4 |
| Twelve-view fine-tuning | Learning rate | Fastener-class prediction head | 3 x 10^-5 |
| Twelve-view fine-tuning | Learning rate | Attribute-head weights | 1 x 10^-5 |
| Twelve-view fine-tuning | Learning rate | Attribute-head biases | 3 x 10^-5 |

Model selection. The final checkpoint was the epoch with the highest macro-average precision across all 20 attributes on the validation subset. No early-stopping rule was applied.

Data split. Both stages used 80% training and 20% validation subsets. Fine-tuning was split at the twelve-view sample level so that all views of the same geometry remained in the same subset.

Batch size. Attribute pretraining used 192 single-view images per batch. Fine-tuning used 16 twelve-view samples, corresponding to 192 rendered images per optimization step.

## S6. Additional reproducibility settings

Only settings that materially affect reproducibility and are not fully specified in the main text or Appendices A-D are listed. Each row presents one parameter or criterion together with its value, unit, and operational meaning.

**Table S11. Contact acquisition and common physics parameters.**

| Parameter or criterion | Value | Unit | Meaning / application |
|---|---|---|---|
| Stage length unit | 1 | mm | Length scale used for collision geometry and motion parameters. |
| Collision representation | SDF | - | Applied to all component collision meshes to represent concave and finely detailed contact geometries. |
| SDF resolution for bolts and nuts | 384 | - | Resolution used for SDF generation. The relatively compact bolt and nut geometries allow threaded features to be represented at this resolution. |
| SDF resolution for other components | 768 | - | Resolution used for SDF generation. The higher resolution preserves small threaded holes and other local contact features within components whose overall dimensions are substantially larger than those features. |
| Initial liaison-edge criterion | At least one Found or Persist event during two physics updates | - | Registers an undirected contact edge between two distinct components. |
| Pairwise contact-generation distance | 0.001 | mm | Maximum separation distance at which contact generation between two collision shapes can begin. |
| Contact offset per collision shape | 0.0005 | mm | Assigned to each collision shape to produce the pairwise contact-generation distance of 0.001 mm. |
| Rest offset | 0.0 | mm | No positive equilibrium clearance or prescribed geometric penetration was introduced. |
| Contact-report force threshold | 0.0 | N | Minimum reported contact force for initial liaison-edge acquisition; not a geometric clearance threshold. |
| Physics timestep $\Delta t_{sim}$ | 1/120 | s | Physics integration interval. |
| Position iterations | 16 | iterations/step | Position-level contact and constraint solve. |
| Velocity iterations | 1 | iteration/step | Velocity-level contact and constraint solve. |
| Ambient gravitational acceleration | 0 | m/s^2 | Disabled during the virtual evaluations. |
| Static friction coefficient | 0.30 | - | Default component material outside ghost-guided path validation. |
| Dynamic friction coefficient | 0.15 | - | Default component material outside ghost-guided path validation. |
| Directional probing acceleration, $a_{\mathrm{probe}}$ | 7.2 | mm/s^2 | Applied successively along $+X, -X, +Y, -Y, +Z$, and $-Z$ during initial liaison-graph construction. Defined as $a_{\mathrm{probe}} = 2\delta_{\mathrm{probe}}/(2\Delta t_{\mathrm{sim}})^2$. |
| Reference probing distance, $\delta_{\mathrm{probe}}$ | 0.001 | mm | Reference displacement used to determine the perturbation acceleration. |

**Table S12. Visibility and tool-access implementation parameters.**

| Parameter or criterion | Value | Unit | Meaning / application |
| --- | --- | --- | --- |
| Camera directions | 12 | views | Directions corresponding to the vertices of an icosahedron. |
| Image resolution | 600 x 600 | pixels | Used for semantic visibility images and isolated-component MVCNN images. |
| MVCNN component framing | Bounding-sphere normalization | - | Normalizes isolated components of different physical sizes to a consistent proportion of the camera view used during fine-tuning. |
| Visibility threshold | 1 | pixel in any view | A component is visible when its semantic unique identifier appears in at least one pixel in at least one camera view. |
| Tool-access collision criterion | No Found event involving the proxy | - | Implementation of the collision-free placement condition in Section 3.2.2. |
| State-visibility camera framing | Initial-assembly framing | - | Camera positions and look-at point retained across state transitions. |

**Table S13. Fastener-size estimation and removal parameters.**

| Parameter or criterion | Value | Unit | Meaning / application |
| --- | --- | --- | --- |
| Axial slice centers | 0.48, 0.50, 0.52 | fraction of bolt length | Locations used to estimate the nominal shank diameter. |
| Slice width | 0.10 | fraction of bolt length | Axial width of each mesh slice used for diameter estimation. |
| Minimum points per slice | 50 | mesh points | Rejects under-sampled slices. |
| Slice-radius statistic | 95th percentile | - | Estimates the outer shank radius while reducing sensitivity to isolated mesh extremes. |
| Cross-slice aggregation | 2 x median valid radius | - | Produces the final estimated diameter. |
| Metric-size tolerance | 10 | % relative deviation | Assigns the estimated diameter to a supported nominal metric size; otherwise Other. |
| Fastener-direction grouping tolerance | 15 | degrees | Assigns a removal direction to an existing group when its angle from the closest current representative direction does not exceed this value. |
| Maximum fastener-direction groups | 3 | groups | Creates at most three direction groups; after this limit is reached, each subsequent direction is assigned to the group with the closest current representative direction. |
| Axial holding force | 0.5 | N | Reduces bolt recontact after thread disengagement. |
| Removal-trial limit | Loop counter > 1000 | iterations | Nominally about 8.3 s at the 1/120 s physics timestep. |
| Bolt-removal completion threshold | $L_b$ | mm | $L_b$ is the axial length of each bolt; all operated bolts must reach this projected displacement along their estimated removal directions. |

**Table S14. Size-dependent torque values used in the final rerun.**

| Bolt size | Torque (N m) |
|---|---|
| M2 | 0.18 |
| M2.5 | 0.35 |
| M3 | 0.68 |
| M4 | 1.60 |
| M5 | 3.20 |
| M6 | 5.40 |
| M8 | 13.1 |
| M10 | 26.0 |
| M12 | 45.3 |
| Other | 45.3 |

The listed values were applied as torques about the estimated bolt axes, not as angular accelerations.

**Table S15. Controller settings for ghost-guided physics validation.**

| Parameter or criterion | Value | Unit | Meaning / application |
|---|---|---|---|
| Position proportional gain, kp,pos | 100 | s^-1 | Translational velocity command generated from positional error. |
| Position damping gain, kd,pos | 0.5 | - | Damps the current linear velocity. |
| Rotation proportional gain, kp,rot | 0 | s^-1 | Exact orientation tracking is not imposed. |
| Rotation damping gain, kd,rot | 0.5 | - | Damps the current angular velocity. |
| Linear-speed scaling coefficient, $c_v$ | 1.0 | - | Configurable coefficient used to scale the linear-velocity limit relative to the FIM grid spacing and physics timestep. |
| Maximum linear speed | $c_v$ h / dt | mm/s | Configurable upper bound on commanded linear velocity, scaled by FIM grid spacing h and physics timestep dt. |
| Maximum angular speed | 360 | degrees/s | Angular-velocity command limit. |
| Maximum tracking steps per point | 30 | physics steps | Allows at most 0.25 s for each ghost point at 120 Hz. |
| Immediate rejection threshold | 30h | mm | Rejects the candidate path if positional tracking error exceeds this value. |

**Table S16. Loading, reaction-sampling, and direction-selection settings for residual interlock-like direction extraction.**

| Parameter or criterion | Value | Unit | Meaning / application |
|---|---|---|---|
| Loading mode | Acceleration | – | Mode used for the opposing endpoint loads. |
| Loading magnitude | 200.0 per endpoint | Simulator acceleration input | Equal magnitudes with opposite signs; no SI-unit conversion is asserted. |

| Parameter or criterion | Value | Unit | Meaning / application |
|---|---|---|---|
| Initial waiting interval | 5 | Simulation updates | Waiting interval after simulation playback and loading begin, before reaction sampling starts. |
| Reaction sampling count | 5 | Samples per trial | Contact reports collected after the initial waiting interval. |
| Sampling interval | 1 | Simulation update | Reports cleared before each update; the resulting contacts contribute to that sample. |
| Contact-contribution tolerance | $1\times10^{-12}$ | Simulator impulse inputs | Numerical tolerance applied to the normal-impulse magnitude and fallback impulse norm; contributions remaining below this tolerance are discarded. |
| Opposing-direction threshold | -0.99 | Dot product | Maximum dot product for treating the two trial-derived candidate directions as nearly opposite. |
| Endpoint-center alignment threshold | 0.95 | Absolute dot product | Minimum alignment required for using the endpoint-center direction to resolve the sign ambiguity between nearly opposite trial-derived directions. |
| Endpoint-OBB-axis alignment threshold | 0.98 | Absolute dot product | Minimum alignment required to replace the reaction-derived direction with the nearest endpoint OBB axis while preserving its sign. |

**Parameters reported elsewhere.** The occupancy- and FIM-grid adaptation rule, point-arrival threshold, zero-friction validation condition, and collision handling are specified in Section 3.2.7. FIM propagation and path-selection parameters are specified in Appendix C. Residual-interface and initial-direction screening rules are specified in Sections 3.1.3 and 3.2.5. Cylindrical contact-rotatable relation parameters are specified in Appendix D and are not repeated here.